\documentclass[conference]{IEEEtran}
\usepackage{times}

\usepackage[numbers]{natbib}
\usepackage{multicol}

\usepackage[utf8]{inputenc} % allow utf-8 input
\usepackage[T1]{fontenc}    % use 8-bit T1 fonts
\usepackage{url}            % simple URL typesetting
\usepackage{booktabs}       % professional-quality tables
\usepackage{amsfonts}       % blackboard math symbols
\usepackage{amsbsy}
\usepackage{nicefrac}       % compact symbols for 1/2, etc.
\usepackage{microtype}      % microtypography
\usepackage{subcaption}

\usepackage{graphicx}
\usepackage{graphics}
\usepackage{multirow}
\usepackage{xcolor}
\usepackage{amsmath}
\usepackage{float}
\usepackage[bottom]{footmisc}
\usepackage{wrapfig}

\usepackage{algorithm}
\usepackage{algorithmic}

\DeclareMathOperator*{\argmax}{arg\,max}
\DeclareMathOperator*{\argmin}{arg\,min}

\newcommand*\diff{\mathop{}\!\mathrm{d}}

\newcommand{\bs}{\mathbf{s}}

\newcommand{\met}{RHPO}
\newcommand{\method}{Regularized Hierarchical Policy Optimization}

\begin{document}

\title{Compositional Transfer in \\Hierarchical Reinforcement Learning}

\author{
\authorblockN{Markus Wulfmeier\textsuperscript{*}, 
    Abbas Abdolmaleki\textsuperscript{*},
    Roland Hafner,
    Jost Tobias Springenberg,\\
    Michael Neunert,
    Tim Hertweck,
    Thomas Lampe,
    Noah Siegel,
    Nicolas Heess,
    Martin Riedmiller}
\authorblockA{DeepMind,
London, United Kingdom}
}

\maketitle

\begin{abstract}
The successful application of general reinforcement learning algorithms to real-world robotics applications is often limited by their high data requirements. We introduce Regularized Hierarchical Policy Optimization (RHPO) to improve data-efficiency for domains with multiple dominant tasks and ultimately reduce required platform time. To this end, we employ compositional inductive biases on multiple levels and corresponding mechanisms for sharing off-policy transition data across low-level controllers and tasks as well as scheduling of tasks. The presented algorithm enables stable and fast learning for complex, real-world domains in the parallel multitask and sequential transfer case. We show that the investigated types of hierarchy enable positive transfer while partially mitigating negative interference and evaluate the benefits of additional incentives for efficient, compositional task solutions in single task domains. Finally, we demonstrate substantial data-efficiency and final performance gains over competitive baselines in a week-long, physical robot stacking experiment. 
\end{abstract}

\IEEEpeerreviewmaketitle

\footnotetext[1]{Correspondence to: mwulfmeier, abdolmaleki@google.com Shared first-authorship. }

\section{Introduction}\label{sec:introduction}

Creating real-world systems that learn to achieve many goals directly through interaction with their environment is one of the long-standing dreams in robotics. Although recent successes in deep (reinforcement) learning for computer games (Atari \citep{mnih2013playing}, StarCraft \citep{alphastarblog}), Go \citep{silver2017mastering} and other simulated environments (e.g. \citep{openai2018dexterous}) have demonstrated the potential of these methods when large amounts of training data are available, the high cost of data acquisition has limited progress for many problems involving systems directly acting in the physical world. 

Data efficiency in machine learning generally relies on inductive biases or prior knowledge to guide and accelerate the learning process. One strategy for injecting prior knowledge that is widely and successfully used in robotics learning problems is the use of human expert demonstrations to bootstrap the learning process. But the perspective of a system with a permanent embodiment capable of achieving many goals in a persistent environment provides us with a complementary opportunity: an efficient learning strategy should allow us to share and reuse experience across tasks --  such that the system does not have to experience or learn the same thing multiple times, and such that solutions to simpler tasks can bootstrap the learning of harder ones.

Rather than providing prior knowledge or biases specific to a particular task this suggests focusing on more general inductive biases that facilitate the sharing and reuse of experience and knowledge across tasks while allowing other aspects of the domain to be learned \citep{caruana1997multitask}. 
Previous approaches to transfer learning have, for example, built on optimizing initial parameters \citep[e.g.][]{finn2017model}, sharing models and parameters across tasks either in the form of policies or value functions \citep[e.g.][]{rusu2016progressive,teh2017distral,galashov2018information}, data-sharing across tasks \citep[e.g.][]{riedmiller2018learning,andrychowicz2017hindsight}, or through the use of task-related auxiliary objectives \citep{jaderberg2016reinforcement,wulfmeier2017mutual}. 
Transfer between tasks can, however, lead to either constructive or destructive transfer for humans \citep{singley1989transfer} as well as for machines \citep{pan2010survey,torrey2010transfer}. That is, jointly learning to solve different tasks can provide both benefits and disadvantages for individual tasks, depending on their similarity. Finding a mechanism that enables transfer where possible but avoids interference is one of the long-standing research challenges.

\begin{figure}[t]
    \centering
    \begin{tabular}{ccc}
    \includegraphics[height=4.5cm]{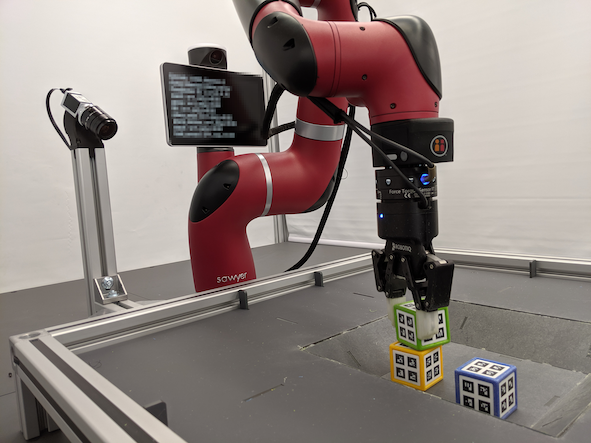} \\
    \includegraphics[height=1.8cm]{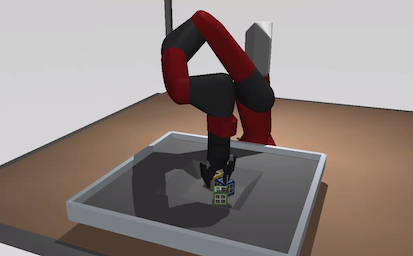} 
    \includegraphics[height=1.8cm]{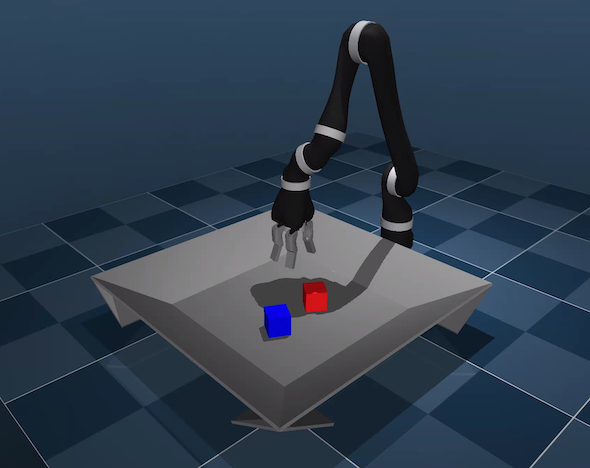} 
    \includegraphics[height=1.8cm]{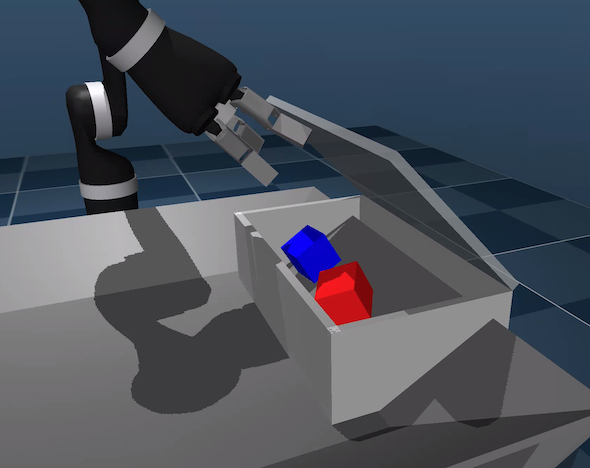}
    \end{tabular}
    \caption{\small Top: Overview of the real robot setup with the Sawyer robot performing the Pile1 task. Screen pixelated for anonymization. Bottom Left: Simulated Sawyer performing the same task. Bottom Middle \& Right: Respectively Pile2 \& Cleanup2 setup with a simulated Jaco arm. 
    }
    \label{fig:robot_photo}
    \vspace{-5mm}
\end{figure}

In this paper, we propose a general reinforcement learning architecture that benefits from learning multiple tasks simultaneously and is sufficiently data-efficient and reliable to solve non-trivial manipulation tasks from scratch directly on robotics hardware. We achieve efficiency through three forms of transfer: (1) robust off-policy learning allows to effectively share all generated transition data across tasks and skills; (2) a modular hierarchical policy architecture allows skills to be directly reused across tasks; and (3) switching between the execution of policies for different tasks within a single episode leads to effective exploration.

The model uses deep neural networks to parameterize state-conditional Gaussian mixture distributions as agent policies, similar to Mixture Density Networks \citep{bishop1994mixture}. To obtain robust and versatile low-level behaviors in the multitask setting we shield the mixture components from information about the task at hand. Task information is thus only communicated through the choice of mixture component by the high-level controller, and the mixture components are trained as domain-dependent but task-independent skills. To efficiently optimize hierarchical policies in a multitask setting, %end-to-end, 
we develop robust off-policy learning schemes enabling us to use all transition data to train each low-level controller independent of the actually executed one. We focus on Maximum A-Posteriori Policy Optimization (MPO) \cite{abdolmaleki2018maximum} but also consider a variant of Stochastic Value Gradients (SVG) \citep{heess2015learning}. For both algorithms we employ trust-region like constraints at both levels of the hierarchy.

We evaluate the approach on several real and simulated robotics manipulation tasks and demonstrate that it outperforms competitive baselines. In particular, it dramatically improves data efficiency on a challenging real-world robotics manipulation task similar to the one considered in \cite{riedmiller2018learning}: 
Our model learns to stack blocks from scratch on a single Sawyer robot arm within about a week at which point it demonstrates up to three times higher performance compared to our baselines. 
We further perform a number of careful ablations. These highlight, among others, the importance of the hierarchical architecture and the importance of the trust-region like constraints for the stability of the learning scheme. 
Finally, to gain a better understanding of the role of this type of hierarchy in RL, we compare its benefits in the single task and multitask setting. We find that it shows clear benefits advantages in the multitask setting. However, it can fail to improve performance in the single-task case, where additional incentives are required to encourage component specialization similar to the multitask case. These results shed further light on the interaction of model and domain in RL.

In summary, our contributions are as follows,
\begin{itemize}
    \item Algorithmic improvements: 
    We propose a new method for robust and efficient off-policy optimization of hierarchical policies. Our approach controls the rate of change at both levels of the hierarchy via trust-region like constraints thus ensuring stable learning. Furthermore, it can use all data to train any given low-level component, independent of the component which generated the transition. This enables data efficient training with experience replay and data sharing across tasks.
    \item Performance improvements:
    We evaluate our approach on a range of real and simulated robotic manipulation domains. The results confirm that the algorithm scales to complex tasks and significantly reduces interaction time. Particular benefits arise in more complex task sets and the low-data regime. When learning to stack from scratch on the Sawyer robot arm in a week-long experiment, the approach demonstrates up to three times better performance for the most complex tasks.
    \item Investigation of benefits, shortcomings and requirements: We perform a careful analysis and ablation of our algorithm and its properties, highlighting in particular, the impact of individual algorithmic and environment properties, as well was the overall robustness to hyperparameter settings.
\end{itemize}

\section{Preliminaries}
 
We consider a multitask reinforcement learning setting with an agent operating in a Markov Decision Process (MDP) consisting of the state space $\mathcal{S}$, the action space $\mathcal{A}$, the transition probability $p(s_{t+1}|s_t,a_t)$ of reaching state $s_{t+1}$ from state $s_t$ when executing action $a_t$. The actions are drawn from a probability distribution over actions $\pi(a | s)$ referred to as the agent's policy. Jointly, the transition dynamics and policy induce the marginal state visitation distribution $p(s)$. The discount factor $\gamma$ together with the reward $r(s, a)$ gives rise to the expected reward, or value, of starting in state $s$ (and following $\pi$ thereafter) $V^{\pi}(s) = \mathbb{E}_{\pi}[\sum^\infty_{t=0} \gamma^t r(s_t, a_t) | s_0 = s, a_{t} \sim \pi(\cdot|s_t), s_{t+1} \sim  p(\cdot|s_t,a_t)]$. 
We define multitask learning over a set of tasks $i \in I$ with common agent embodiment as follows: We assume shared state and action spaces and shared transition dynamics; tasks only differ in their reward function $r_i(s, a)$. We consider task conditional policies $\pi(a | s, i)$ with the overall objective defined as

\begin{eqnarray}
\begin{aligned}
J(\pi) &=\mathbb{E}_{i \sim I}\Big[\mathbb{E}_{\pi, p\left(s_0\right)}\Big[\sum_{t=0}^{\infty} \gamma^{t} r_i\left(s_{t}, a_{t}\right) | s_{t+1} \sim p(\cdot|s_t,a_t) \Big]\Big] \nonumber\\
&= \mathbb{E}_{i \sim I}\Big[\mathbb{E}_{\pi, p\left(s\right)}\big[ Q^\pi(s, a, i) \big]\Big],
\end{aligned}
\end{eqnarray}

where all actions are drawn according to the policy $\pi$ conditioned on task $i$, that is, $a_t \sim \pi(\cdot | s_t, i)$ and we used the following definition of the task-conditional state-action value function (Equation \ref{q_task}). 

\begin{eqnarray}\label{q_task}
\begin{aligned}
Q^{\pi}(s,a,i) = \mathbb{E}_\pi \biggl[ \sum_{t=0}^{\infty} \gamma^{t} r_i\left(s_{t}, a_{t}\right) | a_0 = a, \\
s_0 = s, a_{t} \sim \pi(\cdot|s_t,i), s_{t+1} \sim  p(\cdot|s_t,a_t) \biggr] 
\end{aligned}
\end{eqnarray}

\section{Method}
This section introduces \method~(\met) which focuses on efficient training of modular policies by sharing data across tasks. 
We first describe the underlying class of mixture policies, followed by details on the critic-weighted maximum likelihood optimization objective used to update structured hierarchical policies in a multitask, off-policy setting. For efficiency in the multitask case, \met~extends data-sharing and scheduling mechanisms from Scheduled Auxiliary Control with randomized scheduling (SAC-U) \citep{riedmiller2018learning}.

\subsection{Hierarchical Policies}\label{sec:HierarchicalPolicies}
\begin{figure}[hb]
% \vskip 0.2in
\begin{center}
\centerline{
  \includegraphics[height=0.46\columnwidth]{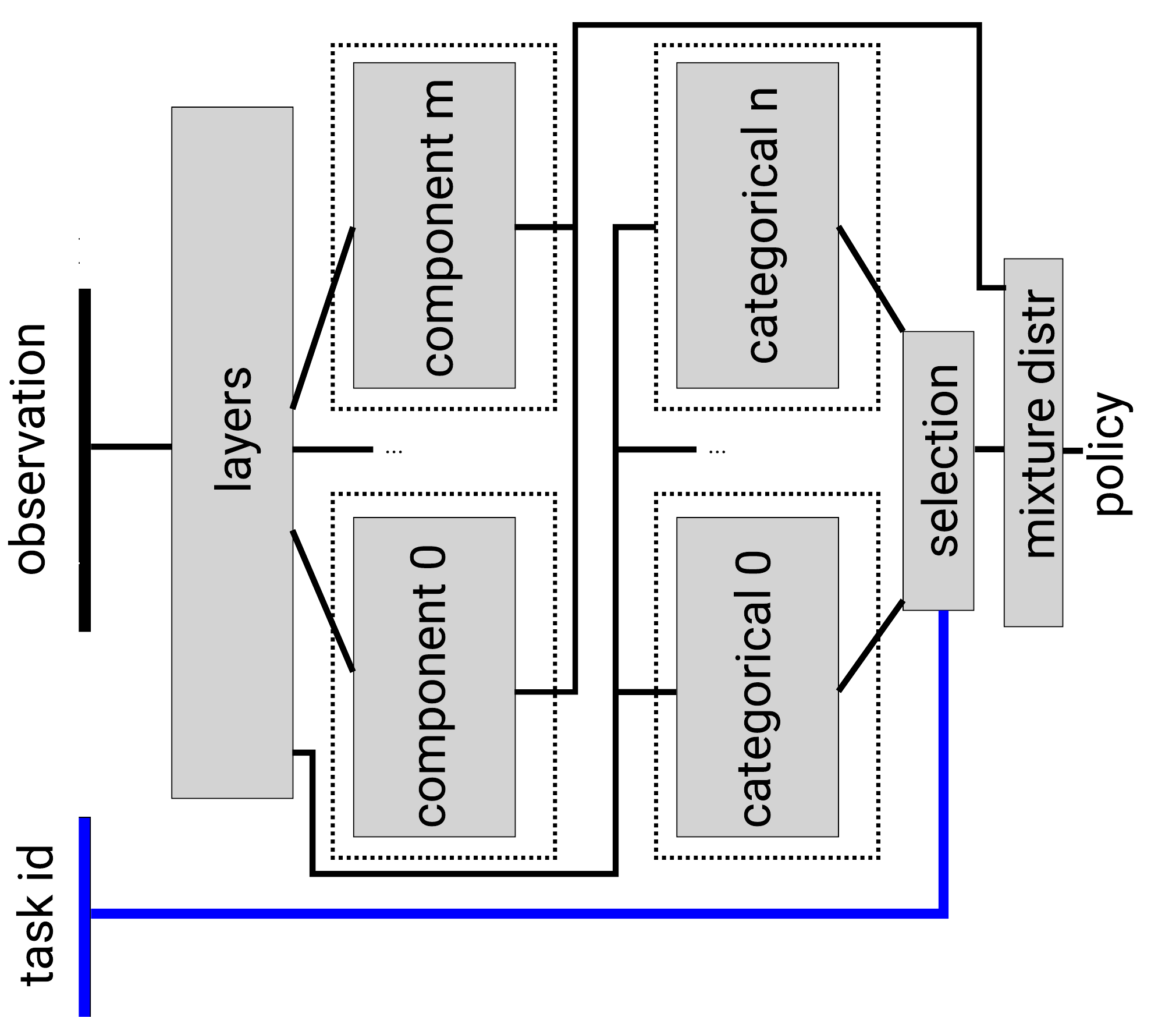}
}
\caption{\small \label{fig:sac_networks_alternative_policy_main}The hierarchical multitask policy architectures used in this paper. Note that only the high-level controller of mixture distribution is conditioned on the task ID and low level components are shared among tasks. A detailed description can be found in Appendix \ref{app:architectures}.}
\end{center}
% \vskip -0.2in
\end{figure}

We start by defining the hierarchical policy class which supports sharing sub-policies across tasks. Formally, we decompose the per-task policy $\pi(a | s, i)$ as
\begin{eqnarray}\label{eq:policy}
\pi_{\theta}(a | s, i) = \sum^M_{o=1} \pi^L_\theta \left(a | s, o\right) \pi^H_\theta\left(o | s, i\right),
\end{eqnarray}
where $\pi^H$ and $\pi^L$ respectively represent a high-level switching controller (a categorical distribution) and a low-level sub-policy (components of the resulting mixture distribution), and $o$ is the index of the sub-policy. $\theta$ denotes the parameters of both $\pi^H$ and $\pi^L$, which we seek to optimize. While the number of components has to be decided externally, \met~is robust with respect to this parameter (Appendix \ref{app:components}).
Note that in the above formulation {\it only the high-level controller $\pi_H$ is conditioned on the task information $i$}. This choice introduces a form of information asymmetry \citep{galashov2018information,tirumala2019exploiting,heess2016learning} that enables the low-level policies to acquire general, task-independent behaviours. 
This choice strengthens the decomposition of tasks across domains and prevents degenerate solutions that bypass the high-level controller. Intuitively, these sub-policies can be understood as building reflex-like low-level control loops, which perform domain-dependent but task-independent behaviours and can be modulated by higher cognitive functions with knowledge of the task at hand. Figure \ref{fig:sac_networks_alternative_policy_main} illustrates the used hierarchical policy architecture.  

\vspace{-6mm}
\subsection{Data-efficient Multitask Policy Optimization}\label{sec:method}
In the following sections, we present the core principles underlying \met; for the complete pseudocode algorithm please see Algorithm \ref{alg:learner} and Appendix \ref{sec:Pseudocode}. 
We build on an Expectation-Maximization based policy optimization algorithm (similar to MPO \citep{abdolmaleki2018relative}) and adapt it to the application to hierarchical policies in the multitask case. We update the parametric policy in 2 stages and decouple the policy improvement step from the fitting of the parametric policy. 

We begin by describing the policy improvement steps below, assuming that we have an approximation of the ground-truth state-action value function $\hat{Q}(s, a, i) \approx {Q}^\pi(s, a, i)$ available (see Equation \eqref{eq:objective_q_value} for details on learning $\hat{Q}$ from off-policy data). Starting from an initial policy $\pi_{\theta_0}$ we can then iterate the following steps to improve the policy $\pi_{\theta_k}$:

\vspace{2mm}

\begin{itemize}[leftmargin=*]
    \item[] \textbf{Policy Evaluation:} Update $\hat{Q}$ such that $\hat{Q}(s, a, i) \approx \hat{Q}^{\pi_{\theta_{k}}}(s, a, i)$, see Equation \eqref{eq:objective_q_value}.
    \item[] \textbf{Policy Improvement:}
    \begin{itemize}
        \item \textbf{Step 1:} Obtain $q_k = \arg \max_q J(q)$, under $\mathrm{KL}$ constraints with $\pi_{ref} = \pi_{\theta_k} $ (Equation \eqref{eq:objective_q}).
        \item \textbf{Step 2:}  Obtain \\ $\theta_{k+1} = \arg \min_{\theta} \mathbb{E}_{s \sim \mathcal{D}, i \sim I}\Big[ \mathrm{KL}\big( q_k(\cdot | s, i) \| \pi_{\theta}(\cdot | s, i) \big) \Big]$, under additional regularization (Equation \eqref{eq:objective_pi}).
    \end{itemize}
\end{itemize}

\vspace{2mm}

\paragraph{Policy Improvement 1: Obtaining Non-parametric Policies}
Concretely, we first introduce an intermediate non-parametric policy $q(a | s, i)$ and optimize $J(q)$ while staying close, in expectation, to a reference policy $\pi_{ref}(a | s, i)$ 
\begin{eqnarray}
  \begin{aligned}
    \max_{q} J(q) =\ \mathbb{E}_{i \sim I}&\Big[\mathbb{E}_{q, s \sim \mathcal{D}}\big[ \hat{Q}(s, a, i) \big]\Big], \\
    \text{s.t. } \mathbb{E}_{s \sim \mathcal{D}, i \sim I}&\Big[
    \mathrm{KL}\big(q(\cdot|s, i) \| \pi_{ref}(\cdot|s, i)\big)\Big] \le \epsilon,
  \end{aligned}
  \label{eq:objective_q}
\end{eqnarray}
where $\mathrm{KL}(\cdot \| \cdot)$ denotes the Kullback Leibler divergence, $\epsilon$ defines a bound on the KL, $\mathcal{D}$ denotes the data contained in a replay buffer. 

We find the intermediate policy $q$ by maximizing Equation \eqref{eq:objective_q} and can obtain a closed-form solution with a non-parametric policy for each task, as 
\begin{equation} 
q_k(a|s,i) \propto \pi_{\theta_k}(a|s,i) \exp\left({\frac{\hat{Q}(s,a,i)}{\eta}}\right),
\end{equation}
where $\eta$ is a temperature parameter (corresponding to a given bound $\epsilon$) that is obtained by optimizing the dual function,

\begin{eqnarray}
\begin{aligned}
    g(\eta) &= \eta\epsilon+\eta\mathbb{E}_{s \sim \mathcal{D}, i \sim I}\Big[\log\Big(\int \pi_{\theta_k}(a|s,i)\\
    &\exp\left(\frac{\hat{Q}(s,a,i)}{\eta}\right)\diff a\Big)\Big]
    \label{eq:dual_eta1},
\end{aligned}
\end{eqnarray}
(see Appendix \ref{sec:dualfunctionderivation} for a detailed derivation of the dual function).
This policy representation is independent of the form of the parametric policy $\pi_{\theta_k}$; i.e. $q$ only depends on $\pi_{\theta_k}$ through its requirement for obtaining samples. This, crucially, makes it easy to employ complicated structured policies (such as the one introduced in Section \ref{sec:HierarchicalPolicies}). The only requirement here, and in the following steps, is that we must be able to sample from $\pi_{\theta_k}$ and calculate the gradient (w.r.t. $\theta_k$) of its log density (but the sampling process itself need not be differentiable).

\paragraph{Policy Improvement  2: Fitting Parametric Policies}
In the second step we fit a policy to the non-parametric distribution obtained from the previous calculation by minimizing the divergence $
\mathbb{E}_{s \sim \mathcal{D}, i \sim I}[ \mathrm{KL}( q_k(\cdot | s, i) \| \pi_{\theta}(\cdot | s, i))].$
Assuming that we can sample from $q_k$ this step corresponds to maximum likelihood estimation (MLE). 
Furthermore, we introduce a trust-region constraint on policy updates. In this way, we can regularize towards a target policy, effectively mitigating optimization instabilities. 
Trust-region constraints have been used in on- and off-policy RL \citep{schulman2015trust,abdolmaleki2018relative}. We adapt the formulation of \citep{abdolmaleki2018relative} to our hierarchical setting, and as the analysis in Section \ref{sec:sim_multitask} shows, it is critical for the success of our algorithm.
Formally, we aim to obtain the solution in Equation \ref{eq:objective_pi}, where $\epsilon_{m}$ defines a bound on the change of the new policy. 

Here, we drop constant terms and the negative sign in the second line (turning $\min$ into $\max$), and explicitly insert the definition $\pi_\theta(a | s, i) = \sum^M_{o=1} \pi_L\left(a | s, o\right) \pi_H\left(o | s, i\right)$, highlighting that we are marginalizing over the high-level choices in this fitting step. The update is independent of the specific policy component from which the action was sampled, enabling joint updates of all components. This reduces the variance of the update and also enables efficient off-policy learning. 

\begin{eqnarray}
\begin{aligned}
    \theta_{k+1} &= \arg \min_{\theta} \mathbb{E}_{s \sim \mathcal{D}, i \sim I}\Big[ \mathrm{KL}\big( q_k(\cdot | s, i) \| \pi_{\theta}(\cdot | s, i) \big) \Big] \\
    & = \arg \max_{\theta} \mathbb{E}_{s \sim \mathcal{D}, i \sim I}\Bigg[ \mathbb{E}_{\pi_{\theta_k}} \Big[
    \exp({\nicefrac{\hat{Q}(s,a,i)}{\eta}}) 
    \\
    & {\log \sum^M_{o=1} \pi^L_\theta\left(a | s, o\right) \pi^H_\theta\left(o | s, i\right)
    \Big] \Bigg],} \\
    \text{s.t. } & \mathbb{E}_{s \sim \mathcal{D}, i \sim I} \Bigg[ \textrm{KL}(\pi^{H}_{\theta_k}(o | s, i) \| \pi^{H}_\theta(o | s, i)) + \\
    &\frac{1}{M}\sum_{o=1}^M \textrm{KL}(\pi^{L}_{\theta_k}(a | s, o) \| \pi^{L}_\theta(a | s, o)) \Bigg] < \epsilon_{m}
\end{aligned}
\label{eq:objective_pi}
\end{eqnarray}

Different approaches can be used to control convergence for both the high-level categorical choices and the action choices to change slowly throughout learning.
The average KL constraint in Equation  \eqref{eq:objective_pi} is similar in nature to an upper bound on the computationally intractable KL divergence between the two mixture distributions and has been determined experimentally to perform better in practice than simple bounds. 
In practice, in order to control the change of the high level and low level policies independently we decouple the constraints to be able to set different $\epsilon$ for the means ($\epsilon_{\mu}$), covariances ($\epsilon_{\Sigma}$) and the categorical distribution ($\epsilon_{\alpha}$) in case of a mixture of Gaussian policy. 
To solve Equation \eqref{eq:objective_pi}, we first employ Lagrangian relaxation to make it amenable to gradient based optimization and then perform a fixed number of gradient descent steps (using Adam \citep{kingma2014adam}); a detailed overview can be found in Algorithm \ref{alg:learner} as well as with further information in the Appendix \ref{sec:policyDerivation}.

\paragraph{Policy Evaluation} \label{par:policy_eval}
For data-efficient off-policy learning of $\hat{Q}$ we experience sharing across tasks and switching between tasks within one episode for improved exploration by adapting the initial state distribution of each task based on other tasks \citep{riedmiller2018learning}.

Formally, we assume access to a replay buffer containing data gathered from all tasks. 
For each trajectory snippet $\tau = \lbrace (s_0, a_0, R_0), \dots, (s_L, a_L, R_L) \rbrace$ we record the rewards for all tasks $R_t = [r_{i_1}(s_t, a_t), \dots, r_{i_{|I|}}(s_t, a_t)]$ as a vector in the buffer. Using this data we define the retrace objective for learning $\hat{Q}$, parameterized via $\phi$, following \citep{munos2016safe, riedmiller2018learning} as
\begin{eqnarray}
\begin{aligned}
\min_\phi L(\phi) = \sum_{i \sim I} \mathbb{E}_{\tau \sim \mathcal{D}} \Big[ \big( r_i(s_t, a_t) + \label{eq:objective_q_value} \\
\gamma Q^{ret}(s_{t+1}, a_{t+1}, i) - \hat{Q}_\phi(s_t, a_t, i))^2 \Big], 
\end{aligned}
\end{eqnarray}
where $Q^{ret}$ is the L-step retrace target \citep{munos2016safe}, see the Appendix \ref{sec:qtrace} for details.

\begin{algorithm}[t]
\caption{RHPO - Asynchronous Learner}\label{alg:learner}
\begin{algorithmic}
\STATE \textbf{Input:} $N_{steps}$ number of update steps, $N_\text{targetUpdate}$ update steps between target update, $N_s$ number of action samples per state, KL regularization parameters $\epsilon$, initial parameters for $\pi,~\eta$ and $\phi$
\STATE initialize N = 0
\WHILE{$k \leq N_\text{steps}$}
% \STATE update replay buffer $B$ with received trajectories
\FOR{$k$ in $[0...N_\text{targetUpdate}]$}
\STATE sample a batch of trajectories $\tau$ from replay buffer $B$ 
\STATE sample $N_s$ actions from $\pi_{\theta_k}$ to estimate expectations below
\STATE // compute mean gradients over batch for policy, Lagrangian multipliers and Q-function
\STATE $\delta_\pi \leftarrow -\nabla_\theta \sum_{s_t \in \tau} \sum_{j=1}^{N_s} [ \exp\left(\frac{Q(s_t,a_j,i)}{\eta}\right)$
\STATE \hfill $\log \pi_{\theta}(a_j | s_t, i) ] $ following Eq. \ref{eq:objective_pi}
\STATE $\delta_{\eta} \leftarrow \nabla_\eta g(\eta) = \nabla_\eta \eta\epsilon+\eta \sum_{s_t \in \tau} \log \frac{1}{N_s} \sum_{j=1}^{N_s}[ $
\STATE \hfill $\exp\left(\frac{Q(s_t,a_j,i)}{\eta}\right) ]$ following Eq. \ref{eq:dual_eta1}
\STATE $\delta_Q \leftarrow \nabla_{\phi} \sum_{i \sim I} \sum_{(s_t, a_t) \in \tau} \big( \hat{Q}_\phi(s_t, a_t, i) -
  Q^{\text{ret}} \big)^2$ 
\STATE \hfill  with $Q^{\text{ret}}$ following Eq. \ref{eq:objective_q_value}
\STATE // apply gradient updates
\STATE $\pi_{\theta_{k+1}} =$ optimizer\_update($\pi$, $\delta_\pi$), 
\STATE $\eta =$ optimizer\_update($\eta$, $\delta_\eta$) 
\STATE $\hat{Q}_\phi =$ optimizer\_update($\hat{Q}_\phi$, $\delta_Q$)
\STATE $k = k + 1$
\ENDFOR
\STATE // update target networks
\STATE $\pi' = \pi$, $Q' = Q$
\ENDWHILE
\end{algorithmic}
\end{algorithm}

\section{Experiments}\label{sec:experiments}
In the following sections, we investigate the effects of training hierarchical policies in single and multitask domains. In particular, we demonstrate that \met~can provide compelling benefits for multitask learning in real and simulated robotic manipulation tasks and significantly reduce platform interaction time. 
For the final experiment, a stacking task on a physical Sawyer robot arm, \met~ achieves a dramatic performance improvement after a week of training compared to several strong baselines.
We further investigate \met~in a sequential transfer setting and find that when pre-trained skills (i.e.\ low-level components) are available \met~can provide additional improvements in data efficiency. 

Finally, we perform a number of ablations to emphasize the importance of trust-region constraints for the high-level controller and to understand the relative role of hierarchy in the single-task and multitask setting:
In the single-task case, using domains from the DeepMind Control Suite \citep{tassa2018deepmind}, we first demonstrate that our hierarchy on its own can fail to improve performance and that for the model to exploit compositionality in this setting, additional incentives for component specialization are required.

For all tasks and algorithms, we use a distributed actor-critic framework (similar to \citep{espeholt2018impala}) with flexible hardware assignment \citep{buchlovsky2019tf}. We perform critic and policy updates from a replay buffer, which is asynchronously filled by a set of actors. In all figures with error bars, we visualize mean and variance derived from 3 runs.
Additional details of task hyperparameters as well as results for ablations and the full set of tasks from the multitask domains are provided in the Appendix \ref{sec:ExperimentalSetup}. \footnote{Additional details and the appendix can be found under \url{https://sites.google.com/corp/view/rhpo}}

\subsection{Simulated Multitask Experiments}\label{sec:sim_multitask}

We use three simulated multitask scenarios with the Kinova Jaco and Rethink Robotics Sawyer robot arms to test in a variety of conditions. The three scenarios each consist of tasks of different difficulties and vary in their overall complexity. The least difficult scenario is \textbf{Pile1}: Here, the seven tasks of interest range from simple reaching for a block over tasks like grasping it, 
to the final task of stacking the block on top of another block. 
The two more difficult scenarios are \textbf{Pile2} and \textbf{Cleanup2}. \textbf{Pile2} includes stacking with both blocks on top of the respective other block, resulting in a setting with 10 tasks. \textbf{Cleanup2} includes harder tasks such as opening a box and placing blocks into this box, consisting of a total of 13 tasks. 
In addition to the experiments in simulation, which are executed with 5 actors in a distributed setting, we also investigate the \textbf{Pile1} multitask domain (same rewards and setup) on a single, physical robot in Section \ref{sec:real_multitask}. 

Our main comparison evaluates \textbf{RHPO} with hierarchical policies against SAC \citep{riedmiller2018learning} with a flat, monolithic policy shared across all tasks which is provided with the additional task id as input (displayed as \textbf{SAC-U-Monolithic}) as well as policies with task dependent heads (displayed as \textbf{SAC-U-Independent}) following \citep{riedmiller2018learning} -- both using MPO as the optimizer. Furthermore, we compare against a re-implementation of SAC using SVG \citep{heess2015learning} as actor-critic based optimizer which uses the reparameterization trick (displayed as SAC-U[SVG]).  In order to compare with gradient-based hierarchical policy updates (such option critic \citep{bacon2017option}) as well as investigating the application of the proposed hierarchical model for other RL algorithms; we also use SVG (with continuous relaxation of the Categorical distribution \citep{maddison2016concrete, jang2016categorical}) instead of MPO to optimize the hierarchical model with results included in the Pile1 experiments (displayed as RHPO[SVG]) with additional results in Appendix \ref{app:svg_experiments}. These comparisons furthermore strengthen our choice for critic-weighted likelihood instead of reparametrization gradient-based policy optimizer.

The main SAC baselines provide the two opposite, naive perspectives on transfer: by using the same monolithic policy across tasks we enable positive as well as negative interference and independent policies prevent policy-based transfer. After experimentally confirming the robustness of \met~with respect to the number of low-level sub-policies (see Appendix \ref{app:components}), we set $M$ proportional to the number of tasks in each domain.

\begin{figure}[h]
    \centering
    \begin{tabular}{cc}
    \includegraphics[trim=30 0 0 0,width=.22\textwidth]{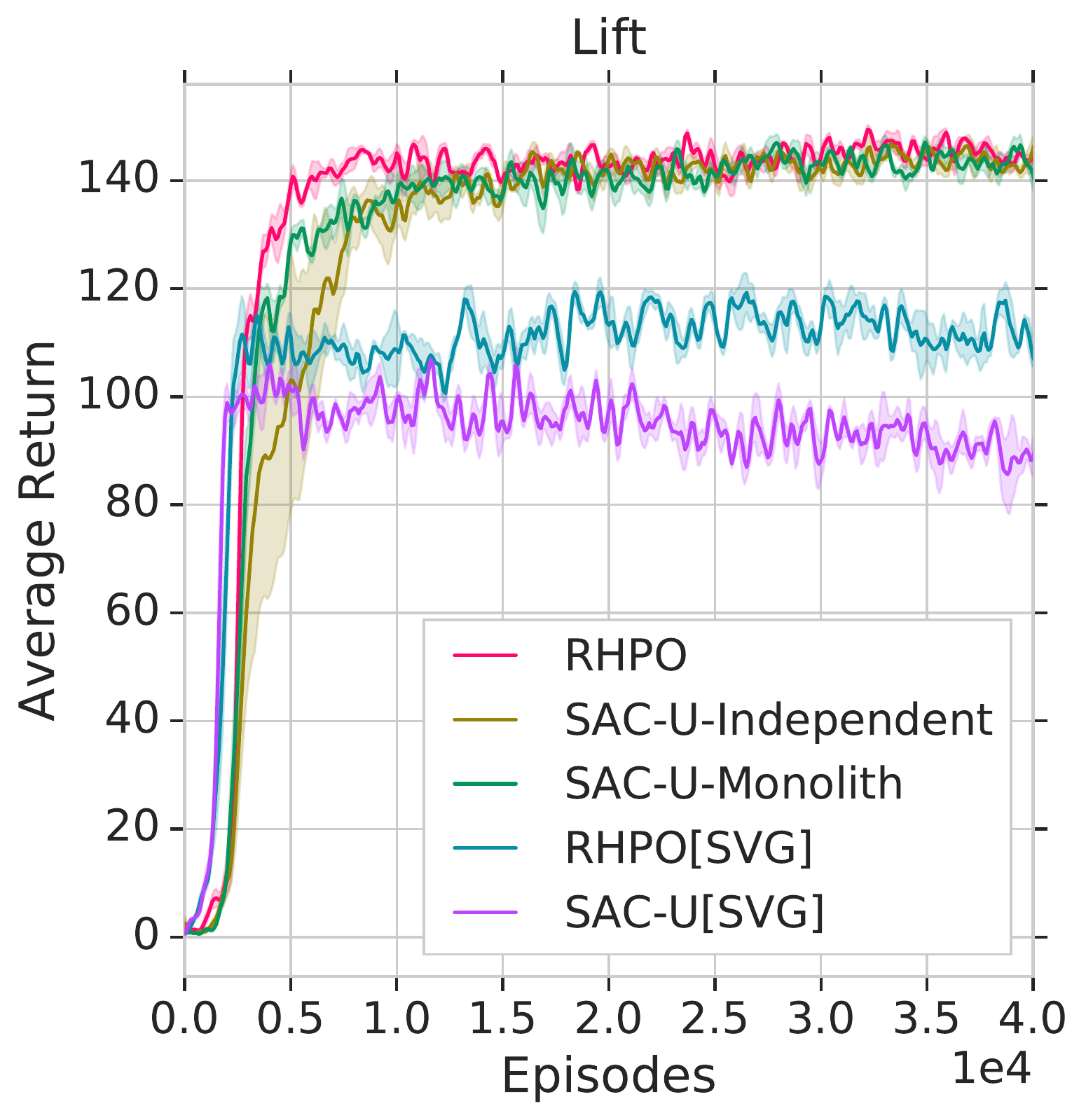} &
    \includegraphics[trim=30 0 0 0,width=.22\textwidth]{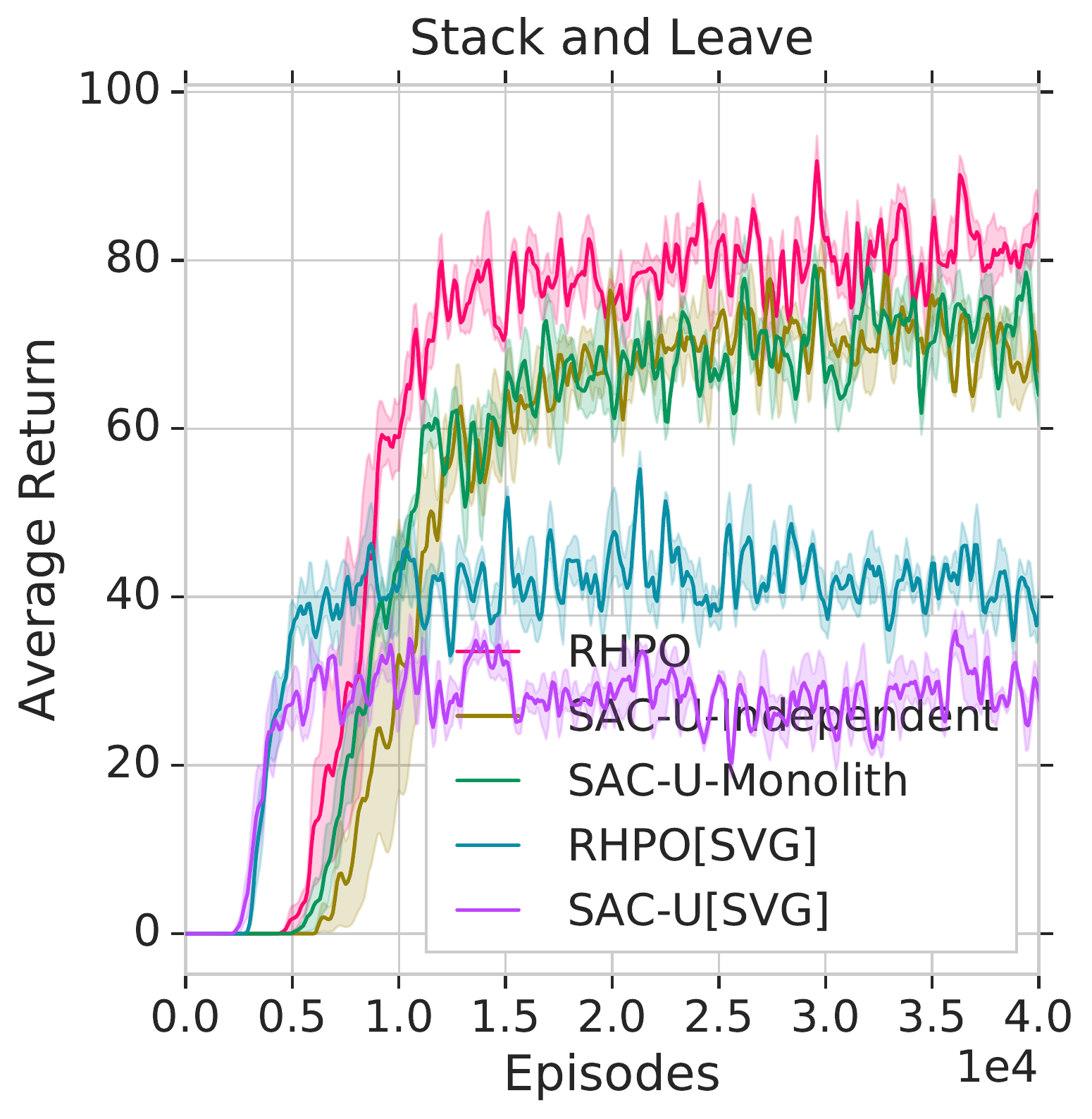} \\ 
    \includegraphics[trim=30 0 0 0,width=.22\textwidth]{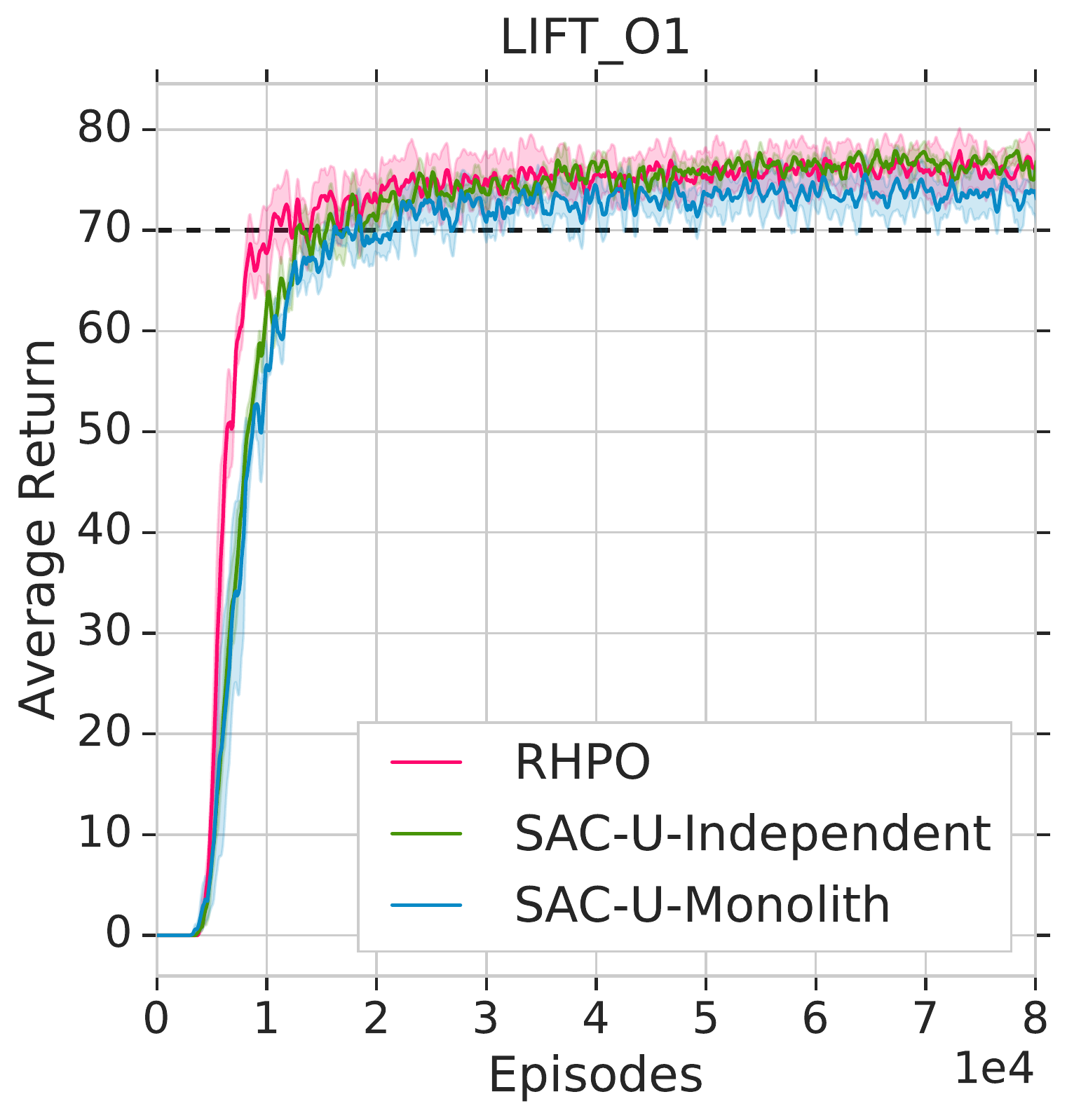} &
    \includegraphics[trim=30 0 0 0,width=.22\textwidth]{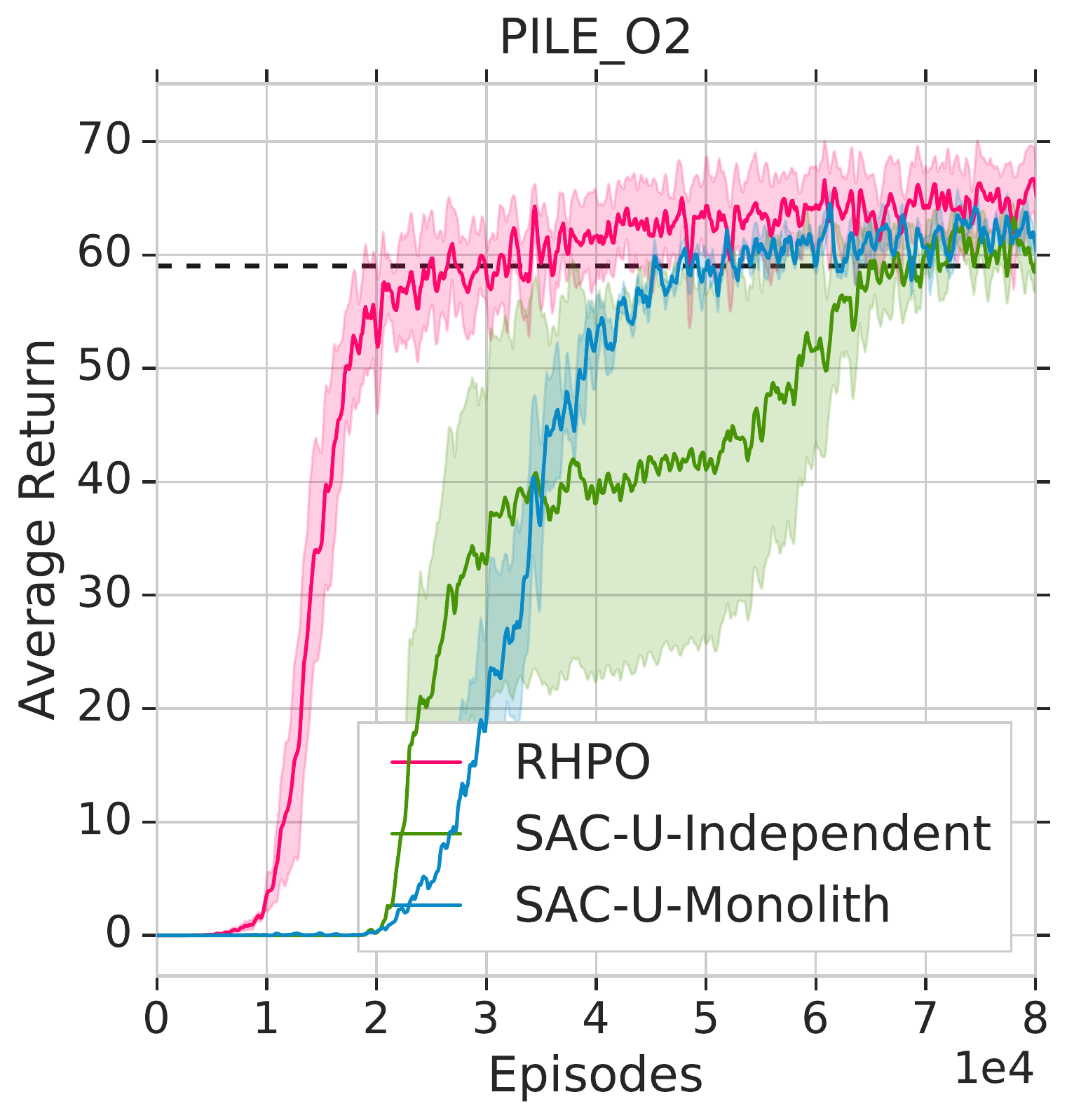}  \\
    \includegraphics[trim=30 0 0 0,width=.22\textwidth]{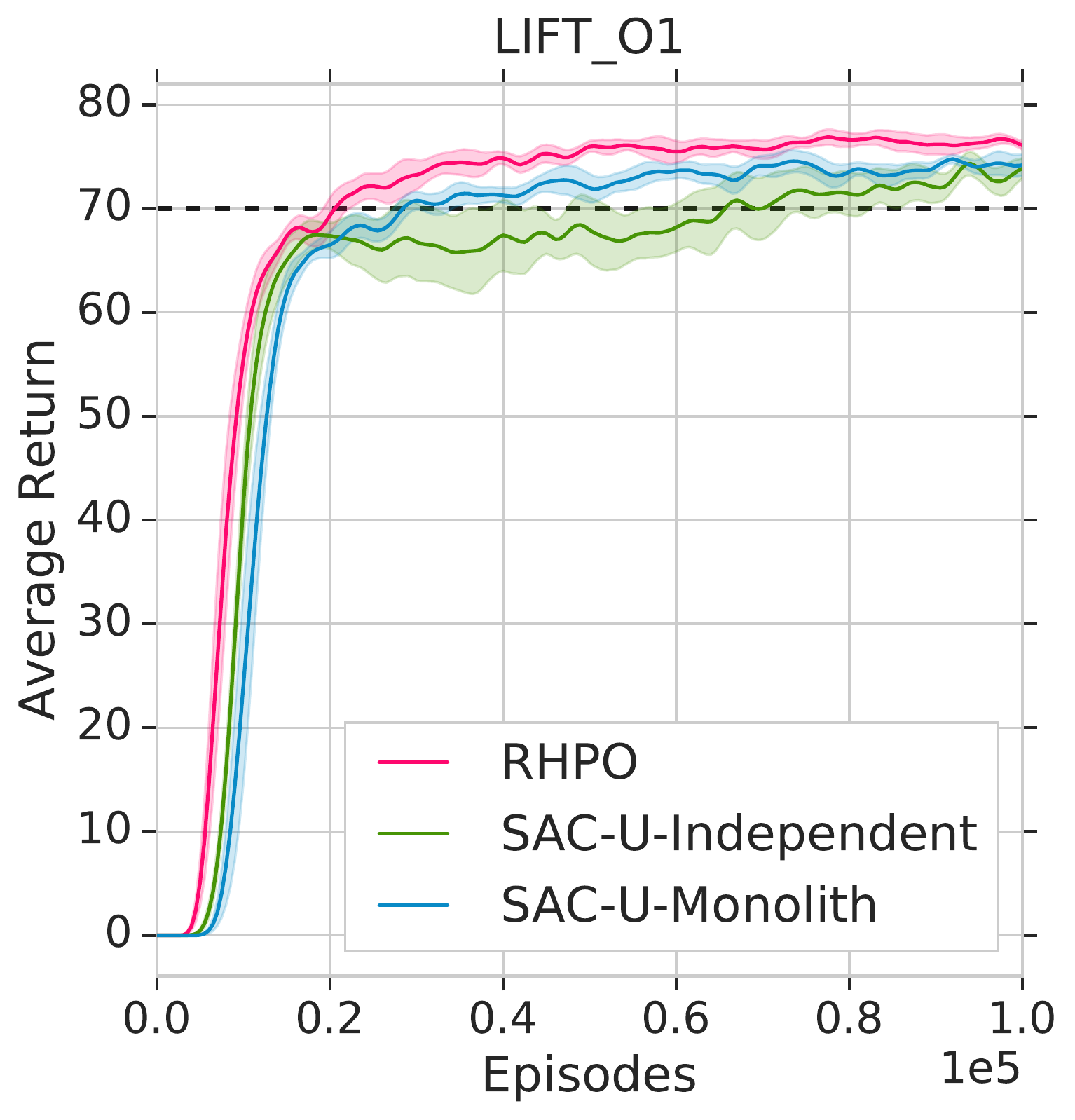} &
    \includegraphics[trim=30 0 0 0,width=.22\textwidth]{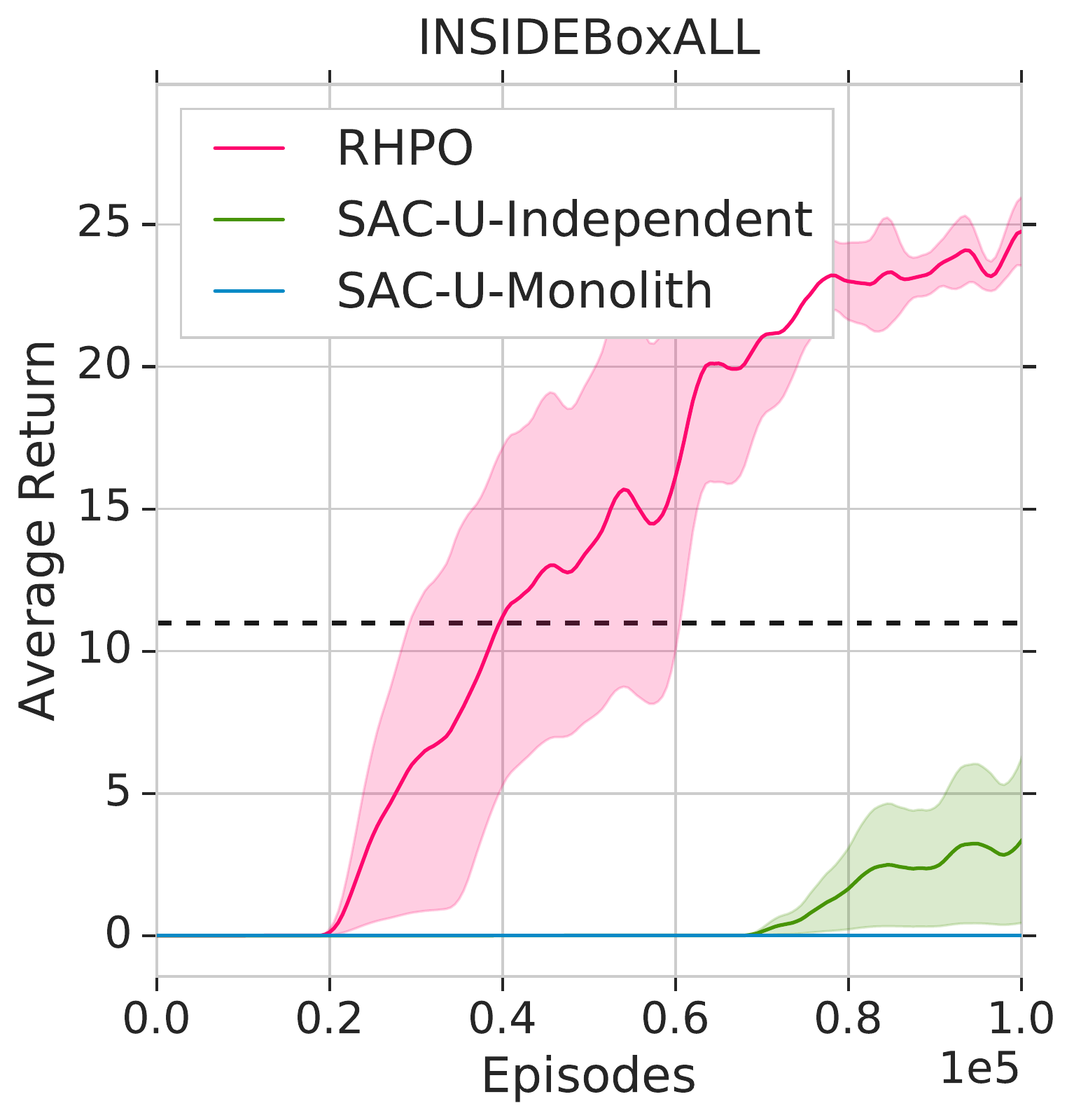} \\
    \end{tabular}
    \caption{\small Results for the multitask robotic manipulation experiments in simulation. The dashed line corresponds to the performance of the SVG-based implementation of SAC-U. From top to bottom: 2 tasks from the Pile1, Pile2 \& Cleanup2 domains. We show averages over 3 runs each, with corresponding standard deviation. \met~outperforms both baselines across all tasks with the benefits increasing for more complex domains. }
    \label{fig:multitask_experiments}
\end{figure}

Figure \ref{fig:multitask_experiments} demonstrates that \met~outperforms the monolithic as well as the independent baselines (based on SAC). For simple tasks such as the Pile1 domain, the difference is small, but as the number of tasks grows and the complexity of the domain increases (cf. Pile2 and Cleanup2), the advantage of composing learned behaviours across tasks becomes more significant. 
We further observe that using MPO instead of SVG \citep{heess2015learning} as policy optimizer results in an improvement for the baselines. This effect becomes more pronounced for the hierarchical policies.

\vspace{-3mm}

\subsection{Physical Robot Experiments}\label{sec:real_multitask}
For real-world experiments, data-efficiency is crucial. We perform all experiments in this section relying on a single robot (single actor) -- demonstrating the benefits of \met~in the low data regime. The performed task is the real world version of the Pile1 task described in Section \ref{sec:sim_multitask}. Given the higher cost of experiment time, the robot experiments additionally emphasize the requirements for hyperparameter robust algorithms which is further investigated in Section \ref{sec:pablations}.

\begin{figure}[t]
    \centering
    \begin{tabular}{cc}
    \includegraphics[trim=30 0 0 0,width=0.23\textwidth]{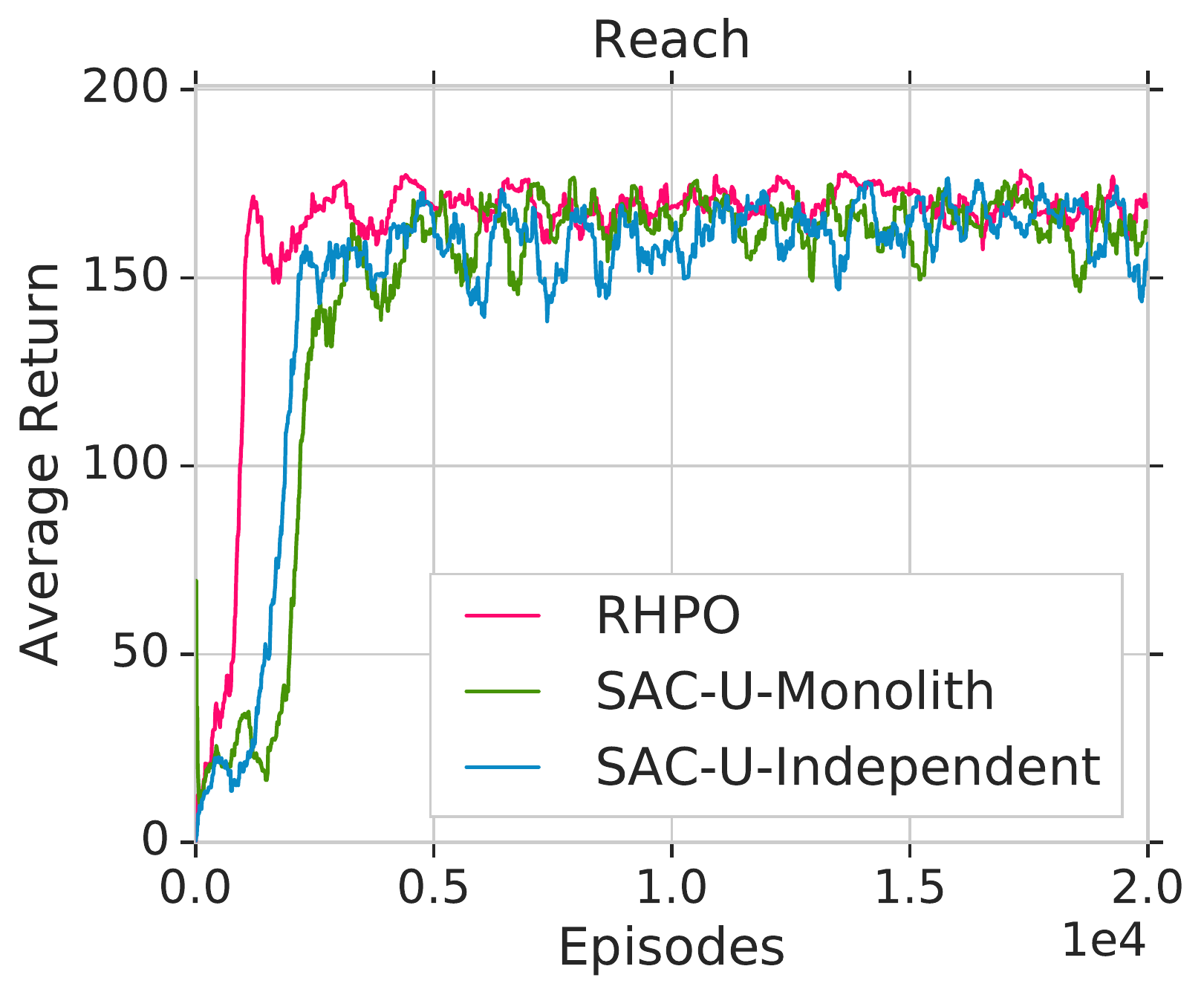}  &
    \includegraphics[trim=30 0 0 0,width=0.23\textwidth]{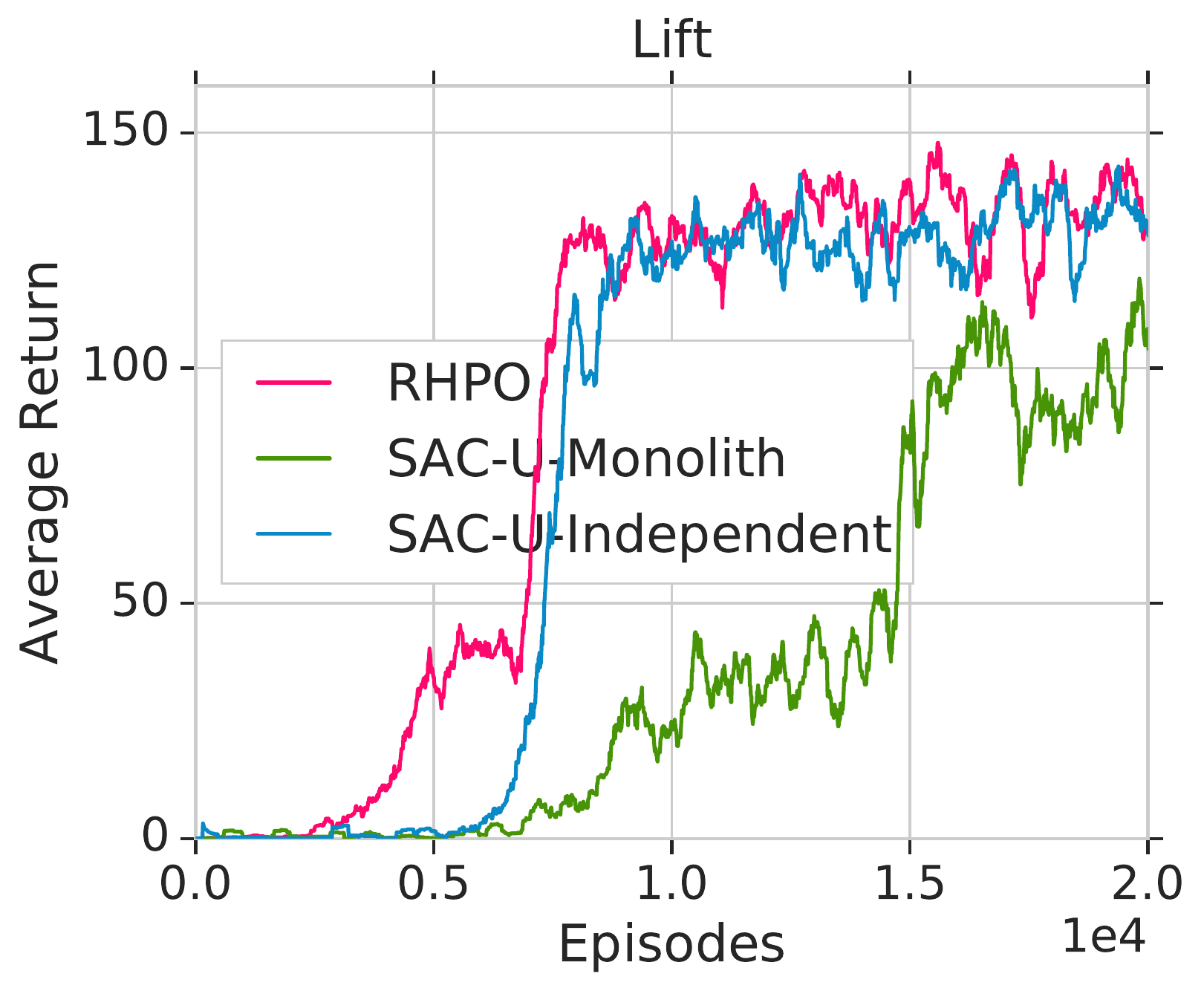}
    \\
    \includegraphics[trim=30 0 0 0,width=0.23\textwidth]{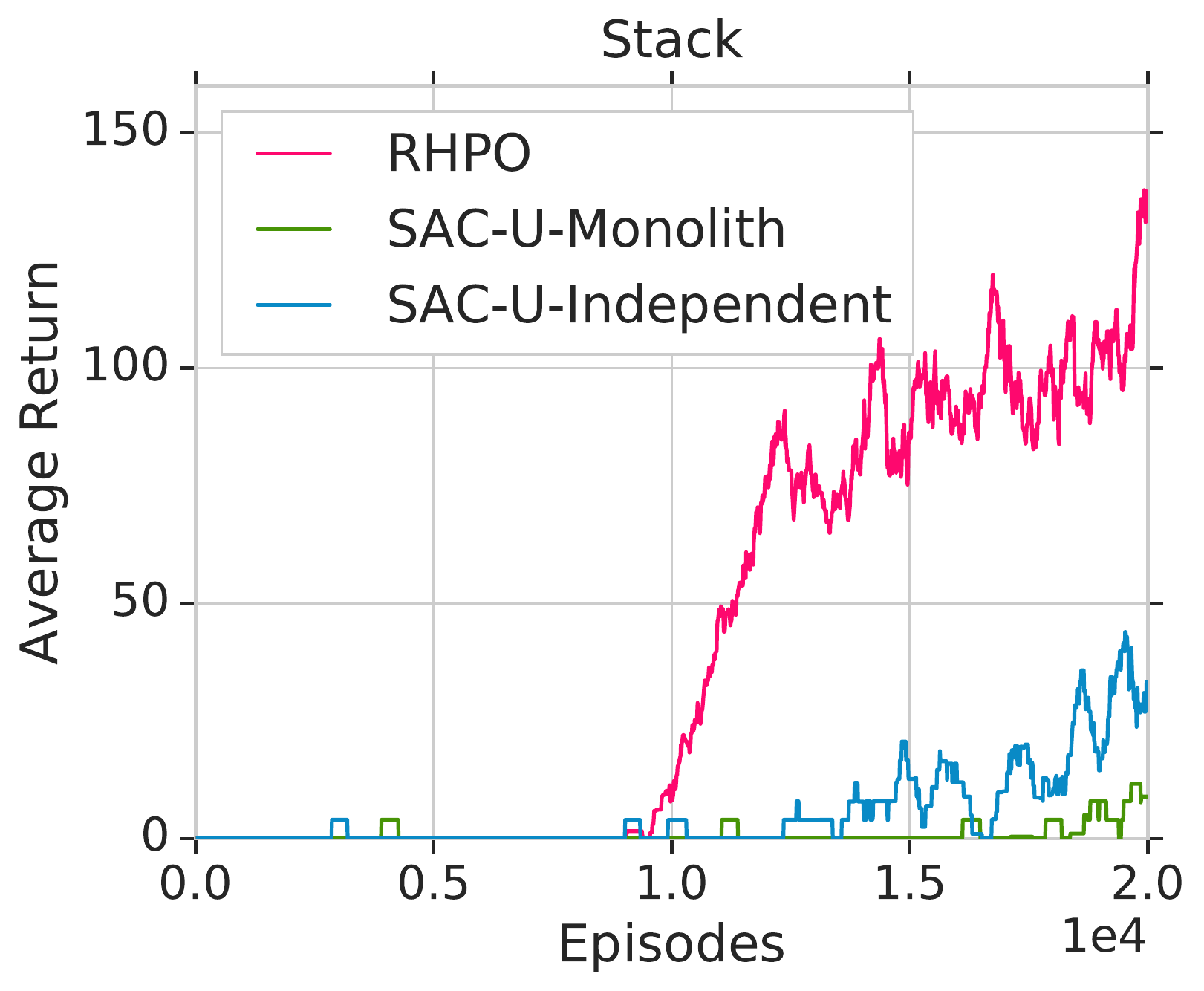} &
    \includegraphics[trim=30 0 0 0,width=0.23\textwidth]{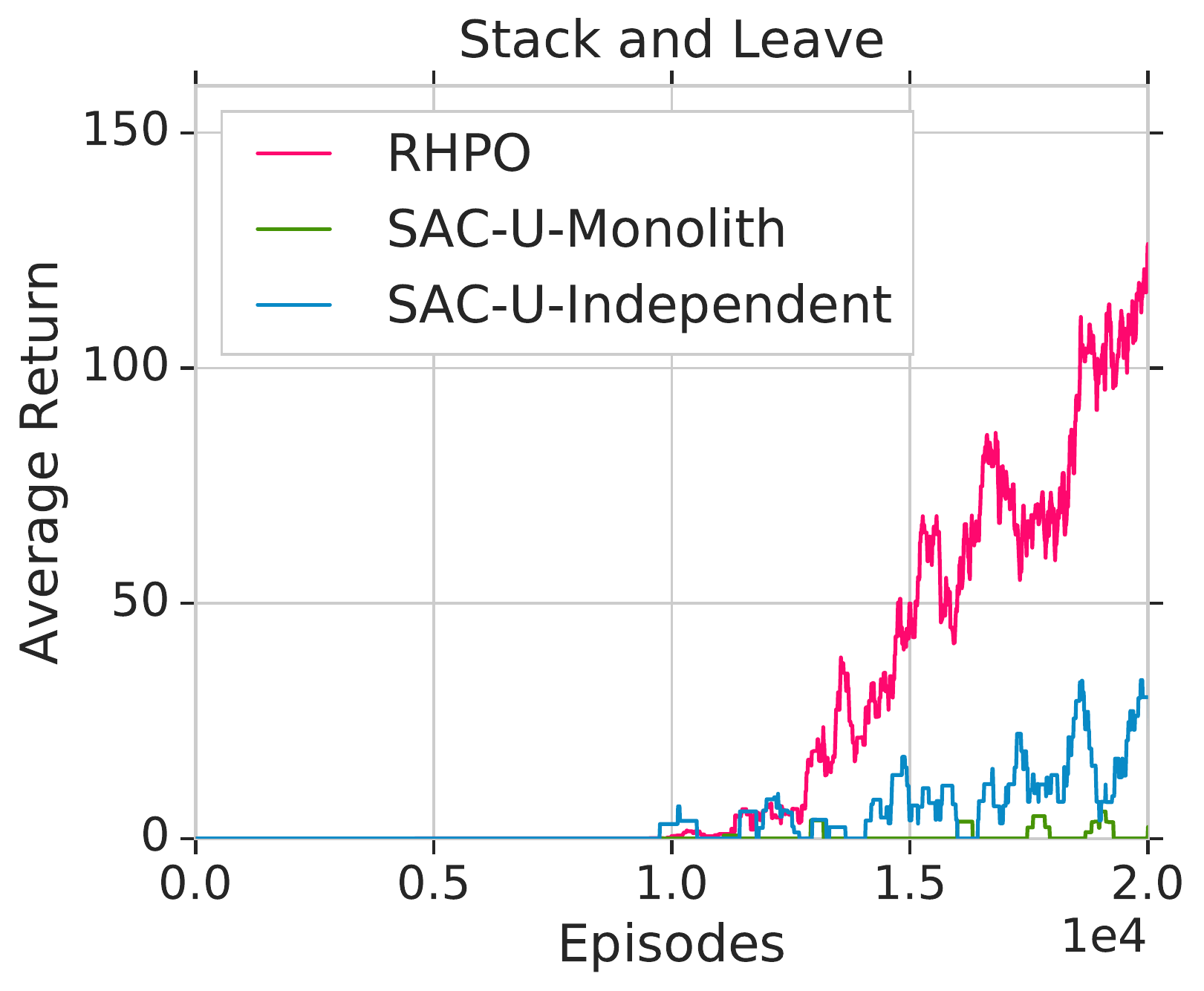} 
    \end{tabular}

    \caption{\small Robot Experiments. Left: While simpler tasks such as reaching are learned with comparable efficiency, the later, more complex tasks are acquired significantly faster with \met. }
    \label{fig:robot_experiments}
\end{figure}

The setup for the experiments consists of a Sawyer robot arm mounted on a table, equipped with a Robotiq 2F-85 parallel gripper. A basket of size $20\text{cm}^2$ in front of the robot contains the three cubes. Three cameras on the basket track the cubes using fiducials (augmented reality tags).
As in simulation, the agent is provided with proprioception information (joint positions, velocities and torques), a wrist sensor's force and torque readings, as well as the cubes' poses -- estimated via the fiducials. The agent's action is five dimensional and consists of the three Cartesian translational velocities, the angular velocity of the wrist around the vertical axis and the speed of the gripper's fingers.

Figure \ref{fig:robot_experiments} plots the learning progress on the real robot for four of the tasks, from simpler reach and lift tasks and the stack and final stack-and-leave task -- which is the main task of interest. Plots for the learning progress of all tasks are given in the appendix \ref{sec:completeResults}. As can be observed, all methods manage to learn the reach task quickly (within about a few thousand episodes) but only \met~with a hierarchical policy is able to learn the stacking task (taking about 15 thousand episodes to obtain good stacking success), which takes about 8 days of training on the real robot with considerably slower progress for all baselines taking multiple weeks for completion.

\begin{figure}[h]
    \centering
    \begin{tabular}{cc}
    \includegraphics[scale=0.3]{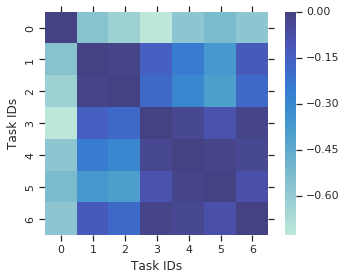} &
    % \caption*{Task \\Similarity}
    % \end{subfigure}%\hspace*{-0.9em}
    % \begin{subfigure}{0.2\textwidth}
    \includegraphics[scale=0.3]{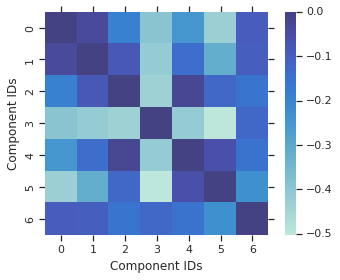} \\
    Task Similarity & Component Similarity
    
        % \caption*{Component \\Similarity}
    % \end{subfigure}
    \end{tabular}

    \caption{\small Similarities between tasks (based on their distribution over components) and similarities between components (based on the distribution over tasks which apply them). }
    \label{fig:similarity}
\end{figure}

To provide further insight into the learned representation we compute distributions for each component over the tasks which activate it, as well as distributions for each task over which components are being used. For each set of distributions, we determine the Battacharyya distance metric to determine the similarity between tasks and the similarity between components in Figure \ref{fig:similarity} (right). The plots demonstrate how the components specialize, but also provide a way to investigate our tasks, showing e.g. that the first reach task is fairly independent and that the last four tasks are comparably similar regarding the high-level components applied for their solution.

\vspace{-2mm}

\subsection{Sequential Transfer}\label{sec:sequential}
\vspace{-2mm}
\begin{figure}[h]
    \centering
    \begin{tabular}{cc}
    \includegraphics[trim=40 0 0 0, width=.23\textwidth]{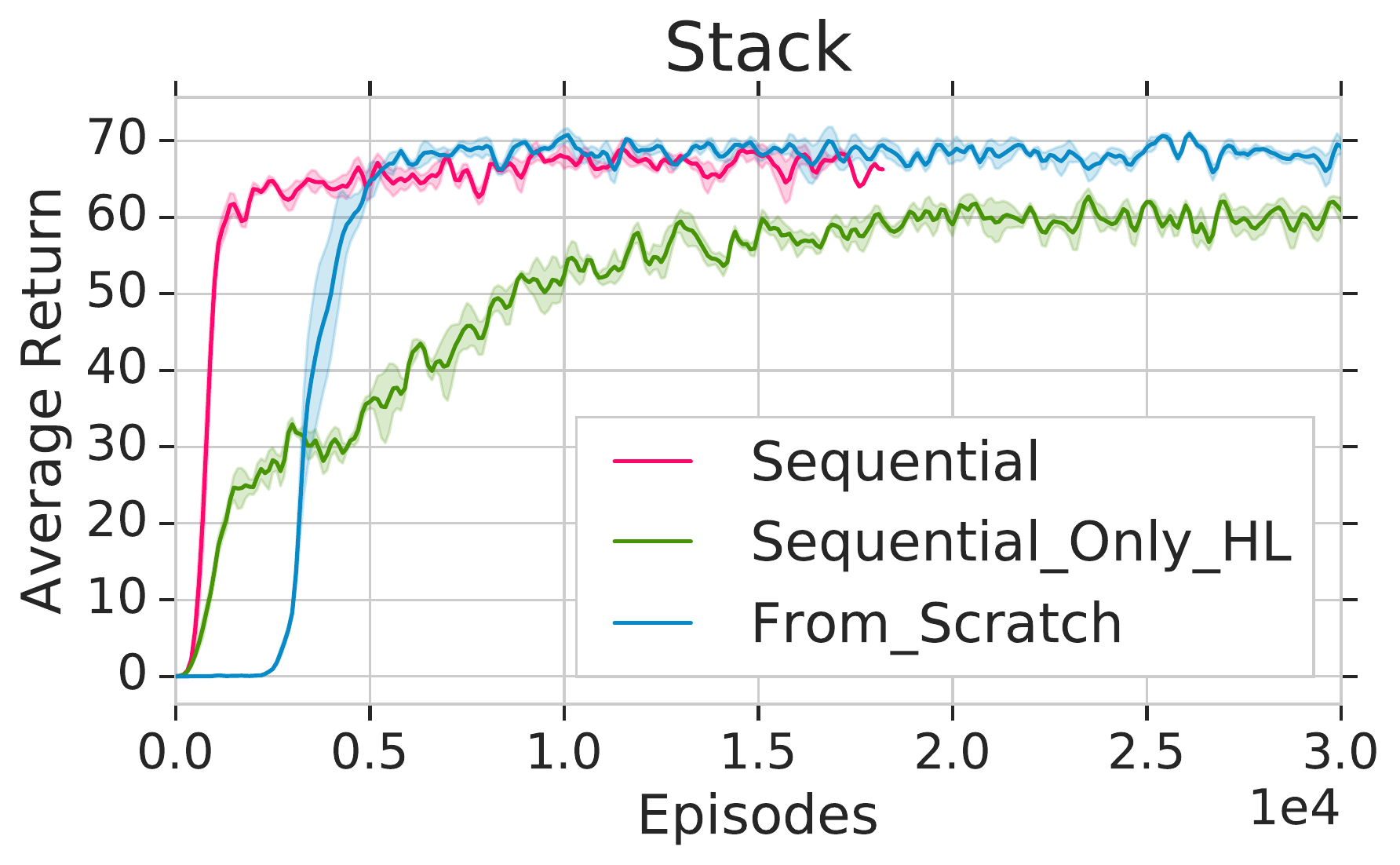}&
    \includegraphics[trim=40 0 0 0, width=.23\textwidth]{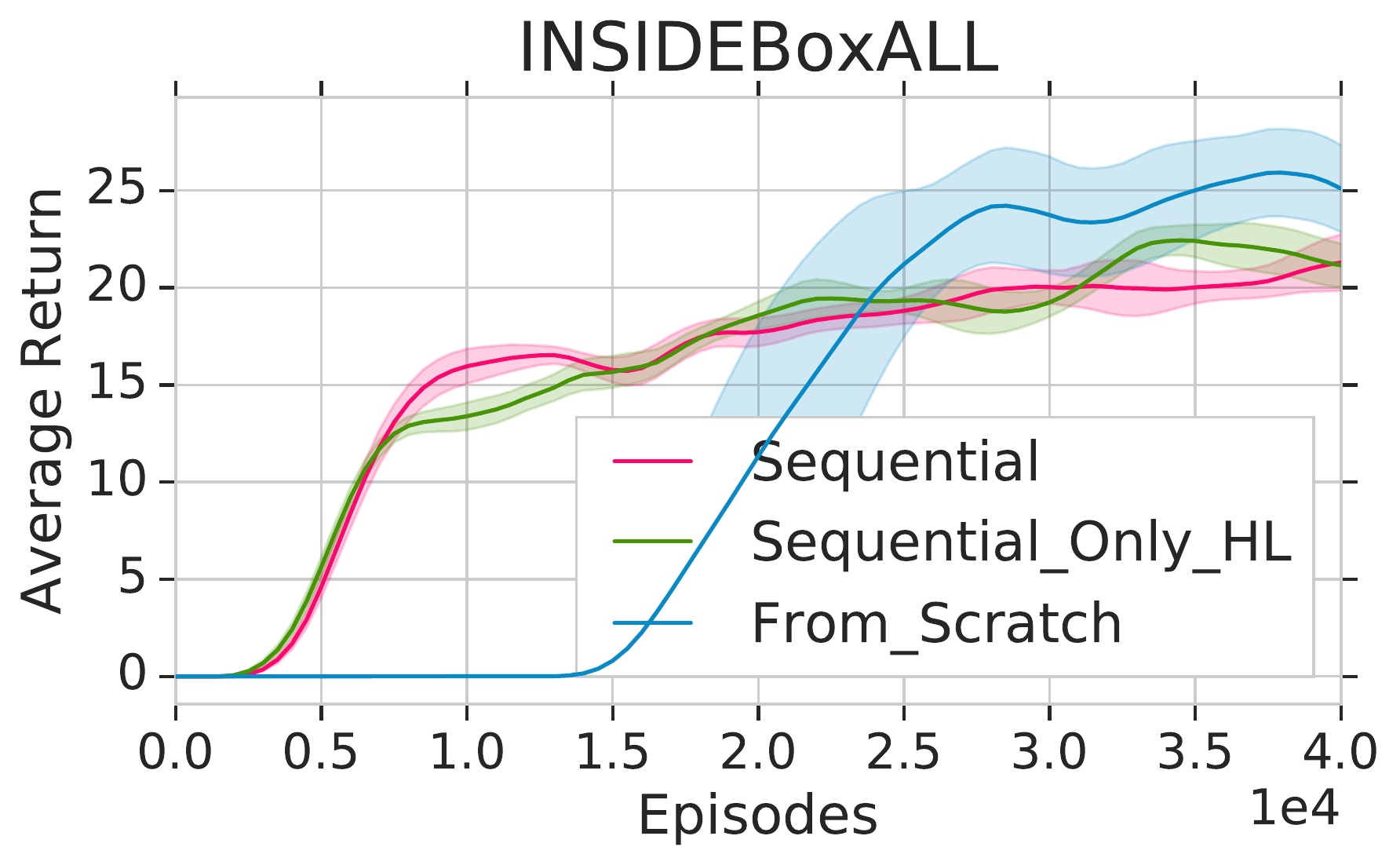}\\
    % Stack % Stack
    \end{tabular}
    \caption{\small Sequential transfer experiments: the models are first trained with all but the final task in the Pile1 and Cleanup2 domains, and finally we train the models to adapt to the final task by either training 1- only a high-level controller or 2-a high-level controller as well as an additional component. 
    }
    \label{fig:sequential_experiments}
\end{figure}

\met~is well suited for sequential transfer learning as it allows to use pre-trained low-level components to solve new tasks.  
%based on solving new tasks with pre-trained low-level components. 
To investigate performance in adapting pre-trained multitask policies to novel tasks, we train agents to fulfill all but the final task in the Pile1 and Cleanup2 domains and subsequently evaluate training the models on the final task. 
We consider two settings for the final policy: we either introduce only a new high-level controller (Sequential-Only-HL) or both an additional shared component as well as a new high-level controller (Sequential).
Figure \ref{fig:sequential_experiments} displays that in the sequential transfer setting, starting from a policy trained on a set of related tasks results in up to 5 times more data-efficiency in terms of actor episodes on the final task than training the same policy from scratch. We observe that the final task can be solved by only reusing low-level components from previous tasks if the final task is the composition of previous tasks. This is the case for the final task in Cleanup2 which can be completed by sequencing the previously learned components and in contrast to Pile1 where the final letting go of the block after stacking is not required for earlier tasks.

\subsection{Simulated Single Task Experiments}\label{sec:sim_singletask}

\begin{figure}[h]
    \centering
    \begin{tabular}{c}
    \includegraphics[width=.48\textwidth]{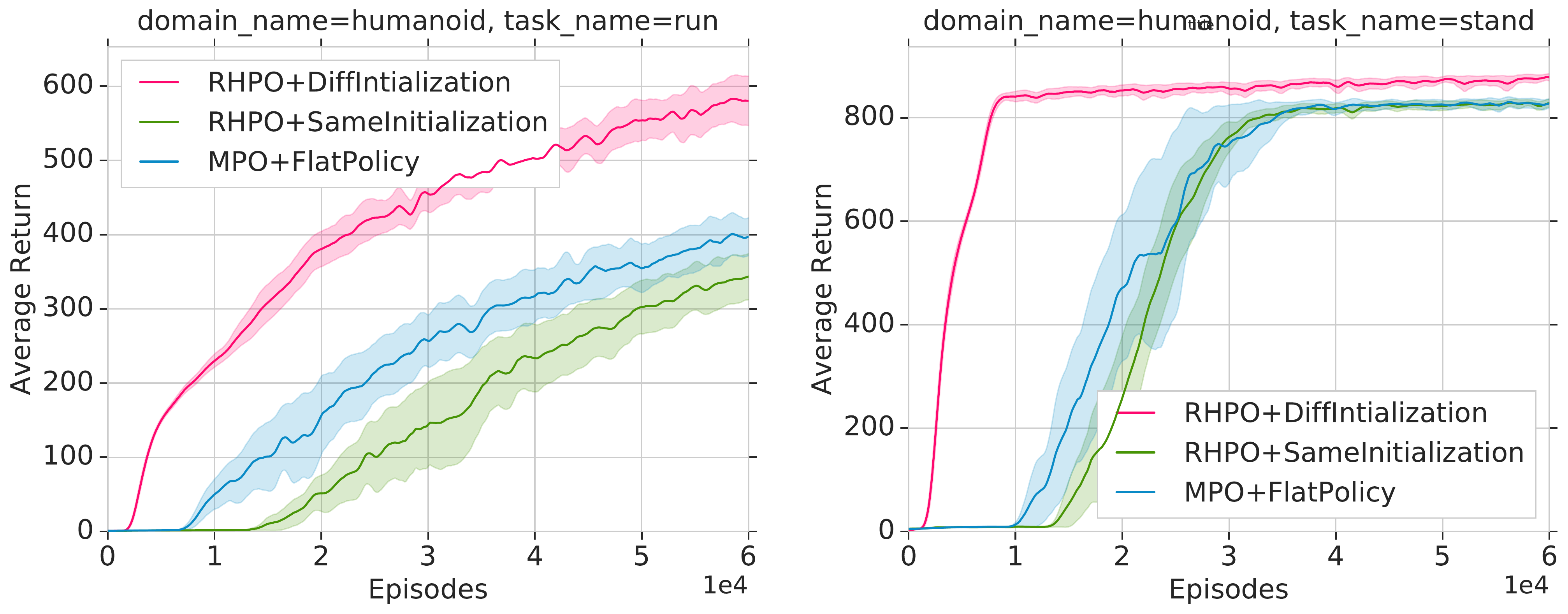}
    \end{tabular}
    \caption{\small Using \met~with different component initialization (red curve) demonstrates benefits over homogeneous initialization as well as the flat Gaussian policy. The plot shows that the simple change in initialization is sufficient to enable component specialization and the correlated improvement in performance.}
    \label{fig:singleTask_experiments}
\end{figure}

We consider two high-dimensional tasks for continuous control: humanoid-run and humanoid-stand from \citet{tassa2018deepmind} and  compare MPO with a flat Gaussian policy to \met~with a mixture of Gaussians with three components. 
% In the presented setup, learner and actors work asynchronously. 
Figure \ref{fig:singleTask_experiments} shows the results in terms of the number of episodes. 

When both the flat and hierarchical policies are initialized with means close to zero, \met~performs comparable to a flat policy and learns similar means and variances for all components as the model fails to decompose the learned behavior. If, however, the hierarchical policy is initialized with distinct means for different components (here, for the three components ranging for all dimensions from the minimum to maximum of the allowed action range, i.e.\ -1, 1), we observe significantly improved performance and component specialization.

\subsection{Performance Ablations}\label{sec:pablations}
We perform a series of ablations based on the earlier introduced Pile1 domain, providing additional insights into benefits, shortcomings and relevant hyperparameters of \met.

First, we display the importance of choice of regularization in Figure \ref{fig:catkl_experiments1} with complete results in Appendix \ref{app:regularisation}. We are able to demonstrate the effect of weakening the constraint by setting the epsilon value higher (here: to 1.). This setting prevents convergence of the policy to capable solutions and emphasizes the necessity of constraining the update steps. In addition, very small values can slow down convergence. However, in the present experiments a range of about 2 orders of magnitude results in good performance.

\begin{figure}[h]
    \centering
    \begin{tabular}{cc}
    \includegraphics[trim=50 0 0 0,width=.23\textwidth]{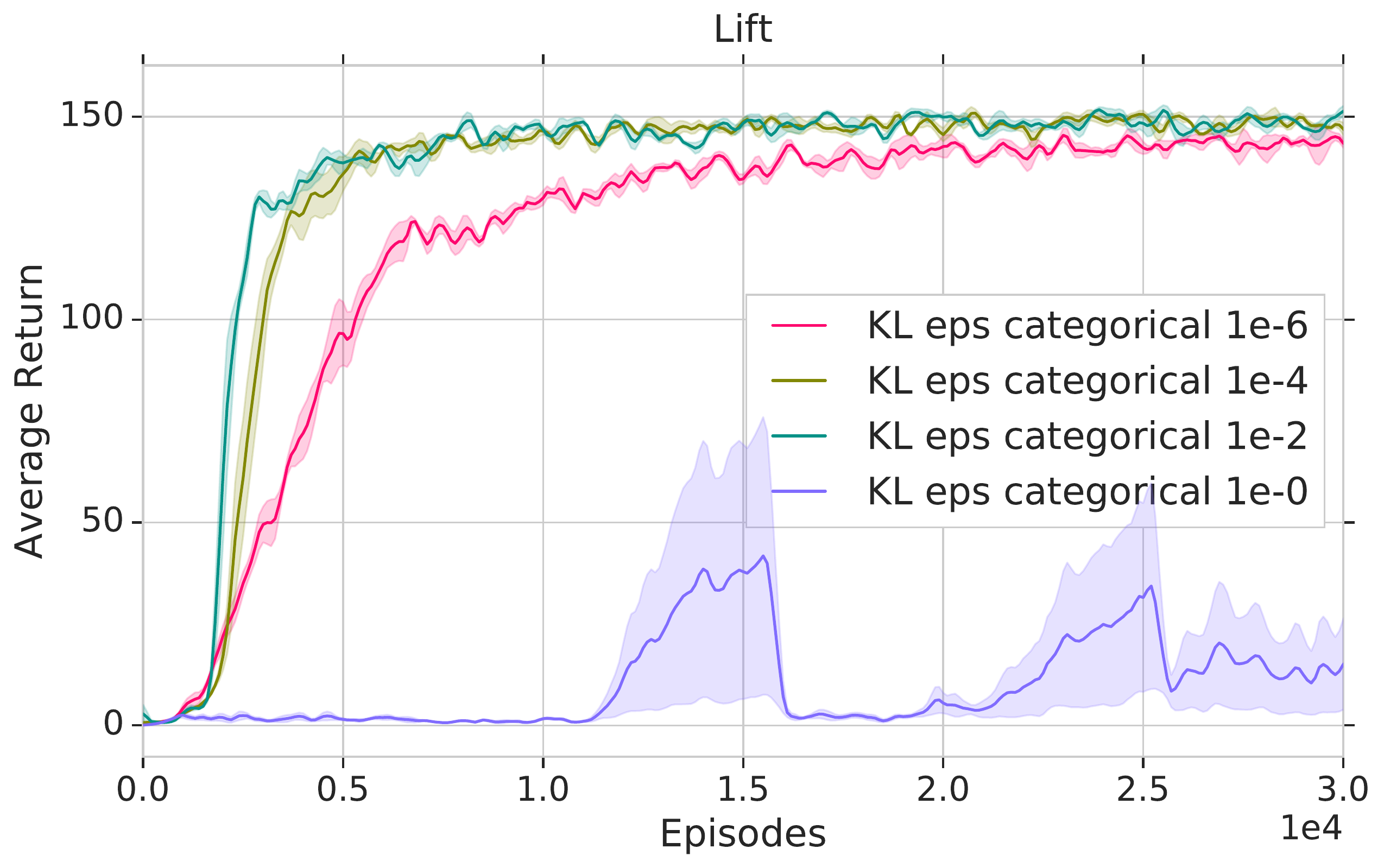} &
    \includegraphics[trim=50 0 0 0,width=.23\textwidth]{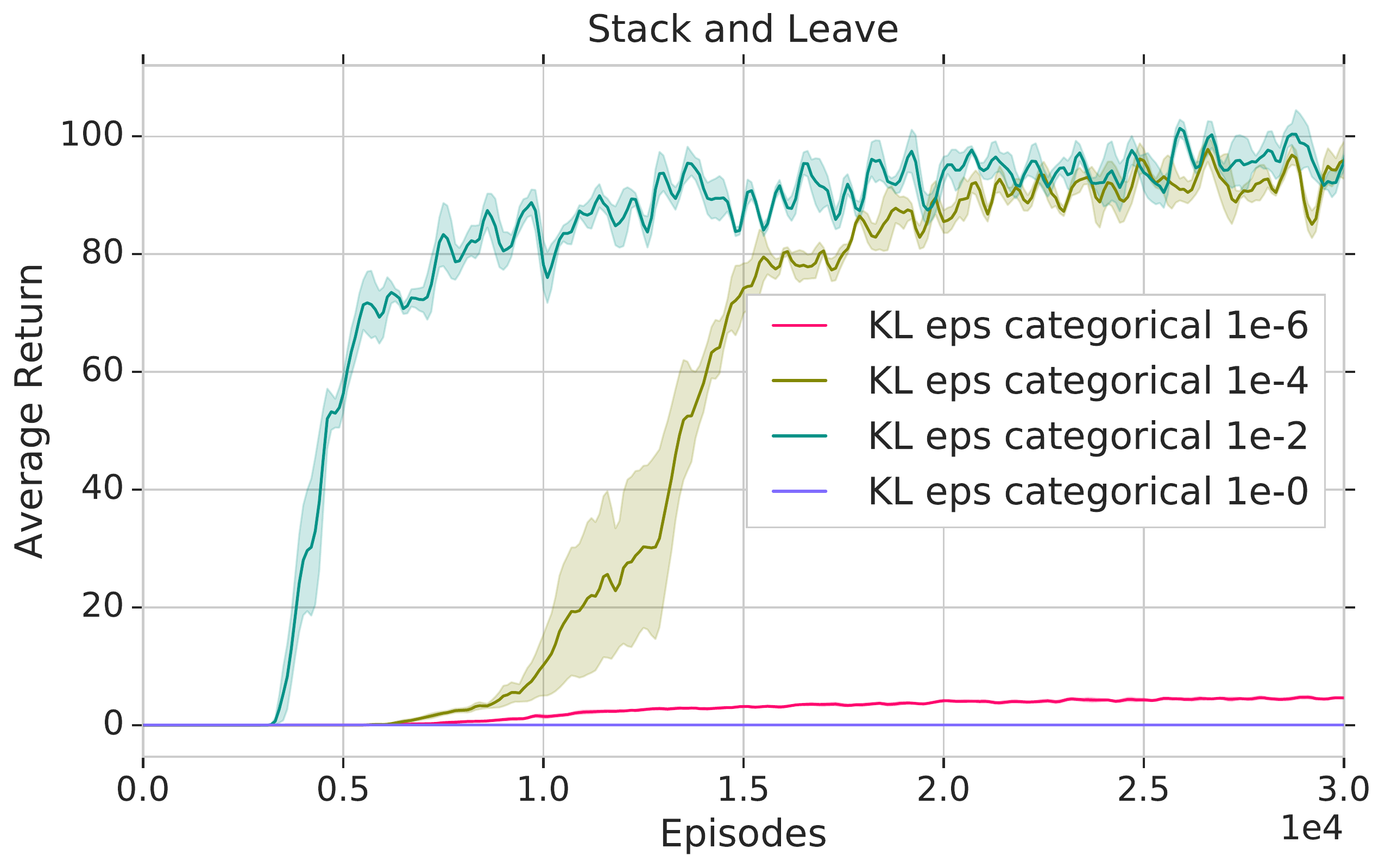} 
    \end{tabular}
    \caption{\small Results for sweeping the KL constraint between 1e-6 and 1. for 2 tasks in the Pile1 domain. For very weak constraints the model does not converge successfully, while for very strong constraints it only converges very slowly. }
    \label{fig:catkl_experiments1}
\end{figure}

\begin{figure}[h]
    \centering
    \begin{tabular}{c}
    \includegraphics[width=.45\textwidth]{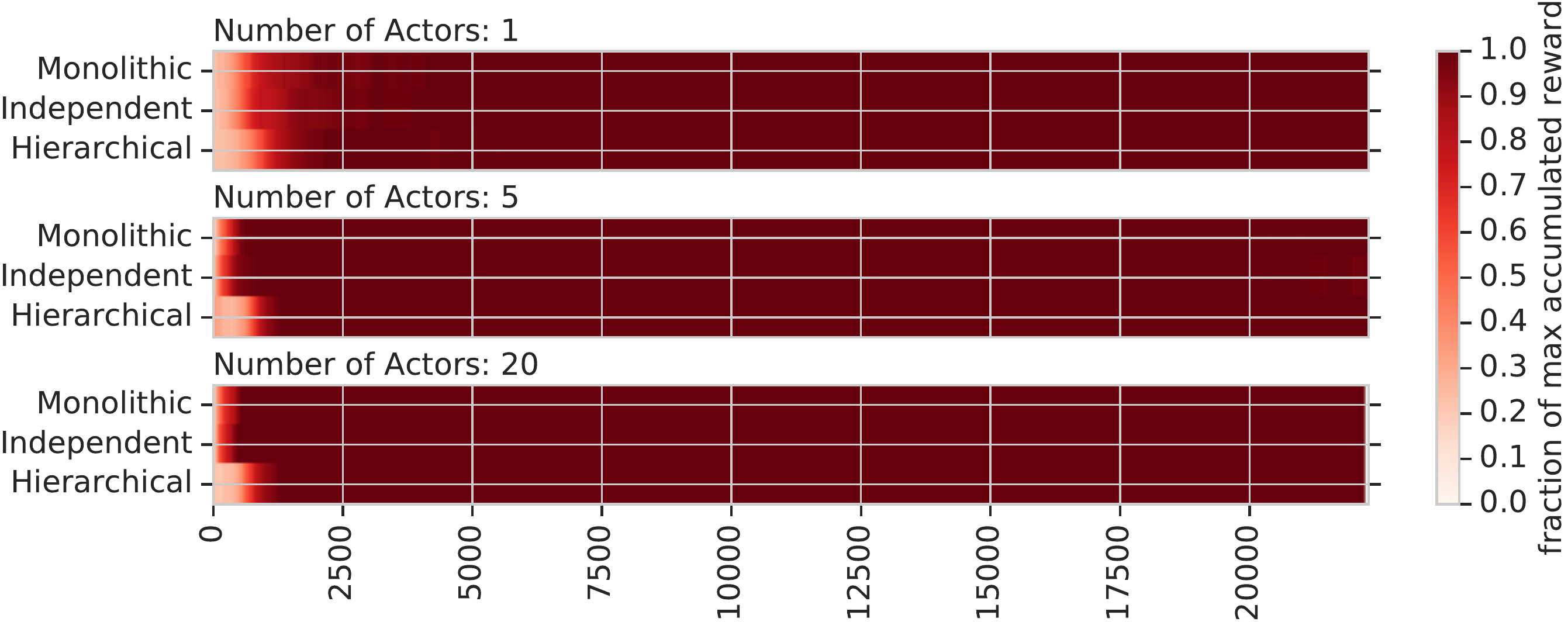}\\
    \small Reach\\
    % \includegraphics[width=.7\textwidth]{figures/actor_ablations/actor_ablation_barplot_1_5_20_1.pdf}\\
    % Grasp\\
    \includegraphics[width=.45\textwidth]{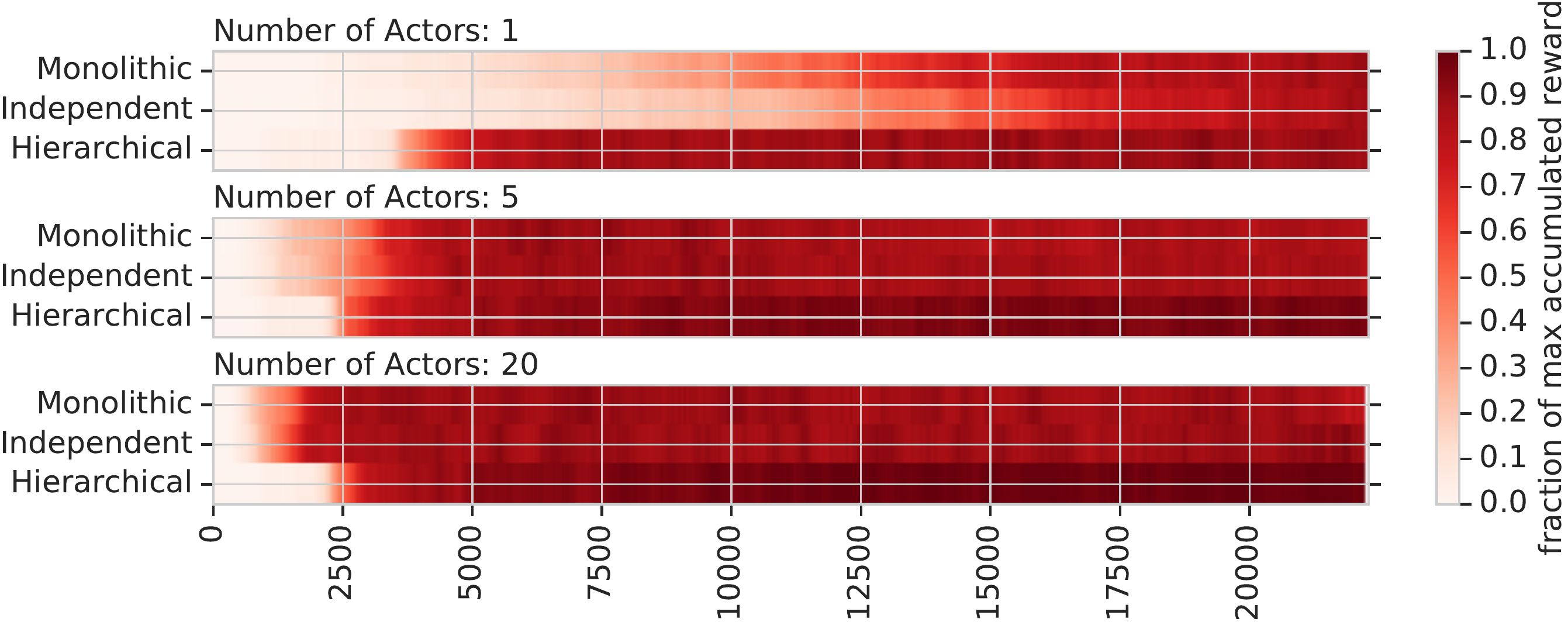}\\
    \small Lift\\
    % \includegraphics[width=.7\textwidth]{figures/actor_ablations/actor_ablation_barplot_1_5_20_3.pdf}\\ 
    % Place Wide\\
    % \includegraphics[width=.7\textwidth]{figures/actor_ablations/actor_ablation_barplot_1_5_20_4.pdf}\\
    % Place Narrow\\
    % \includegraphics[width=.7\textwidth]{figures/actor_ablations/actor_ablation_barplot_1_5_20_4.pdf}\\
    % Place Narrow\\
    % \includegraphics[width=.7\textwidth]{figures/actor_ablations/actor_ablation_barplot_1_5_20_5.pdf} \\
    \includegraphics[width=.45\textwidth]{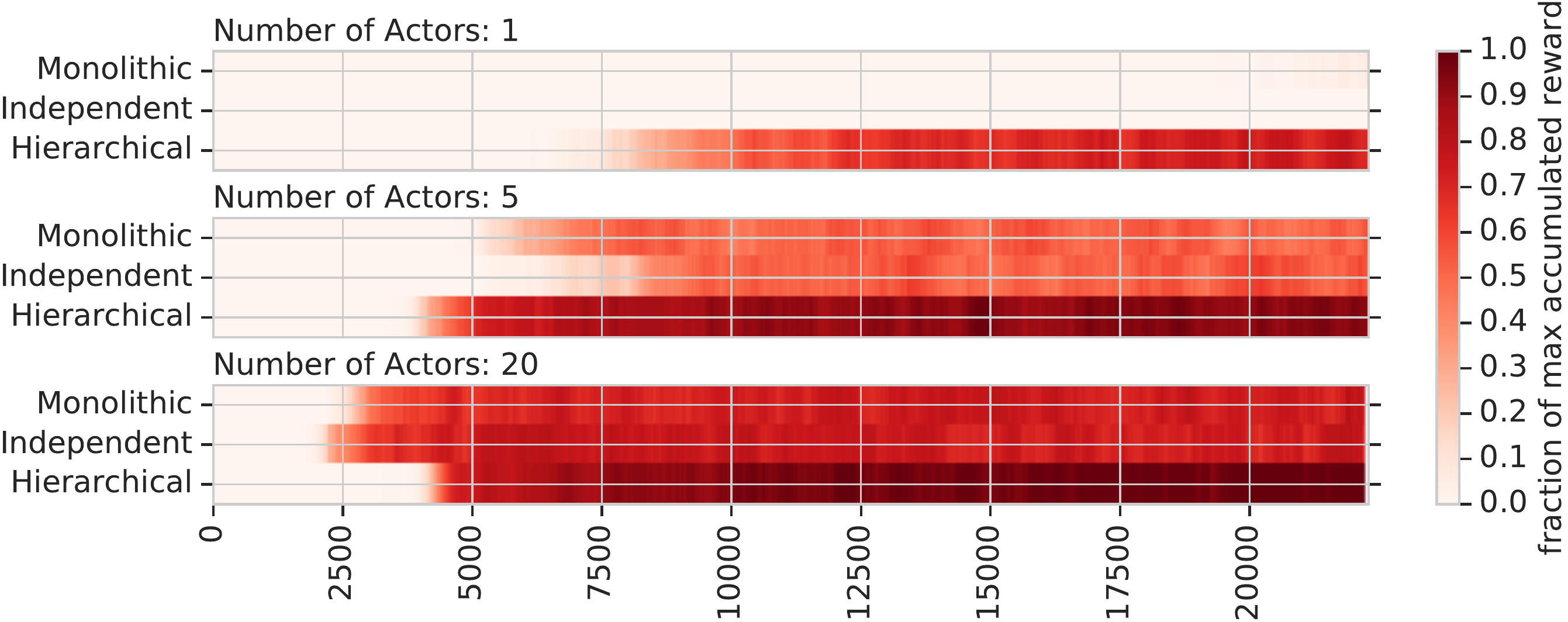} \\
    \small Stack\\
    % \includegraphics[width=.7\textwidth]{figures/actor_ablations/actor_ablation_barplot_1_5_20_6.pdf} \\
    % Stack and Leave
    \end{tabular}
    \caption{\small Results for ablating the number of data-generating actors in the Pile1 domain. We can see that the benefit of hierarchical policies is stronger for more complex tasks and lower data rates. However, even with 20 actors we see better final performance and stability}
    \label{fig:multitask_experiments_datarate1}
\end{figure}

We additionally ablate over the number of data-generating actors in Figure \ref{fig:multitask_experiments_datarate1} to evaluate all approaches with respect to data rate and illustrate how \met~is particularly relevant at lower data rates such as given by real-world robotics applications (with results for all tasks in Appendix \ref{app:datarate}). Here, \met~always provides stronger final performance and learns significantly faster in one actor case as common for robot experiments. By running with multiple actors, we increase the rate of data generation such that in asynchronous settings, the speed of the learner becomes more important. Since training our hierarchical policies is computationally slightly more costly, the benefits become smaller for easier tasks\footnote{In asynchronous RL systems, the update rate of the learner can have a significant impact on the performance when evaluated over data generated. }.

Finally, we demonstrate the robustness of \met~with respect to the number of sub-policies in Figure \ref{fig:multitask_experiments_datarate1} (with complete results in Appendix \ref{app:components}) and connected simplicity of determining this hyperparameter which for all other experiments is simply set proportionally to the number of tasks.

\section{Related Work}\label{sec:related}

Transfer learning, in particular in the multitask context, has long been part of machine learning (ML) for data-limited domains \citep{caruana1997multitask,torrey2010transfer,pan2010survey,taylor2009transfer}. Commonly, it is not straightforward to train a single model jointly across different tasks as the solutions to tasks might not only interfere positively but also negatively \citep{wang2018characterizing}. 
Preventing this type of forgetting or negative transfer presents a challenge for biological \citep{singley1989transfer} as well as artificial systems \citep{french1999catastrophic}. In the context of ML, a common scheme is the reduction of representational overlap  \citep{french1999catastrophic,rusu2016progressive,wang2018characterizing}.
\citet{bishop1994mixture} utilize neural networks to parametrize mixture models for representing multi-modal distributions thus mitigating shortcomings of non-hierarchical approaches. \citet{rosenstein2005transfer} demonstrate the benefits of hierarchical classification models to limit the impact of negative transfer.

Hierarchical approaches have a long history in the reinforcement learning literature \citep[e.g.][]{sutton1999between,dayan1993feudal}.
Prior work commonly benefits from combining hierarchy with additional inductive biases such as \citep{vezhnevets2017feudal, nachum2018data, nachum2018near,xie2018transferring} which employ different rewards for different levels of the hierarchy rather than optimizing a single objective for the entire model as we do. 
Other works have shown the additional benefits for the stability of training and data-efficiency when sequences of high-level actions are given as guidance during optimization in a hierarchical setting \citep{shiarlis2018taco, andreas2017modular,tirumala2019exploiting}.
Instead of introducing additional training signals, we directly investigate the benefits of compositional hierarchy as provided structure for transfer between tasks.

Hierarchical models for probabilistic trajectory modelling have been used for the discovery of behavior abstractions as part of an end-to-end reinforcement learning paradigm \citep[e.g.][]{teh2017distral,igl2019multitask,tirumala2019exploiting,galashov2018information} where the models act as learned inductive biases that induce the sharing of behavior across tasks. 
In a vein similar to the presented algorithm, \citep[e.g][]{heess2016learning,tirumala2019exploiting} share a low-level controller across tasks but modulate the low-level behavior via a continuous embedding rather than picking from a small number of mixture components.
In related work  \cite{hausman2018learning,haarnoja2018latent} learn hierarchical policies with continuous latent variables optimizing the entropy regularized objective.

Similar to our work, the options framework \citep{sutton1999between,precup2000temporal} supports behavior hierarchies, where the higher level chooses from a discrete set of sub-policies or ``options'' which commonly are run until a termination criterion is satisfied. The framework focuses on the notion of temporal abstraction. A number of works have proposed practical and scalable algorithms for learning option policies with reinforcement learning \citep[e.g.][]{bacon2017option,zhang2019dac,smith2018inference,riemer2018learning,harb2018waiting} or criteria for option induction \citep[e.g.][]{harb2018waiting,harut2019termination}. 
Rather than the additional inductive bias of temporal abstraction, we focus on the investigation of composition as type of hierarchy in the context of single and multitask learning while demonstrating the strength of hierarchical composition to lie in domains with strong variation in the objectives - such as in multitask domains.
We additionally introduce a hierarchical extension of SVG \citep{heess2015learning}, to investigate similarities to work on the option critic \citep{bacon2017option}. 

With the use of KL regularization to different ends in RL, work related to \met~focuses on contextual bandits \citep{daniel2016hierarchical}. 
The algorithm builds on a 2-step EM like procedure to optimize linearly parametrized mixture policies. However, their algorithm has been used only with low dimensional policy representations, and in contextual bandit and other very short horizon settings.
Our approach is designed to be applicable to full RL problems in complex domains with long horizons and with high-capacity function approximators such as neural networks. This requires robust estimation of value function approximations, off-policy correction, and additional regularization for stable learning.

\section{Discussion}\label{sec:discussions}

We introduce \met, a novel algorithm for robust training of hierarchical policies in multitask settings. \met~ consistently outperforms competitive baselines which either handle tasks independently or implicitly share experience by reusing data across tasks. Especially for complex tasks or in a low data regime, as encountered in robotics applications, we strongly reduce the number of environment interactions and improve final performance as well as learning robustness and sensitivity to hyper-parameters. Our results show that the algorithm scales to complex, real-world domains and provides an important step towards the deployment of RL algorithms on robotic systems.

Algorithmically, our method highlights the importance of trust-region-like regularization for stable optimization of hierarchical policies. Furthermore, our update rules in combination with mixture policies and hindsight reward assignments enable training for any task and skill independent of the data source. This enables efficient learning of the hierarchical policies in an off-policy setting, which is important for data efficient learning.   

Conceptually, our results demonstrate that hierarchical policies can be an effective way of sharing skills or behavior components across tasks, both in multitask (Sections \ref{sec:sim_multitask}-\ref{sec:real_multitask}) as well as in transfer settings (Section \ref{sec:sequential}) and partially mitigate negative interference between tasks in the parallel multitask learning scenario. 
Furthermore, we find that their benefits are complementary to off-policy sharing of transition data across tasks (e.g. SAC-X \citep{riedmiller2018learning}, HER \citep{andrychowicz2017hindsight}).
Valuable directions for future work include the direct extension to multilevel hierarchies and the identification of basis sets of behaviours which perform well on wide ranges of possible tasks given a known domain.

We believe that especially in domains with consistent agent embodiment and high costs for data generation learning tasks jointly and information sharing is imperative. Our results suggest that a system that is exposed to a rich set of tasks or experiences and has appropriate means for reusing knowledge can learn to solve non-trivial problems directly from interaction with its environment. \met~combines several ideas that we believe will be important: sharing data across tasks and skills across tasks with compositional policy representations, robust optimization, and efficient off-policy learning. Although we have found this particular combination of components to be very effective we believe it is just one instance of -- and step towards -- a spectrum of efficient learning architectures that will unlock further applications of RL both in simulation and, more importantly, on physical hardware.

\section*{Acknowledgments}
The authors would like to thank Michael Bloesch, Jonas Degrave, Joseph Modayil and Doina Precup for helpful discussion and relevant feedback for shaping our submission. 
As robotics (and AI research) is a team sport we'd additionally like to acknowledge the support of Francesco Romano, Murilo Martins, Stefano Salicetti, Tom Roth{\"o}rl and Francesco Nori on the hardware and infrastructure side as well as many others of the DeepMind team.

\bibliography{main}
\bibliographystyle{plainnat}

\clearpage
\appendix

\section{Appendix}
\subsection{Additional Derivations \label{sec:additiona_derivations}}

In this section we explain the detailed derivations for training hierarchical policies parameterized as a mixture of Gaussians. 
\subsubsection{Obtaining Non-parametric Policies \label{sec:dualfunctionderivation}}
In each policy improvement step, to obtain non-parametric policies for a given state and task distribution, we solve the following program:

\begin{equation*}
  \begin{aligned}
  & \max_q \mathbb{E}_{\mu(s), i \sim I} \big[ \mathbb{E}_{q(a|s,i)} \big[\hat{Q}(s,a,i)] \big] \\  
  & s.t. \mathbb{E}_{\mu(s), i \sim I} \big[ \textrm{KL}(q(\cdot|s,i) , \pi(\cdot|s,i,\theta_t) ) \big] < \epsilon\\
  & s.t. \mathbb{E}_{\mu(s), i \sim I} \big[ \mathbb{E}_{q(a | s)} \big [1 \big] \big] = 1.
  \end{aligned}
\end{equation*}

To make the following derivations easier to follow we open up the expectations, writing them as integrals explicitly. For this purpose let us define the joint distribution over states $s \sim \mu(s)$ together with randomly sampled tasks $i \sim I$ as $\mu(s,i) = p(s | \mathcal{D}) \mathcal{U}(i \in I)$, where $\mathcal{U}$ denotes the uniform distribution over possible tasks. This allows us to re-write the expectations that include the corresponding distributions, i.e. $\mathbb{E}_{\mu(s), i \sim I}[1] = \mathbb{E}_{\mu(s, i)}[1]$, but again, note that $i$ here is not necessarily the task under which $s$ was observed.
We can then write the Lagrangian equation corresponding to the above described program as

\begin{equation*}
  \begin{aligned}
L(q,&\eta,\gamma) = \int \mu(s,i)\int q(a|s,i) \hat{Q}(s,a,i) \diff a \diff s \diff i \\
&+\eta\left(\epsilon - \int \mu(s,i)\int q(a|s,i)\log \frac{q(a|s,i)}{\pi(a|s,i,\theta_t)\diff a \diff s \diff i}\right) \\&+ \gamma\left(1 - \iint \mu(s,i) q(a|s,i)\diff a \diff s \diff i\right).
  \end{aligned}
\end{equation*}
Next we maximize the Lagrangian $L$ w.r.t the primal variable $q$. The derivative w.r.t $q$ reads,

\begin{equation*}
\begin{aligned}
{\partial q}L(q,\eta,\gamma) = &\hat{Q}(a,s,i) - \eta\log q(a|s,i) \\
&+\eta\log \pi(a|s,i,\theta_t) - \eta - \gamma.
\end{aligned}
\end{equation*}

Setting it to zero and rearranging terms we obtain

$$q(a|s,i) = \pi(a|s,i,\theta_t)\exp\left(\frac{\hat{Q}(s,a,i)}{\eta}\right)\exp\left(-\frac{\eta+\gamma}{\eta}\right).$$

However, the last exponential term is a normalization constant for $q$. Therefore we can write,

$$\exp\left(\frac{\eta+\gamma}{\eta}\right) = \int \pi(a|s,i, \theta_t)\exp\left(\frac{Q(s,a,i)}{\eta}\right)\diff a$$
\begin{equation}
\begin{aligned}
\frac{\eta+\gamma}{\eta}= \log\left(\int \pi(a|s,i,\theta_t)\exp\left(\frac{Q(s,a,i)}{\eta}\right)\diff a\right).
\label{eq:norm}
\end{aligned}
\end{equation}

Now, to obtain the dual function $g(\eta)$, we plug in the solution to the KL constraint term (second term) of the Lagrangian which yields 
\begin{equation*}
  \begin{aligned}
&L(q,\eta,\gamma) = \int \mu(s,i)\int q(a|s,i) Q(s,a,i) \diff a \diff s \\
&+\eta\biggl(\epsilon - \int \mu(s,i)\int q(a|s,i)\\
& ~~~~~~~~~~~~~~~~~~~ \log {\scriptscriptstyle \frac{\pi(a|s,i,\theta_t)\exp\left(\frac{Q(s,a,i)}{\eta}\right)\exp\left(-\frac{\eta+\gamma}{\eta}\right)}{\pi(a|s,i,\theta_t)} } \diff a \diff s \diff i\biggr) \\
&+ \gamma\left(1 - \iint \mu(s,i) q(a|s,i)\diff a \diff s \diff i\right).
  \end{aligned}
\end{equation*}

After expanding and rearranging terms we get
\begin{equation*}
  \begin{aligned}
& L(q,\eta,\gamma) = \int \mu(s,i)\int q(a|s,i) Q(s,a,i) \diff a \diff s \diff i \\&
- \eta\int \mu(s,i)\int q(a|s,i)\Big[\frac{Q(s,a,i)}{\eta} + \\
& ~~~~~~~~~~~~~~~~~~~ \log \pi(a|s,i;\theta_t) - \frac{\eta+\gamma}{\eta}\Big]\diff a \diff s \diff i \\& + \eta\epsilon
 +\eta\int \mu(s,i)\int q(a|s,i)\log \pi(a|s,i;\theta_t)\diff a \diff s \diff i \\ & + \gamma\left(1 - \iint \mu(s,i) q(a|s,i)\diff a \diff s \diff i\right).
  \end{aligned}
\end{equation*}

Most of the terms cancel out and after rearranging the terms we obtain
\begin{equation*}
  \begin{aligned}
L(q,\eta,\gamma) = \eta\epsilon + \eta\int \mu(s,i)\frac{\eta+\gamma}{\eta}\diff s \diff i.
  \end{aligned}
\end{equation*}

Note that we have already calculated the term inside the integral in Equation \ref{eq:norm}. By plugging in equation \ref{eq:norm} we obtain the dual function
\begin{equation}
\begin{aligned}
    g(\eta) = \eta\epsilon+\eta\int\mu(s,i)\log\Big(\int \pi(a|s,i, \theta_t)\\
    ~~~~~~~~~~~~~~~~~~~~~~ \exp\left(\frac{Q(s,a,i)}{\eta}\right)\diff a\Big)\diff s \diff i
    \label{eq:dual_eta},
\end{aligned}
\end{equation}
which we can minimize with respect to $\eta$ based on samples from the replay buffer.

\subsubsection{Extended Update Rules For Fitting a Mixture of Gaussians \label{sec:policyDerivation}}

After obtaining the non parametric policies, we fit a parametric policy to samples from said non-parametric policies -- effectively employing using maximum likelihood estimation with additional regularization based on a distance function $\mathcal{T}$, i.e,

\begin{equation}
\begin{aligned}
\theta^{(k+1)} &=\arg \min_{\theta} \mathbb{E}_{s \sim \mathcal{D}, i \sim I}\Big[ \mathrm{KL}\big( q_k(\cdot | s, i) \| \pi_{\theta}(\cdot | s, i) \big) \Big] \\
&= \argmin_{\theta}
\mathbb{E}_{s \sim \mathcal{D}, i \sim I, a \sim q(\cdot | s, i)}\Big[ -\log \pi_{\theta}(a |s, i) \Big], \quad \\
&= \argmax_{\theta}
\mathbb{E}_{s \sim \mathcal{D}, i \sim I, a \sim q(\cdot | s, i)}\Big[ \log \pi_{\theta}(a |s, i) \Big], \quad \\&\textrm{s.t.}\:\mathbb{E}_{s \sim \mathcal{D}, i \sim I} \Big[ \mathcal{T}(\pi_{\theta_{k}}(\cdot|s, i) | \pi_{\theta}(\cdot|s, i)) \Big] < \epsilon,
\end{aligned}
\label{eq:mpo_fststep}
\end{equation}

where $\mathcal{T}$ is an arbitrary distance function to evaluate the change of the new policy with respect to a reference/old policy, and $\epsilon$ denotes the allowed change for the policy. 

To make the above objective amenable to gradient based optimization we employ Lagrangian Relaxation, yielding the following primal:

\begin{equation}\label{eq:lagrange}
\begin{aligned}
\max_\theta \min_{\alpha > 0} L(\theta,\alpha) = \mathbb{E}_{ s \sim \mathcal{D},i \sim I, a \sim q(\cdot | s, i)}\Big[ \log \pi_{\theta}(a|s, i) \Big] + \\
\alpha\Big(\epsilon - \mathbb{E}_{s \sim \mathcal{D},i \sim I} \big[ \mathcal{T}(\pi_{\theta_{k}}(\cdot|s, i) , \pi_{\theta}(\cdot|s, i))\big]\Big).
\end{aligned}
\end{equation}

We solve for $\theta$ by iterating the inner and outer optimization programs independently: We fix the parameters $\theta$ to their current value and optimize for the Lagrangian multipliers (inner minimization) and then we fix the Lagrangian multipliers to their current value and optimize for $\theta$ (outer maximization). In practice we found that it is effective to simply perform one gradient step each in inner and outer optimization for each sampled batch of data.

The optimization given above is general, i.e. it works for any general type of policy. As described in the main paper, we consider hierarchical policies of the form 
\begin{eqnarray}\label{eq:policy2}
\pi_{\theta}(a | s, i) = \sum^M_{o=1} \pi^L_\theta \left(a | s, o\right) \pi^H_\theta\left(o | s, i\right).
\end{eqnarray}
In particular, in all experiments we made use of a mixture of Gaussians parametrization, where the high level policy $\pi^H_\theta$ is a categorical distribution over low level $\pi^L_\theta$ Gaussian policies, i.e,
\begin{align*}
\pi_{\theta}(a|s, i) &= \pi(a|\{\alpha_{\theta}^{j},\mu_{\theta}^{j},\Sigma_{\theta}^{j}\}(s)_{j=1...M}) \\
&=  \sum_{j=1}^M \alpha^j_{\theta}(s, i)\mathcal{N}(\mu^j_{\theta}(s),\Sigma^j_{\theta}(s)) \\ 
&\forall s \sum_{j=1}^M \alpha^j(s,i) = 1 \text{, and, } \ \alpha^j(s, i) > 0
\end{align*}
where $j$ denote the index of components and $\alpha$ is the high level policy $\pi^H$ assigning probabilities to each mixture component for a state s given the task and the low level policies are all Gaussian. Here $\alpha^j$s are the probabilities for a categorical distribution over the components.

We also define the following distance function between old and new mixture of Gaussian policies
$$
\mathcal{T}(\pi_{\theta_{k}}(\cdot|s, i),\pi_{\theta}(\cdot|s, i)) = \mathcal{T}_{H}(s, i) + \mathcal{T}_{L}(s)
$$

$$\mathcal{T}_{H}(s,i) = \textrm{KL}(\textrm{Cat}(\{\alpha_{\theta_{k}}^{j}(s, i)\}_{j=1...M}) \| \textrm{Cat}(\{\alpha_{\theta}^{j}(s, i)\}_{j=1...M}))$$

$$\mathcal{T}_{L}(s) = \frac{1}{M}\sum_{j=1}^{M}\textrm{KL}(\mathcal{N}(\mu^j_{\theta_{k}}(s),\Sigma^j_{\theta_{k}}(s)) \| \mathcal{N}(\mu^j_{\theta}(s),\Sigma^j_{\theta}(s)))$$

where $\mathcal{T}_{H}$ evaluates the KL between categorical distributions and $\mathcal{T}_{L}$ corresponds to the average KL across Gaussian components, as also described in the main paper (c.f. Equation 5 in the main paper).

In order to bound the change of categorical distributions, means and covariances of the components independently -- which makes it easy to control the convergence of the policy and which can prevent premature convergence as argued in \citet{abdolmaleki2018relative} -- we separate out the following three intermediate policies

\begin{align*}
\pi_{\theta}^{\Sigma}(a|s, i) = \pi(a|\{\alpha_{\theta_{k}}^{j},\mu_{\theta_{k}}^{j},\Sigma_{\theta}^{j}\}(s, i)_{j=1...M}) \\
\pi_{\theta}^{\mu}(a|s, i) = \pi(a|\{\alpha_{\theta_{k}}^{j},\mu_{\theta}^{j},\Sigma_{\theta_{k}}^{j}\}(s, i)_{j=1...M}) \\
\pi_{\theta}^{\alpha}(a|s, i) = \pi(a|\{\alpha_{\theta}^{j},\mu_{\theta_{k}}^{j},\Sigma_{\theta_{k}}^{j}\}(s, i)_{j=1...M})
\end{align*}

Which yields the following final optimization program

\begin{equation}
\begin{aligned}
\theta^{(k+1)} & = \argmax_{\theta}
\mathbb{E}_{s \sim \mathcal{D}, i \sim I, a \sim q(\cdot | s, i)}\Big[ \\
&\log \pi_{\theta}^{\mu}(a|s, i) + \log \pi_{\theta}^{\Sigma}(a|s, i) + \log \pi_{\theta}^{\alpha}(a|s, i)\Big], \quad \\
&\textrm{s.t.}\:\mathbb{E}_{s \sim \mathcal{D}, i \sim I} \Big[\mathcal{T}(\pi_{\theta_{k}}(a|s, i) | \pi_{\theta}^{\mu}(a|s, i))\Big] < \epsilon_{\mu},
\\
&\textrm{s.t.}\:\mathbb{E}_{s \sim \mathcal{D}, i \sim I} \Big[\mathcal{T}(\pi_{\theta_{k}}(a|s, i) | \pi_{\theta}^{\Sigma}(a|s, i))\Big] < \epsilon_{\Sigma},
\\
&\textrm{s.t.}\:\mathbb{E}_{s \sim \mathcal{D}, i \sim I} \Big[\mathcal{T}(\pi_{\theta_{k}}(a|s, i) | \pi_{\theta}^{\alpha}(a|s, i))\Big] < \epsilon_{\alpha},
\end{aligned}
\label{eq:mpo}
\end{equation}

This decoupling allows us to set different $\epsilon$ values for the change of means, covariances and categorical probabilities, i.e., $\epsilon_\mu , \epsilon_\Sigma, \epsilon_\alpha$. 
Different $\epsilon$ lead to different learning rates. We always set a much smaller epsilon for the covariance and categorical than for the mean. The intuition is that while we would like the distribution to converge quickly in the action space, we also want to keep the exploration both locally (via the covariance matrix) and globally (via the high level categorical distribution) to avoid premature convergence.   

\subsection{Algorithmic Details}
\subsubsection{Pseudocode for the full procedure}\label{sec:Pseudocode}
We provide a pseudo-code listing of the full optimization procedure -- and the asynchronous data gathering -- performed by \met~ in Algorithm \ref{alg:learner1} and \ref{alg:actor}. The implementation relies on Sonnet \citep{sonnetblog} and TensorFlow \citep{tensorflow2015-whitepaper}.

\begin{algorithm}
\caption{RHPO - Asynchronous Learner}\label{alg:learner1}
\begin{algorithmic}
\STATE \textbf{Input:} $N_{steps}$ number of update steps, $N_\text{targetUpdate}$ update steps between target update, $N_s$ number of action samples per state, KL regularization parameters $\epsilon$, initial parameters for $\pi,~\eta$ and $\phi$
\STATE initialize N = 0
\WHILE{$k \leq N_\text{steps}$}
% \STATE update replay buffer $B$ with received trajectories
\FOR{$k$ in $[0...N_\text{targetUpdate}]$}
\STATE sample a batch of trajectories $\tau$ from replay buffer $B$ 
\STATE sample $N_s$ actions from $\pi_{\theta_k}$ to estimate expectations below
\STATE // compute mean gradients over batch for policy, Lagrangian multipliers and Q-function
\STATE $\delta_\pi \leftarrow -\nabla_\theta \sum_{s_t \in \tau} \sum_{j=1}^{N_s} [ \exp\left(\frac{Q(s_t,a_j,i)}{\eta}\right)$
\STATE \hfill $\log \pi_{\theta}(a_j | s_t, i) ] $ following Eq. \ref{eq:objective_pi}
\STATE $\delta_{\eta} \leftarrow \nabla_\eta g(\eta) = \nabla_\eta \eta\epsilon+\eta \sum_{s_t \in \tau} \log \frac{1}{N_s} \sum_{j=1}^{N_s}[ $
\STATE \hfill $\exp\left(\frac{Q(s_t,a_j,i)}{\eta}\right) ]$ following Eq. \ref{eq:dual_eta}
\STATE $\delta_Q \leftarrow \nabla_{\phi} \sum_{i \sim I} \sum_{(s_t, a_t) \in \tau} \big( \hat{Q}_\phi(s_t, a_t, i) -
  Q^{\text{ret}} \big)^2$ 
\STATE \hfill  with $Q^{\text{ret}}$ following Eq. \ref{eq:objective_q_value}
\STATE // apply gradient updates
\STATE $\pi_{\theta_{k+1}} =$ optimizer\_update($\pi$, $\delta_\pi$), 
\STATE $\eta =$ optimizer\_update($\eta$, $\delta_\eta$) 
\STATE $\hat{Q}_\phi =$ optimizer\_update($\hat{Q}_\phi$, $\delta_Q$)
\STATE $k = k + 1$
\ENDFOR
\STATE // update target networks
\STATE $\pi' = \pi$, $Q' = Q$
\ENDWHILE
\end{algorithmic}
\end{algorithm}

\begin{algorithm}
\caption{RHPO - Asynchronous Actor}\label{alg:actor}
\begin{algorithmic}
\STATE \textbf{Input:} $N_\text{trajectories}$ number of total trajectories requested, $T$ steps per episode, $\xi$ scheduling period
\STATE initialize N = 0
% \STATE // Initialize Q-table
% \STATE $ \forall \cT_h, \cT_{0:h-1} : Q(\cT_{0:h-1}, \cT_h) = 0$, $M_{\cT_h} = 0$
%$Q(\cT_{0:h-1}, \cT_h) = \frac{1}{M} \sum_{i=1}^M R^\tau_\cM(\cT_{h:H}),$
\WHILE{$N < N_\text{trajectories}$}
\STATE fetch parameters $\theta$
\STATE // collect new trajectory from environment
\STATE $\tau = \lbrace \rbrace$%, h = 0$
\FOR{$t$ in $[0...T]$}
\IF{$t \pmod{\xi} \equiv 0$}
\STATE // sample active task from uniform distribution
\STATE $i_{act} \sim I$
% \STATE $h = h + 1$
\ENDIF
\STATE $a_t \sim \pi_\theta(\cdot | s_t, i_{act})$
\STATE // execute action and determine rewards for all tasks
\STATE $\bar{r} = \lbrack r_{i_1}(s_t, a_t), \dots, r_{i_{|I|}}(s_t, a_t) \rbrack$
\STATE $\tau \leftarrow \tau \cup \lbrace (s_t, a_t, \bar{r}, \pi_\theta(a_0 | s_t, i_{act})) \rbrace$
\ENDFOR
\STATE send batch trajectories $\tau$ to replay buffer
% \STATE // update Monte Carlo Q for scheduler$P_S$ 
% \FOR{h=0:H}
% \STATE $M_{\cT_h} = M_{\cT_h} + 1$
% \STATE $Q(\cT_{0:h-1}, \cT_h) \pluseq  \frac{R^\tau_\cM(\cT_{h:H}) - Q(\cT_{0:h-1}, \cT_h)}{M}$
% \ENDFOR
\STATE $N = N + 1$
\ENDWHILE
\end{algorithmic}
\end{algorithm}

\subsubsection{Details on the policy improvement step \label{sec:qtrace}}
As described in the main paper we consider the same setting as
scheduled auxiliary control setting (SAC-X) \citep{riedmiller2018learning} to perform policy improvement (with uniform random switches between tasks every N steps within an episode, the SAC-U setting).

Given a replay buffer containing data gathered from all tasks, where for each trajectory snippet $\tau = \lbrace (s_0, a_0, R_0), \dots, (s_L, a_L, R_L) \rbrace$ we record the rewards for all tasks $R_t = [r_{i_1}(s_t, a_t), \dots, r_{i_{|I|}}(s_t, a_t)]$ as a vector in the buffer, we define the retrace objective for learning $\hat{Q}$, parameterized via $\phi$, following \citet{riedmiller2018learning} as
\begin{equation}
\begin{aligned}
  &\min_\phi L(\phi) = \sum_{i \sim I} \mathbb{E}_{\tau \sim \mathcal{D}} \Big[ \big( r_i(s_t, a_t) +  \\
  & \gamma Q^{ret}(s_{t+1}, a_{t+1}, i) - \hat{Q}_\phi(s_t, a_t, i))^2 \Big], \\
  &   \text{with } Q^{\text{ret}}(s_{t}, a_t, i) = \sum_{j=t}^\infty \Big( \gamma^{j-t} \prod_{k=t}^j c_k \Big)\Big[ \delta_Q(s_j, s_{j+1}) \Big], \\ 
  & \delta_Q(s_j, s_{j+1}) =   r_i(s_j, a_j) + \\
  & \gamma \mathbb{E}_{\pi_{\theta_{k}}(a | s_{j+1}, i)},
  [ \hat{Q}_{\phi'}(s_{j+1}, \cdot, i; \phi') ] - \hat{Q}_{\phi'}(s_j, a_j, i),
\end{aligned}
\label{eq:objective_q_value2}
\end{equation}
where the importance weights are defined as $c_k = \min\left(1, \nicefrac{\pi_{\theta_k}(a_k | s_k, i)}{b(a_k | s_k)}\right)$, with $b(a_k | s_k)$ denoting an arbitrary behavior policy; in particular this will be the policy for the executed tasks during an episode as in \citep{riedmiller2018learning}. Note that, in practice, we truncate the infinite sum after $L$ steps, bootstrapping with $\hat{Q}$. We further perform optimization of Equation \eqref{eq:objective_q_value} via gradient descent and make use of a target network~\citep{mnih2015human}, denoted with parameters $\phi'$, which we copy from $\phi$ after a couple of gradient steps.
We reiterate that, as the state-action value function $\hat{Q}$ remains independent of the policy's structure, we are able to utilize any other off-the-shelf Q-learning algorithm such as TD(0) \citep{sutton1988learning}.

Given that we utilize the same policy evaluation mechanism as SAC-U it is worth pausing here to identify the differences between SAC-U and our approach. The main difference is in the policy parameterization: SAC-U used a monolithic policy for each task $\pi(a | s, i)$ (although a neural network with shared components, potentially leading to some implicit task transfer, was used). Furthermore, we perform policy optimization based on MPO instead of using stochastic value gradients (SVG \citep{heess2016learning}). We can thus recover a variant of plain SAC-U using MPO if we drop the hierarchical policy parameterization, which we employ in the single task experiments in the main paper.

\subsubsection{Network Architectures}\label{app:architectures}

\begin{figure}[ht]
\vskip 0.2in
\begin{center}
\centerline{\includegraphics[height=0.35\columnwidth]{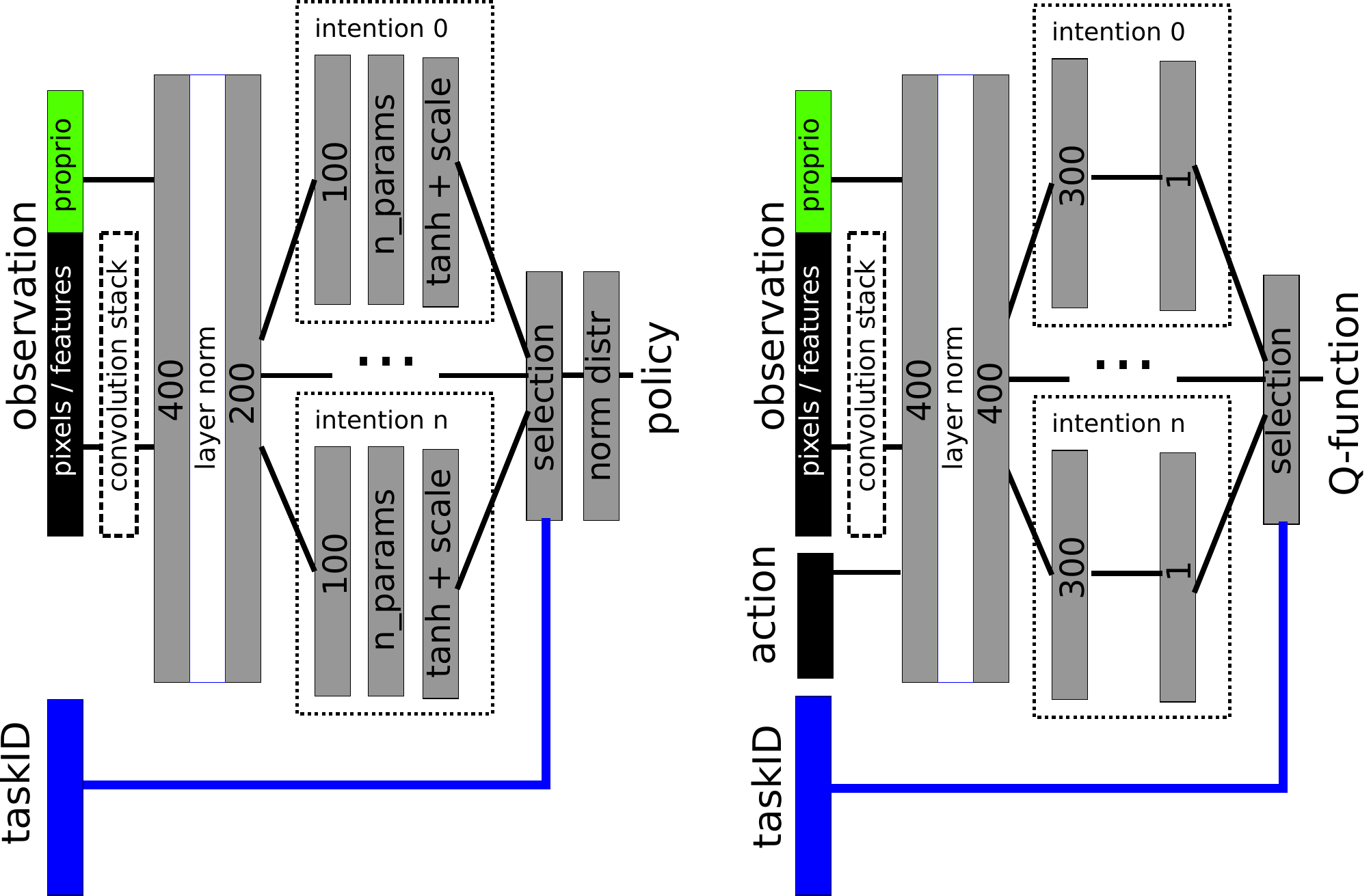}}
\caption{\small \label{fig:sac_networks}Schematics of the used networks. While we use the Q-function (right sub-figure) architecture in all multitask experiments, we investigate variations of the policy architecture (left sub-figure) in this paper (see Figure \ref{fig:sac_networks_alternative_policy}).}
\end{center}
\vskip -0.2in
\end{figure}

To represent the Q-function in the multitask case we use the network architecture from SAC-X (see right sub-figure in Figure \ref{fig:sac_networks}).
The proprioception of the robot, the features of the objects and the actions are fed together in a torso network.
At the input we use a fully connected first layer of 200 units, followed by a layer normalization operator, an optional tanh activation and another fully connected layer of 200 units with an ELU activation function.
The output of this torso network is shared by independent head networks for each of the tasks (or intentions, as they are called in the SAC-X paper).
Each head has two fully connected layers and outputs a Q-value for this task, given the input of the network. Using the task identifier we then can compute the Q value for a given sample by discrete selection of the according head output.

While we use the network architecture for the Q function for all multitask experiments, we investigate different architectures for the policy in this paper.
The original SAC-X policy architecture is shown in Figure \ref{fig:sac_networks} (left sub-figure).
The main structure follows the same basic principle that we use in the Q function architecture. The only difference is that the heads compute the required parameters for the policy distribution we want to use (see subsection \ref{subsec:algo_hypers}). This architecture is referenced as the independent heads (or task dependent heads).

\begin{figure}[ht]
\vskip 0.2in
\begin{center}
\centerline{
  \includegraphics[height=0.3\columnwidth]{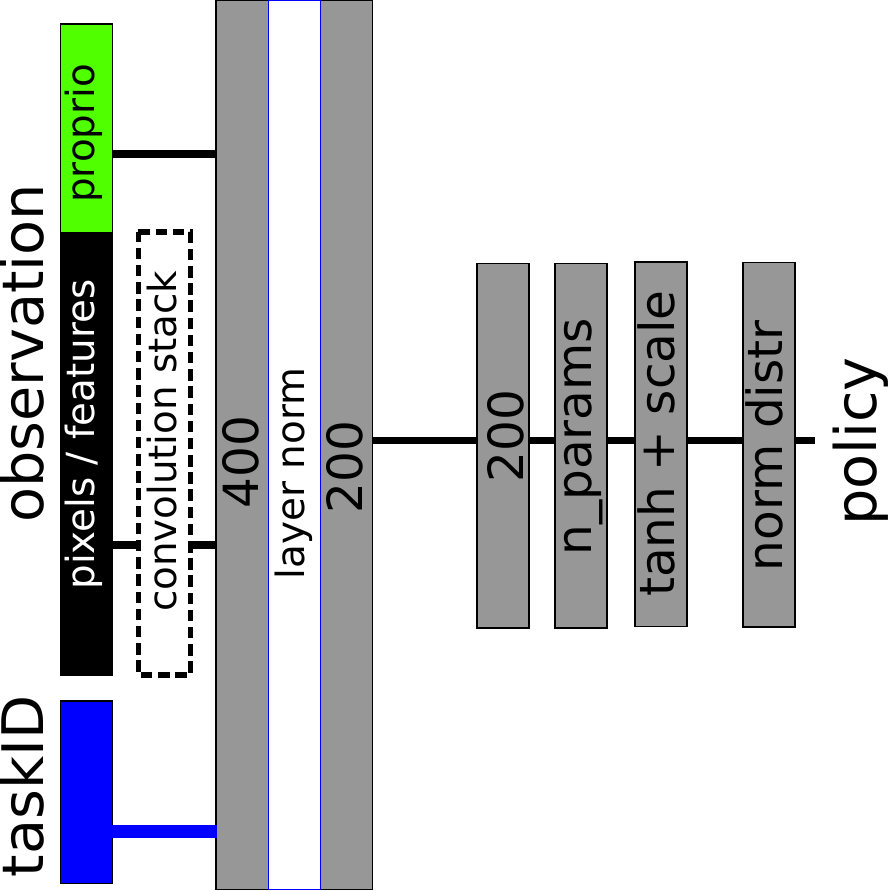}\hspace{0.02\columnwidth}
  \includegraphics[height=0.3\columnwidth]{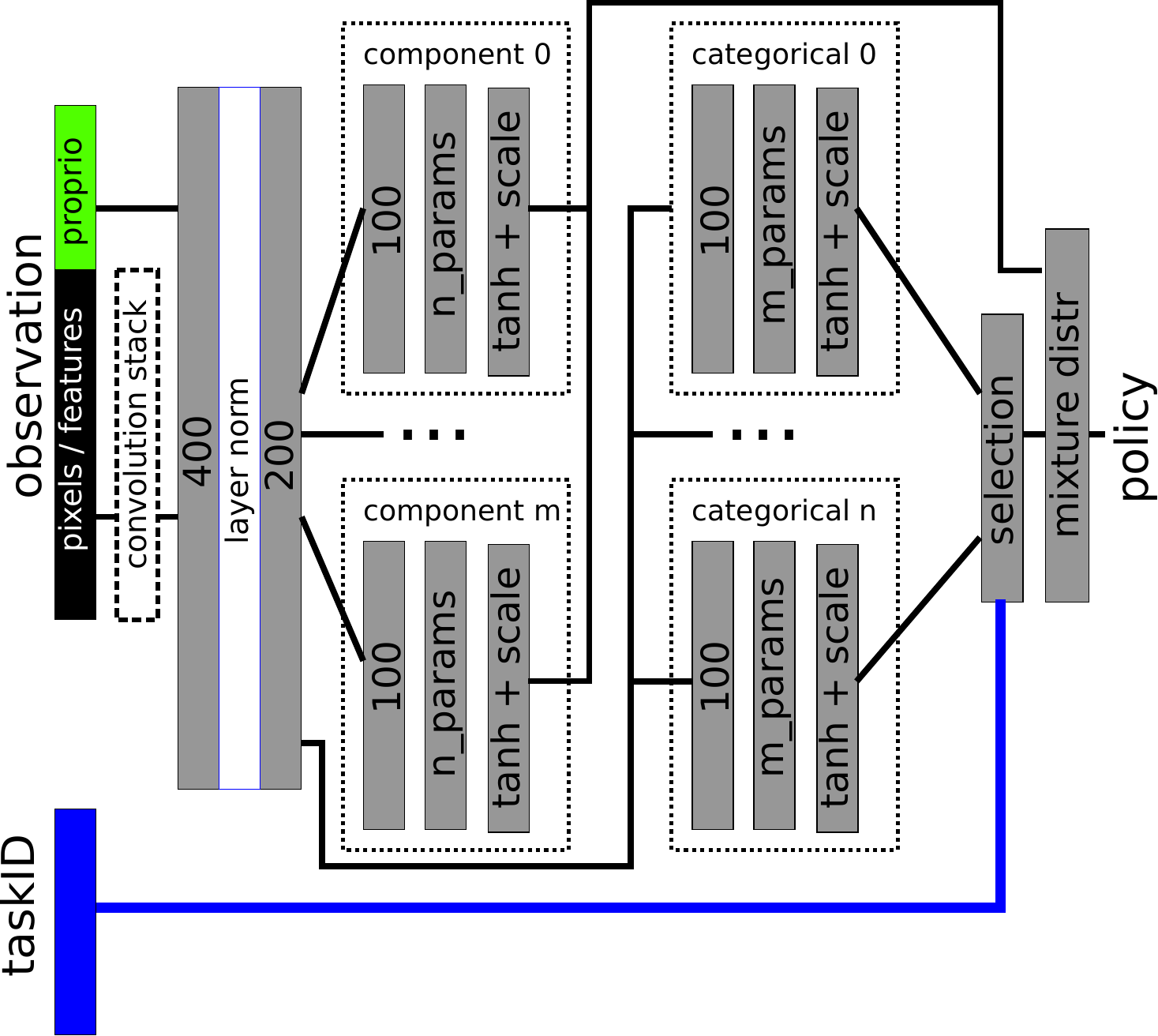}
}
\caption{\small \label{fig:sac_networks_alternative_policy}Schematics of the alternative multitask policy architectures used in this paper. Left sub-figure: the monolithic architecture; Right sub-figure: the hierarchical architecture.}
\end{center}
\vskip -0.2in
\end{figure}

The alternatives we investigate in this paper are the monolithic policy architecture (see Figure \ref{fig:sac_networks_alternative_policy}, left sub-figure) and the hierarchical policy architecture (see Figure \ref{fig:sac_networks_alternative_policy}, right sub-figure).
For the monolithic policy architecture we reduce the original policy architecture basically to one head and append the task-id as a one-hot encoded vector to the input. 
For the hierarchical architecture, we build on the same torso and create a set of networks parameterizing the Gaussians which are shared across tasks and a task-specific network to parameterize the categorical distribution for each task. The final mixture distribution is task-dependent for the high-level controller but task-independent for the low-level policies.

\subsubsection{Algorithm Hyperparameters}
\label{subsec:algo_hypers}

In this section we outline the details on the hyperparameters used for \met~ and baselines in both single task and multitask experiments.
All experiments use feed-forward neural networks. We consider a flat policy represented by a Gaussian distribution and a hierarchical policy represented by a mixture of Gaussians distribution. The flat policy is given by a Gaussian distribution with a diagonal covariance matrix, i.e, 
$
  \pi(a|s,\theta) = \mathcal{N}\left(\mu , 
  \Sigma \right)
$. The neural network outputs the mean $\mu=\mu(s)$ and diagonal Cholesky factors $A=A(s)$, such that $\Sigma = AA^T$. The diagonal factor $A$ has positive diagonal elements enforced by the softplus transform $A_{ii} \leftarrow \log(1 + \exp(A_{ii}))$ to ensure positive definiteness of the diagonal covariance matrix.
Mixture of Gaussian policy has a number of Gaussian components as well as a categorical distribution for selecting the components. The neural network outputs the Gaussian components based on the same setup described above for a single Gaussian and outputs the logits for representing the categorical distribution.
Tables \ref{tab:MPOSingle} show the hyperparameters we used for the single tasks experiments. We found layer normalization and a hyperbolic tangent ($tanh$) on the layer following the layer normalization are important for stability of the algorithms. For \met~ the most important hyperparameters are the constraints in Step 1 and Step 2 of the algorithm.

\begin{table}[t]
\begin{center}
 \begin{tabular}{c||c||c} 
 Hyperparameters & Hierarchical & Single Gaussian \\
 \hline
 Policy net & 200-200-200& 200-200-200\\ 
 Number of actions sampled per state& 10& 10\\
 Q function net & 500-500-500& 500-500-500\\
 Number of components & 3& NA\\
 $\epsilon$ & 0.1& 0.1 \\
 $\epsilon_{\mu}$ & 0.0005& 0.0005 \\
 $\epsilon_{\Sigma}$ & 0.00001& 0.00001\\
  $\epsilon_{\alpha}$ & 0.0001& NA\\
 Discount factor ($\gamma$) & 0.99& 0.99 \\
 Adam learning rate & 0.0002& 0.0002 \\
 Replay buffer size & 2000000& 2000000 \\
 Target network update period & 250 & 250\\
 Batch size & 256& 256\\
 Activation function & elu& elu\\
 Layer norm on first layer & Yes& Yes\\
 Tanh on output of layer norm & Yes& Yes\\
 Tanh on input actions to Q-function & Yes& Yes \\
 Retrace sequence length & 10& 10   
\end{tabular}
\end{center}
\caption{Hyperparameters - Single Task}
\label{tab:MPOSingle}
\end{table}

\begin{table}[t]
\begin{center}
 \begin{tabular}{c||c|c|c} 
 Hyperparameters & Hierarchical & Independent & Monolith \\
 \hline
 Policy torso \\(shared across tasks) & \multicolumn{3}{c}{400-200}\\ 
 \begin{tabular}{@{}c@{}} Policy task-dependent \\ heads \end{tabular} & \begin{tabular}{@{}c@{}} 100 \\ (controller) \end{tabular} & 100& NA\\ 
 \begin{tabular}{@{}c@{}}  Policy shared \\ heads \end{tabular} & \begin{tabular}{@{}c@{}} 100 \\ (components) \end{tabular} & NA & 200\\ 
 Number of action samples& \multicolumn{3}{c}{20}\\
 Q function torso \\(shared across tasks) & \multicolumn{3}{c}{400-400}\\
 Q function head \\(per task)& \multicolumn{3}{c}{300 }\\
 Number of components & \begin{tabular}{@{}c@{}} number of \\ tasks \end{tabular} & NA&NA\\
 Discount factor ($\gamma$) & \multicolumn{3}{c}{0.99} \\
 Replay buffer size & \multicolumn{3}{c}{1e6 * number of tasks} \\
 Target network \\ update period & \multicolumn{3}{c}{500}\\
 Batch size & \multicolumn{3}{c}{256 (512 for Pile1)}\\
\end{tabular}
\end{center}
\caption{Hyperparameters - Multitask. Values are taken from the single task experiments with the above mentioned changes.}
\label{tab:MPOMulti}
\end{table}

% \markus{considered hyperparameters}

\subsection{Additional Details on the SAC-U with SVG baseline}
For the SAC-U baseline we used a re-implementation of the method from \citep{riedmiller2018learning} using SVG \citep{heess2015learning} for optimizing the policy. Concretely we use the same basic network structure as for the "Monolithic" baseline with MPO and parameterize the
policy as 
$$
\pi_\theta = \mathcal{N}(\mu_\theta(s, i),\sigma^2_\theta(s, i) I),
$$
where $I$ denotes the identity matrix and $\sigma_\theta(s, i)$ is computed from the network output via a softplus activation function. 

Together with entropy regularization, as described in \citep{riedmiller2018learning} the policy can be optimized via gradient ascent, following the reparameterized gradient for a given states sampled from the replay:
\begin{equation}
\begin{aligned}
\nabla_\theta \mathbb{E}_{\pi_\theta(a | s, i)}[\hat{Q}(a, s, i)] + \alpha \mathrm{H}\Big(\pi_\theta(a | s, i)\Big),
\end{aligned}
\end{equation}
which can be computed, using the reparameterization trick, as 
\begin{eqnarray}
\begin{aligned}
 \mathbb{E}_{\zeta \sim \mathcal{N}(0, I)}[\nabla_\theta g_\theta(s, i, \zeta) \nabla_g Q(g_\theta(s, i, \zeta), s, i)] + \\
 \alpha \nabla_\theta \mathrm{H}\Big(\pi_\theta(a | s, i)\Big),
\end{aligned}
\label{eq:svg}
\end{eqnarray}
where $g_\theta(s, i, \zeta) = \mu_\theta(s, i) + \sigma_\theta(\bs) * \zeta$ is now a deterministic function of a sample from the standard multivariate normal distribution. See e.g. \cite{heess2015learning} (for SVG) as well as \cite{kingma2013auto} (for the reparameterization trick) for a detailed explanation.

\subsection{Details on the Experimental Setup \label{sec:ExperimentalSetup}}

\subsubsection{Simulation (Single- and Multitask)}

For the simulation of the robot arm experiments the numerical simulator MuJoCo \footnote{MuJoCo: see www.mujoco.org} was used -- using a model we identified from the real robot setup.

We run experiments of length 2 - 7 days for the simulation experiments (depending on the task) with access to 2-5 recent CPUs with 32 cores each (depending on the number of actors) and 2 recent NVIDIA GPUs for the learner. Computation for data buffering is negligible.

\subsubsection{Real Robot Multitask}

Compared to simulation where the ground truth position of all objects is known, in the real robot setting, three cameras on the basket track the cube using fiducials (augmented reality tags).

For safety reasons, external forces are measured at the wrist and the episode is terminated if a threshold of 20N on any of the three principle axes is exceeded (this is handled as a terminal state with reward 0 for the agent), adding further to the difficulty of the task.

The real robot setup differs from the simulation in the reset behaviour between episodes, since objects need to be physically moved around when randomizing, which takes a considerable amount of time. To keep overhead small, object positions are randomized only every 25 episodes, using a hand-coded controller. Objects are also placed back in the basket if they were thrown out during the previous episode. Other than that, objects start in the same place as they were left in the previous episode. The robot's starting pose is randomized each episode, as in simulation.

\subsection{Task Descriptions}

\subsubsection{Pile1}
\label{task_pile1}

\begin{figure}[ht]
\vskip 0.2in
\begin{center}
\centerline{
  \includegraphics[width=0.35\columnwidth]{images/SawyerDE_3bricks.png}
  \includegraphics[width=0.29\columnwidth]{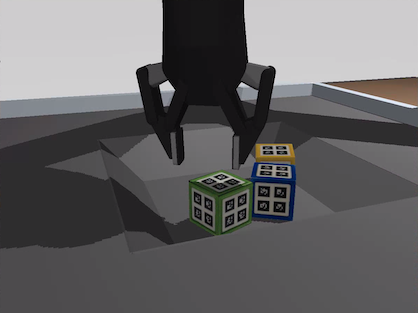}
  \includegraphics[width=0.29\columnwidth]{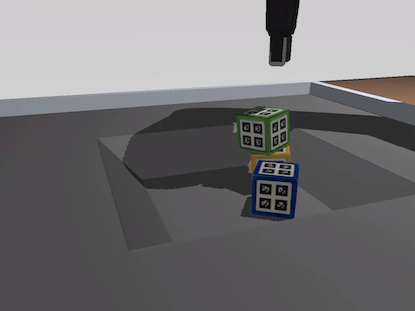}
}
\caption{Sawyer Set-Up.}
\label{fig:sawyer_setup_sim}
\end{center}
\vskip -0.2in
\end{figure}

For this task we have a real setup and a MuJoCo simulation that are well aligned. It consists of a Sawyer robot arm mounted on a table and equipped with a Robotiq 2F-85 parallel gripper. In front of the robot there is a basket of size 20x20 cm which contains three cubes with an edge length of 5 cm (see Figure \ref{fig:sawyer_setup_sim}).

The agent is provided with proprioception information for the arm (joint positions, velocities and torques), and the tool center point position computed via forward kinematics. For the gripper, it receives the motor position and velocity, as well as a binary grasp flag. It also receives a wrist sensor's force and torque readings. Finally, it is provided with the cubes' poses as estimated via the fiducials, and the relative distances between the arm's tool center point and each object. At each time step, a history of two previous observations is provided to the agent, along with the last two joint control commands, in order to account for potential communication delays on the real robot. The observation space is detailed in Table \ref{tab:sawyer_observations}.

The robot arm is controlled in Cartesian mode at 20Hz. The action space for the agent is 5-dimensional, as detailed in Table \ref{tab:pile_actions}. The gripper movement is also restricted to a cubic volume above the basket using virtual walls.

\begin{table}[t]
\caption{Action space for the Sawyer experiments.}
\label{sawyer-action-table2}
\vskip 0.15in
\begin{center}
\begin{small}
\begin{tabular}{lccc}
\toprule
Entry & Dims & Unit & Range \\
\midrule
Translational Velocity in x, y, z & 3 & m/s & [-0.07, 0.07] \\
Wrist Rotation Velocity & 1 & rad/s & [-1, 1] \\
Finger speed & 1 & tics/s & [-255, 255] \\
\bottomrule
\end{tabular}
\end{small}
\end{center}
\vskip -0.1in
\label{tab:pile_actions}
\end{table}

\begin{table}[t]
\caption{Observation used in the experiments with the Sawyer arm. An object's pose is represented as its world coordinate position and quaternion. In the table, $m$ denotes meters, $rad$ denotes radians, and $q$ refers to a quaternion in arbitrary units ($au$).}
\label{tab:sawyer_observations}
\vskip 0.15in
\begin{center}
\begin{small}
\begin{tabular}{lcc}
\toprule
Entry & Dims & Unit \\
\midrule
Joint Position (Arm) & 7 & rad \\
Joint Velocity (Arm) & 7 & rad/s \\
Joint Torque (Arm) & 7 & Nm \\
Joint Position (Hand) & 1 & rad \\
Joint Velocity (Hand) & 1 & tics/s \\
Force-Torque (Wrist) & 6 & N, Nm \\
Binary Grasp Sensor & 1 & au \\
TCP Pose & 7 & m, au \\
Last Control Command (Joint Velocity) & 8 & rad/s \\
Green Cube Pose & 7 & m, au \\
Green Cube Relative Pose & 7 & m, au \\
Yellow Cube Pose & 7 & m, au \\
Yellow Cube Relative Pose & 7 & m, au \\
Blue Cube Pose & 7 & m, au \\
Blue Cube Relative Pose & 7 & m, au \\
\bottomrule
\end{tabular}
\end{small}
\end{center}
\vskip -0.1in
\end{table}

For the Pile1 experiment we use 7 different task to learn, following the SAC-X principles. The first 6 tasks are seen as auxiliary tasks that help to learn the final task (STACK\_AND\_LEAVE(G, Y)) of stacking the green cube on top of the yellow cube.
Overview of the used tasks: 
\begin{itemize}
    \item \textit{REACH(G)}: $stol(d(TCP, G), 0.02, 0.15)$: \\
    Minimize the distance of the TCP to the green cube.
    \item \textit{GRASP}: \\
    Activate grasp sensor of gripper ("inward grasp signal" of Robotiq gripper)
    \item \textit{LIFT(G)}: $slin(G, 0.03, 0.10)$ \\
    Increase z coordinate of an object more than 3cm relative to the table.
    \item \textit{PLACE\_WIDE(G, Y)}: $stol(d(G, Y + [0,0,0.05]), 0.01, 0.20)$\\
    Bring green cube to a position 5cm above the yellow cube.
    \item \textit{PLACE\_NARROW(G, Y)}: $stol(d(G, Y + [0,0,0.05]), 0.00, 0.01)$: \\
    Like PLACE\_WIDE(G, Y) but more precise.
    \item \textit{STACK(G, Y)}: $btol(d_{xy}(G, Y), 0.03) * btol(d_z(G, Y) + 0.05, 0.01) * (1 - \textit{GRASP})$ \\
    Sparse binary reward for bringing the green cube on top of the yellow one (with 3cm tolerance horizontally and 1cm vertically) and disengaging the grasp sensor.
    \item \textit{STACK\_AND\_LEAVE(G, Y)}: $ stol(d_z(TCP, G)+0.10, 0.03, 0.10) * \textit{STACK(G, Y)}$ \\
    Like STACK(G, Y), but needs to move the arm 10cm above the green cube.
\end{itemize}

Let $d(o_i, o_j)$ be the distance between the reference of two objects (the reference of the cubes are the center of mass, TCP is the reference of the gripper), and let $d_{A}$ be the distance only in the dimensions denoted by the set of axes $A$. We can define the reward function details by:

\begin{equation}
stol(v, \epsilon, r) =
\begin{cases}
  1 &\text{iff} \ |v| < \epsilon \\
  1 - tanh^2( \frac{atanh(\sqrt{0.95})}{r} |v|) &\text{else}
\end{cases}
\label{eq:shaped_tolerance}
\end{equation}

\begin{equation}
slin(v, \epsilon_{min}, \epsilon_{max}) =
\begin{cases}
  0 &\text{iff} \ v < \epsilon_{min} \\
  1 &\text{iff} \ v > \epsilon_{max} \\
  \frac{v - \epsilon_{min}}{\epsilon_{max} - \epsilon_{min}}  &\text{else}
\end{cases}
\label{eq:shaped_tolerance2}
\end{equation}

\begin{equation}
btol(v, \epsilon) =
\begin{cases}
  1 &\text{iff} \ |v| < \epsilon  \\
  0 &\text{else}
\end{cases}
\end{equation}

\subsubsection{Pile2}
\label{task_pile2}

\begin{figure}[ht]
\vskip 0.2in
\begin{center}
\centerline{
  \includegraphics[width=0.3\columnwidth]{images/pile2_1.png}
  \includegraphics[width=0.3\columnwidth]{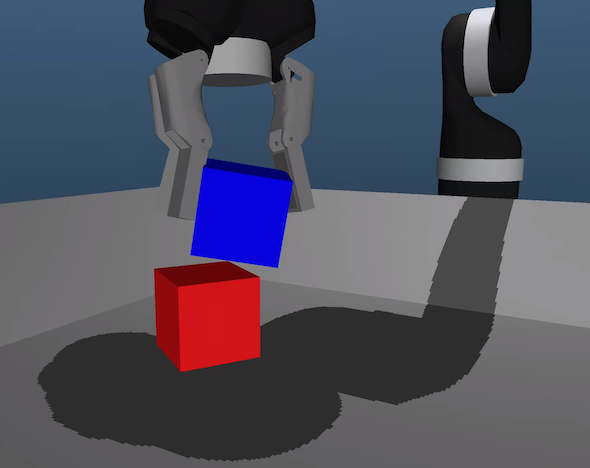}
  \includegraphics[width=0.3\columnwidth]{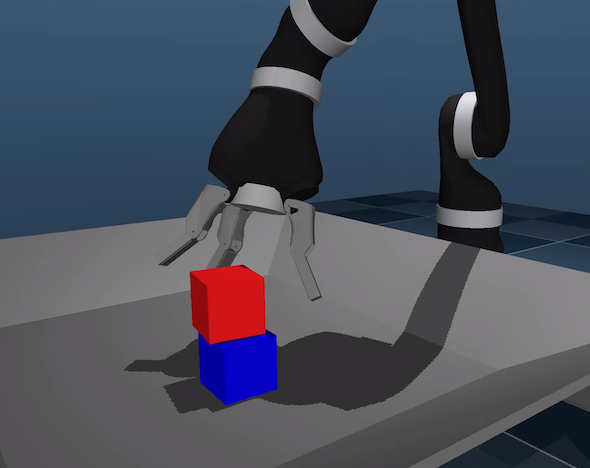}
}
\caption{\small \label{fig:pile2_setup_sim} The \textit{Pile2} set-up in simulation with two main tasks: The first is to stack the blue on the red cube, the second is to stack the red on the blue cube.}
\end{center}
\vskip -0.2in
\end{figure}

For the \textit{Pile2} task, taken from \cite{riedmiller2018learning}, we use a different robot arm, control mode and task setup to emphasize that \met's improvements are not restricted to cartesian control or a specific robot and that the approach also works for multiple external tasks.

Here, the agent controls a simulated Kinova Jaco robot arm, equipped with a Kinova KG-3 gripper. The robot faces a 40 x 40 cm basket that contains a red cube and a blue cube. Both cubes have an edge length of 5 cm (see Figure \ref{fig:pile2_setup_sim}). The agent is provided with proprioceptive information for the arm and the fingers (joint positions and velocities) as well as the tool center point position (TCP) computed via forward kinematics. Further, the simulated gripper is equipped with a touch sensor for each of the three fingers, whose value is provided to the agent as well. Finally, the agent receives the cubes' poses, their translational and rotational velocities and the relative distances between the arm's tool center point and each object. Neither observation nor action history is used in the \textit{Pile2} experiments. The cubes are spawned at random on the table surface and the robot hand is initialized randomly above the table-top with a height offset of up to 20 cm above the table (minimum 10 cm). The observation space is detailed in Table \ref{tab:jaco_observations}.

\begin{table}[t]
\caption{Action space used in the experiments with the Kinova Jaco Arm.}
\label{sawyer-action-table}
\vskip 0.15in
\begin{center}
\begin{small}
\begin{tabular}{lccc}
\toprule
Entry & Dims & Unit & Range \\
\midrule
Joint Velocity (Arm) & 6 & rad/sec & [-0.8, 0.8] \\
Joint Velocity (Hand) & 3 & rad/sec & [-0.8, 0.8] \\
\bottomrule
\end{tabular}
\end{small}
\end{center}
\vskip -0.1in
\label{tab:jaco_actions}
\end{table}

The robot arm is controlled in raw joint velocity mode at 20 Hz. The action space is 9-dimensional as detailed in Table \ref{tab:jaco_actions}. There are no virtual walls and the robot's movement is solely restricted by the velocity limits and the objects in the scene.

Analogous to \textit{Pile1} and the SAC-X setup, we use 10 different task for \textit{Pile2}. The first 8 tasks are seen as auxiliary tasks, that the agent uses to learn the main \textit{two} tasks \textit{PILE\_RED} and \textit{PILE\_BLUE}, which represent stacking the red cube on the blue cube and stacking the blue cube on the red cube respectively. The tasks used in the experiment are:

\begin{itemize}
    \item \textit{REACH(R)} = $stol(d(TCP, R), 0.01, 0.25)$: \\
    Minimize the distance of the TCP to the red cube.
    \item \textit{REACH(B)} = $stol(d(TCP, B), 0.01, 0.25)$: \\
    Minimize the distance of the TCP to the blue cube.
    \item \textit{MOVE(R)} = $slin(|\,linvel(R)\,|, 0, 1)$: \\
    Move the red cube.
    \item \textit{MOVE(B)} = $slin(|\,linvel(B)\,|, 0, 1)$: \\
    Move the blue cube.
    \item \textit{LIFT(R)} = $btol(pos_z(R), 0.05)$ \\
    Increase the z-coordinate of the red cube to more than 5cm relative to the table.
    \item \textit{LIFT(B)} = $btol(pos_z(B), 0.05)$ \\
    Increase the z-coordinate of the blue cube to more than 5cm relative to the table.
    \item \textit{ABOVE\_CLOSE(R, B)} = $above(R, B) * stol(d(R, B), 0.05, 0.2)$ \\
    Bring the red cube to a position above of and close to the blue cube.
    \item \textit{ABOVE\_CLOSE(B, R)} = $above(B, R) * stol(d(R, B), 0.05, 0.2)$ \\
    Bring the blue cube to a position above of and close to the red cube.
    \item \textit{PILE(R)}: \\
    Place the red cube on another object (touches the top). Only given when the cube doesn't touch the robot or the table.
    \item \textit{PILE(B)}: \\
    Place the blue cube on another object (touches the top). Only given when the cube doesn't touch the robot or the table.
\end{itemize}

The sparse reward \textit{above(A, B)} is given by comparing the bounding boxes of the two objects \textit{A} and \textit{B}. If the bounding box of object A is completely above the highest point of object B's bounding box, \textit{above(A, B)} is $1$, otherwise \textit{above(A, B)} is $0$. 

\begin{table}[t]
\caption{\small Observation used in the experiments with the Kinova Jaco Arm. An object's pose is represented as its world coordinate position and quaternion. The lid position and velocity are only used in the \textit{Clean-Up} task. In the table, $m$ denotes meters, $rad$ denotes radians, and $q$ refers to a quaternion in arbitrary units ($au$).}
\label{tab:jaco_observations}
\vskip 0.15in
\begin{center}
\begin{small}
\begin{tabular}{lcc}
\toprule
Entry & Dims & Unit \\
\midrule
Joint Position (Arm) & 6 & rad \\
Joint Velocity (Arm) & 6 & rad/s \\
Joint Position (Hand) & 3 & rad \\
Joint Velocity (Hand) & 3 & rad/s \\
TCP Position & 3 & m \\
Touch Force (Fingers) & 3 & N \\
Red Cube Pose & 7 & m, au \\
Red Cube Velocity & 6 & m/s, dq/dt \\
Red Cube Relative Position & 3 & m \\
Blue Cube Pose & 7 & m, au \\
Blue Cube Velocity & 6 & m/s, dq/dt \\
Blue Cube Relative Position & 3 & m \\
Lid Position & 1 & rad \\
Lid Velocity & 1 & rad/s \\
\bottomrule
\end{tabular}
\end{small}
\end{center}
\vskip -0.1in
\end{table}

\subsubsection{Clean-Up}
\label{task_cleanup}
\begin{figure}[ht]
\vskip 0.2in
\begin{center}
\centerline{
  \includegraphics[width=0.3\columnwidth]{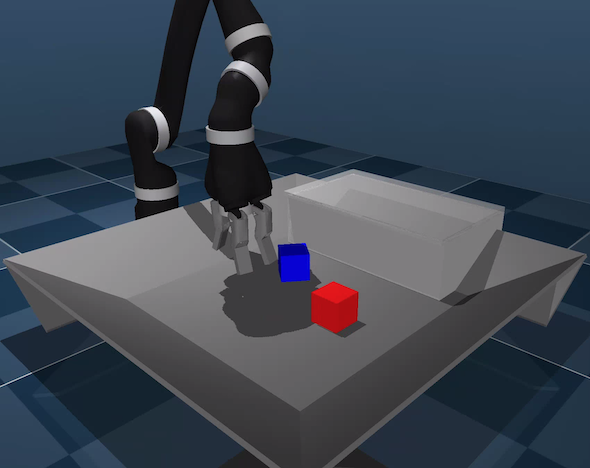}
  \includegraphics[width=0.3\columnwidth]{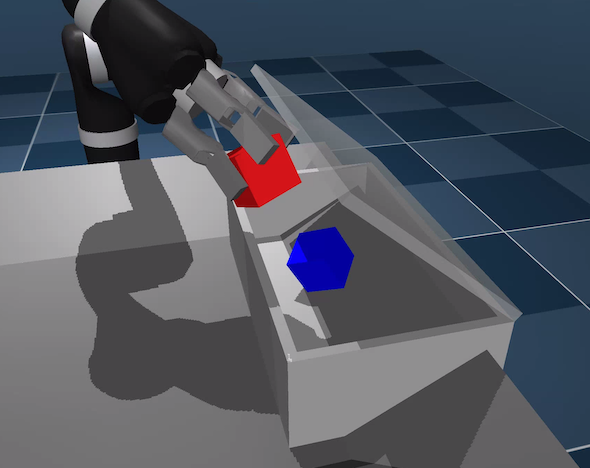}
  \includegraphics[width=0.3\columnwidth]{images/cleanup_3.png}
}
\caption{\small \label{fig:cleanup_setup_sim}The \textit{Clean-Up} task set-up in simulation. The task is solved when both bricks are in the box.}
\end{center}
\vskip -0.2in
\end{figure}

% For the cleanup experiment, we add the lid angle and lid angle velocity, which gives a total of 58 observations for this experiment.
The \textit{Clean-Up} task is also taken from \cite{riedmiller2018learning} and builds on the setup described for the \textit{Pile2} task. Besides the two cubes, the work-space contains an additional box with a moveable lid, that is always closed initially (see Figure \ref{fig:cleanup_setup_sim}). The agent's goal is to clean up the scene by placing the cubes inside the box. In addition to the observations used in the \textit{Pile2} task, the agent observes the lid's angle and it's angle velocity.

Analogous to \textit{Pile2} and the SAC-X setup, we use 13 different task for \textit{Clean-Up}. The first 12 tasks are seen as auxiliary tasks, that the agent uses to learn the main task \textit{ALL\_INSIDE\_BOX}. The tasks used in this experiments are:

\begin{itemize}
    \item \textit{REACH(R)} = $stol(d(TCP, R), 0.01, 0.25)$: \\
    Minimize the distance of the TCP to the red cube.
    \item \textit{REACH(B)} = $stol(d(TCP, B), 0.01, 0.25)$: \\
    Minimize the distance of the TCP to the blue cube.
    \item \textit{MOVE(R)} = $slin(|\,linvel(R)\,|, 0, 1)$: \\
    Move the red cube.
    \item \textit{MOVE(B)} = $slin(|\,linvel(B)\,|, 0, 1)$: \\
    Move the blue cube.
    \item \textit{NO\_TOUCH} = $1 - GRASP$ \\
    Sparse binary reward, given when neither of the touch sensors is active.
    \item \textit{LIFT(R)} = $btol(pos_z(R), 0.05)$ \\
    Increase the z-coordinate of the red cube to more than 5cm relative to the table.
    \item \textit{LIFT(B)} = $btol(pos_z(B), 0.05)$ \\
    Increase the z-coordinate of the blue cube to more than 5cm relative to the table.
    \item \textit{OPEN\_BOX} = $slin(angle(lid), 0.01, 1.5)$ \\
    Open the lid up to 85 degrees.
    \item \textit{ABOVE\_CLOSE(R, BOX)} = $above(R, BOX) * btol(|d(R, BOX)|, 0.2)$ \\
    Bring the red cube to a position above of and close to the box.
    \item \textit{ABOVE\_CLOSE(B, BOX)} = $above(B, BOX) * btol(|d(B, BOX)|, 0.2)$ \\
    Bring the blue cube to a position above of and close to the box.
    \item \textit{INSIDE(R, BOX)} = $inside(R, BOX)$ \\
    Place the red cube inside the box.
    \item \textit{INSIDE(B, BOX)} = $inside(R, BOX)$ \\
    Place the blue cube inside the box.
    \item \textit{INSIDE(ALL, BOX)} = $INSIDE(R, BOX) * INSIDE(B, BOX)$ \\
    Place the all cubes inside the box.
\end{itemize}

The sparse reward \textit{inside(A, BOX)} is given by comparing the bounds of the object A and the box. If the bounding box of object A is completely within the box's bounds \textit{inside(A, BOX)} is $1$, otherwise \textit{inside(A, BOX)} is $0$.

\onecolumn

\subsection{Multitask Results \label{sec:completeResults}}

% \markus{position legend - hierarch are correct and check if names are correct for both baselines}

% \subsubsection{Pile1 -- All Tasks}

\begin{figure}[H]
    \centering
    \begin{tabular}{ccc}
    \includegraphics[width=.24\textwidth]{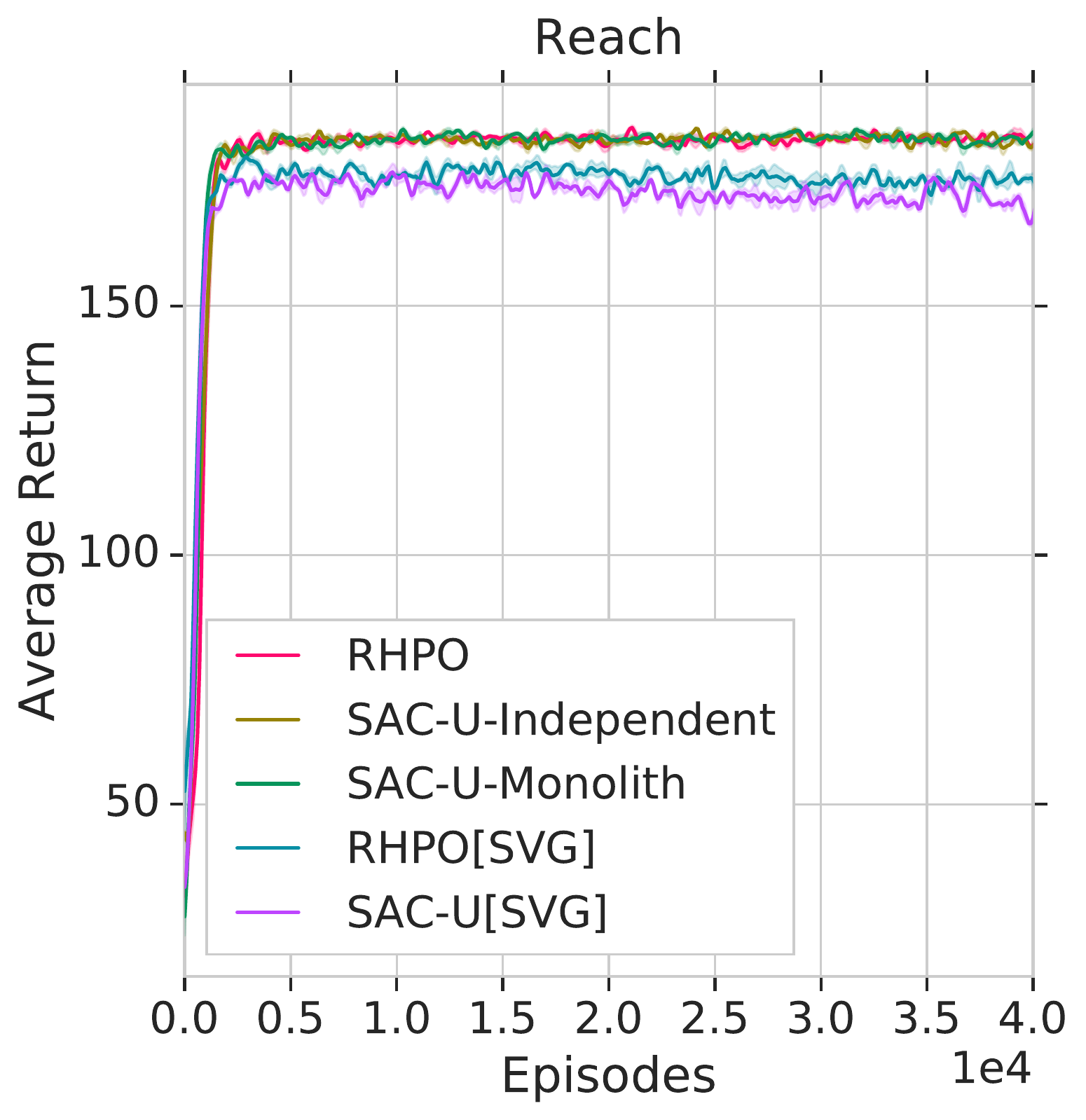}&  
    \includegraphics[width=.24\textwidth]{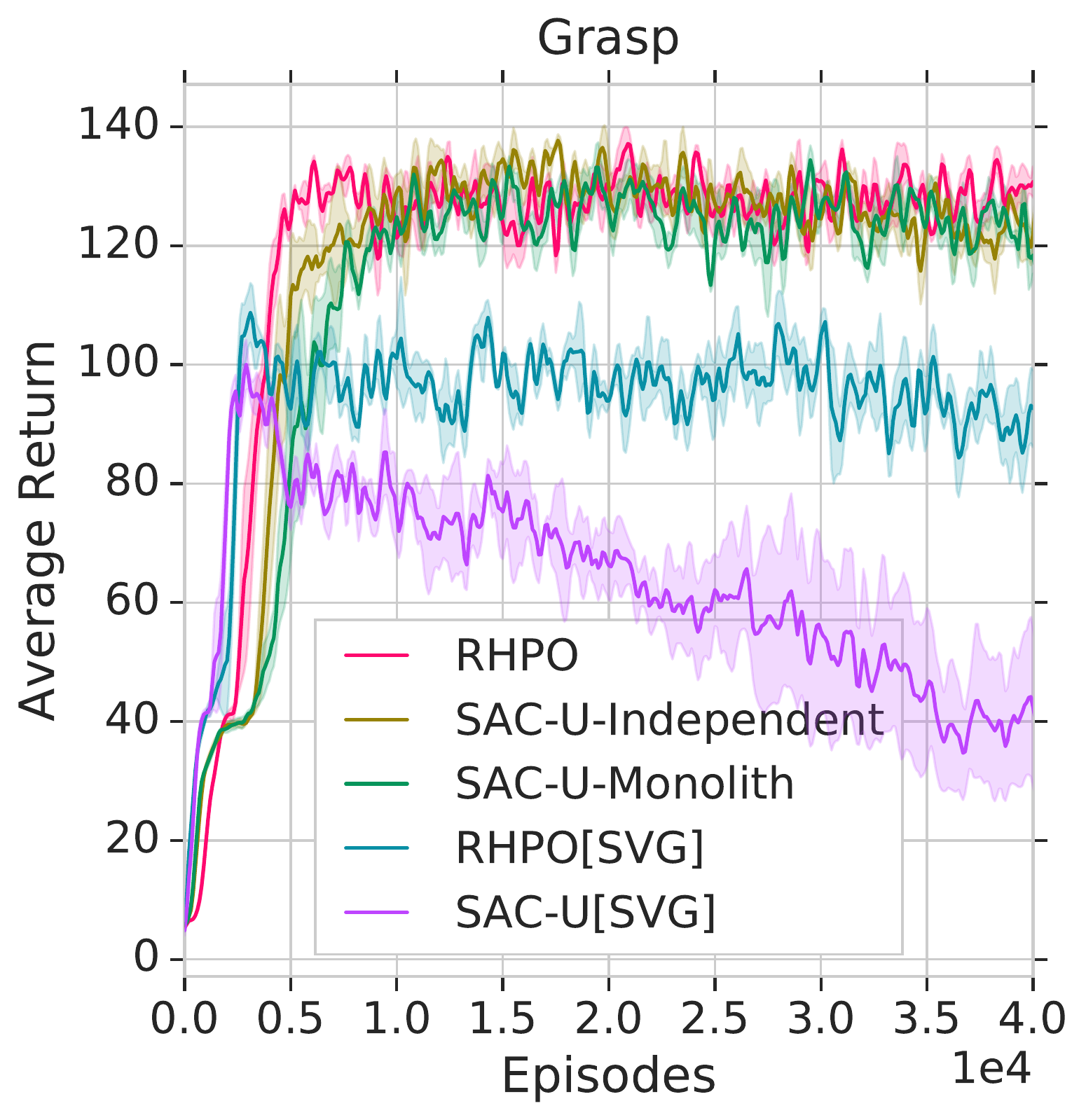}  & 
    \includegraphics[width=.24\textwidth]{figures/pile1/pile1_2.pdf} \\
    \includegraphics[width=.24\textwidth]{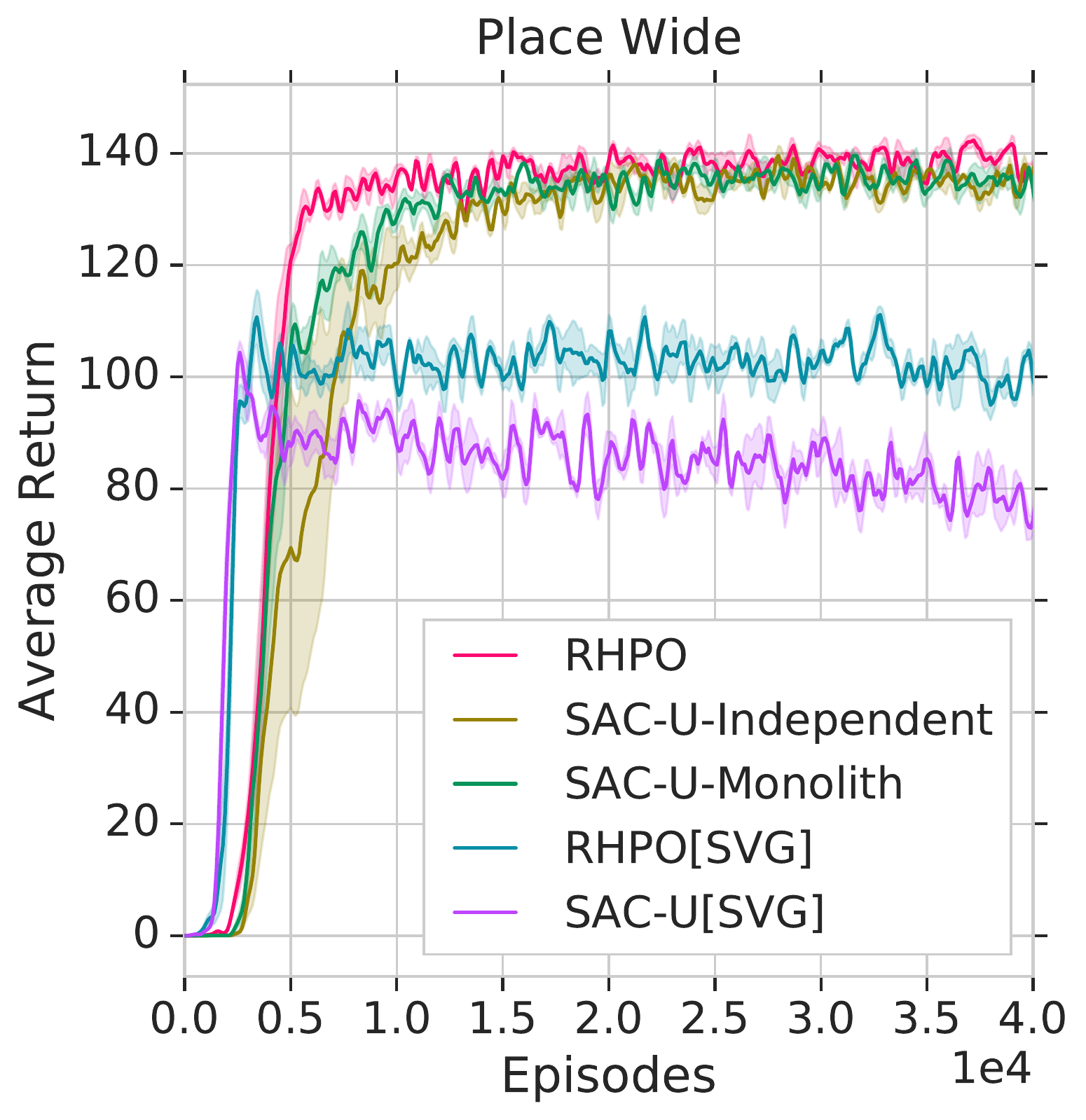}&  
    \includegraphics[width=.24\textwidth]{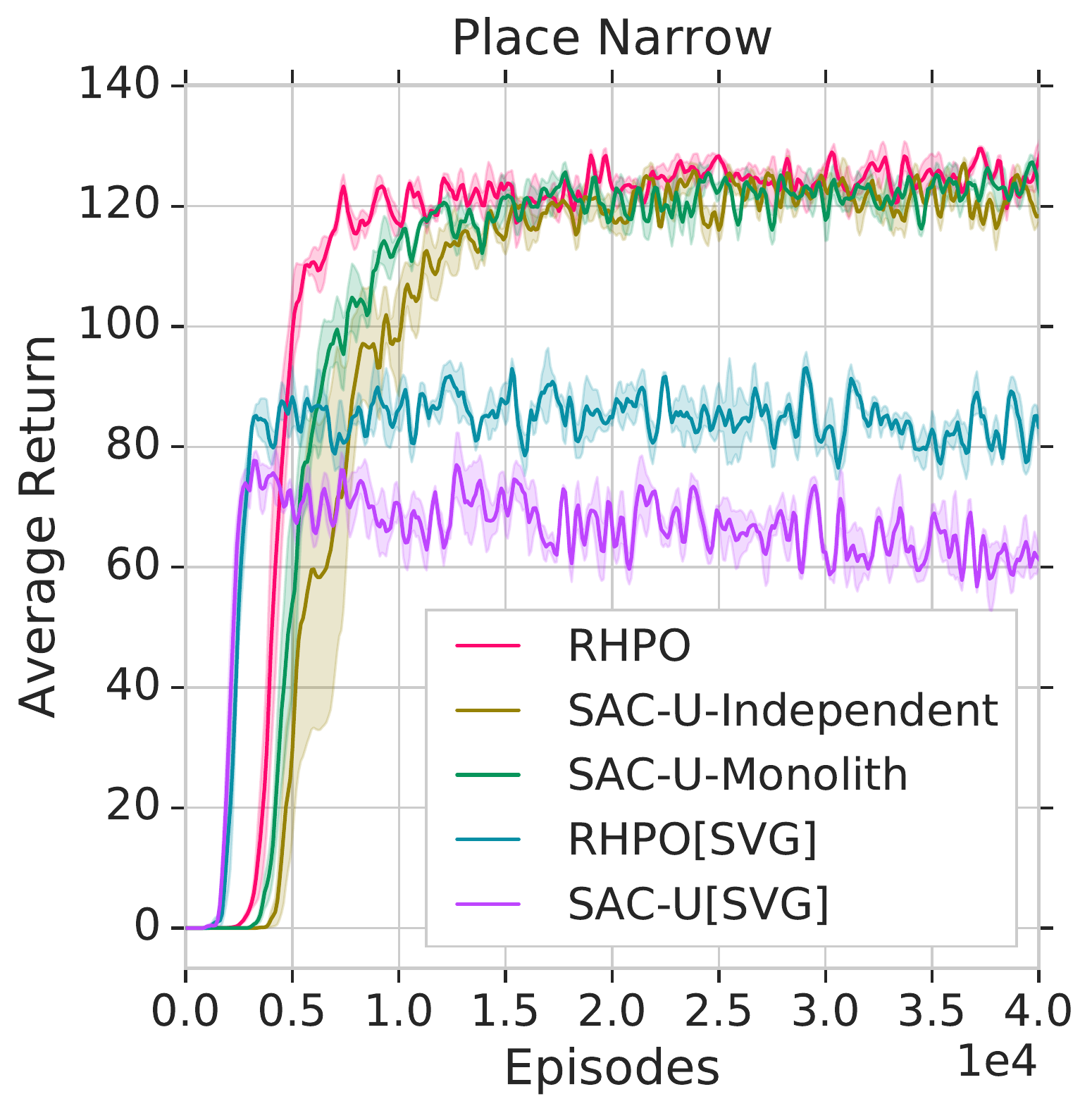}  & 
    \includegraphics[width=.24\textwidth]{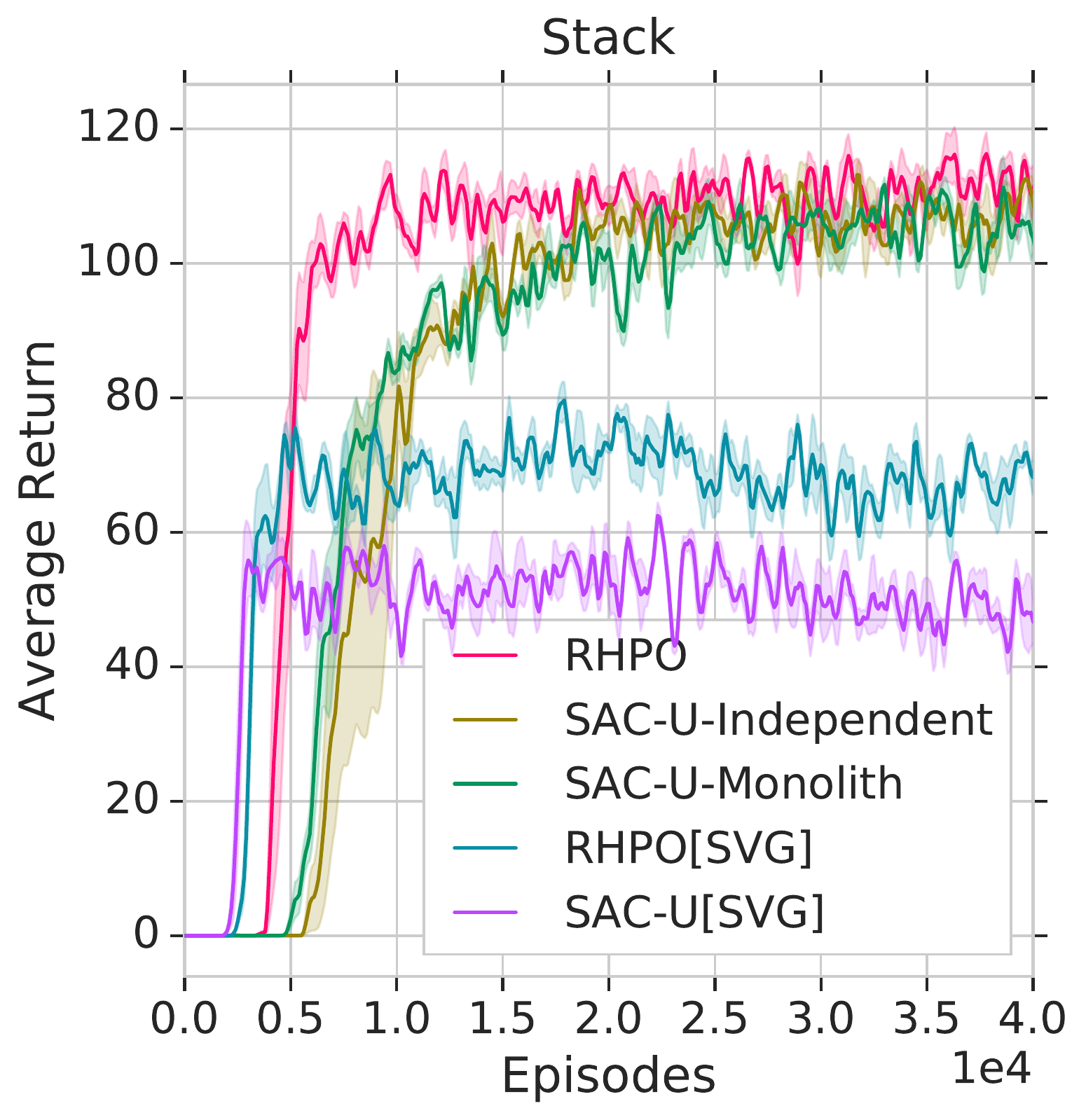} \\
    \includegraphics[width=.24\textwidth]{figures/pile1/pile1_6.pdf}
    \end{tabular}
    \caption{\small Pile1: Complete results for all tasks from the Pile1 domain. The dotted line represents standard SAC-U after the same amount of training. Results show that using hierarchical policy leads to best performance.}
    \label{fig:multitask_experiments_pile1}
\end{figure}

\newpage

% \subsubsection{Pile2 -- All Tasks}

\begin{figure}[H]
    \centering
    \begin{tabular}{ccc}
    \includegraphics[width=.25\textwidth]{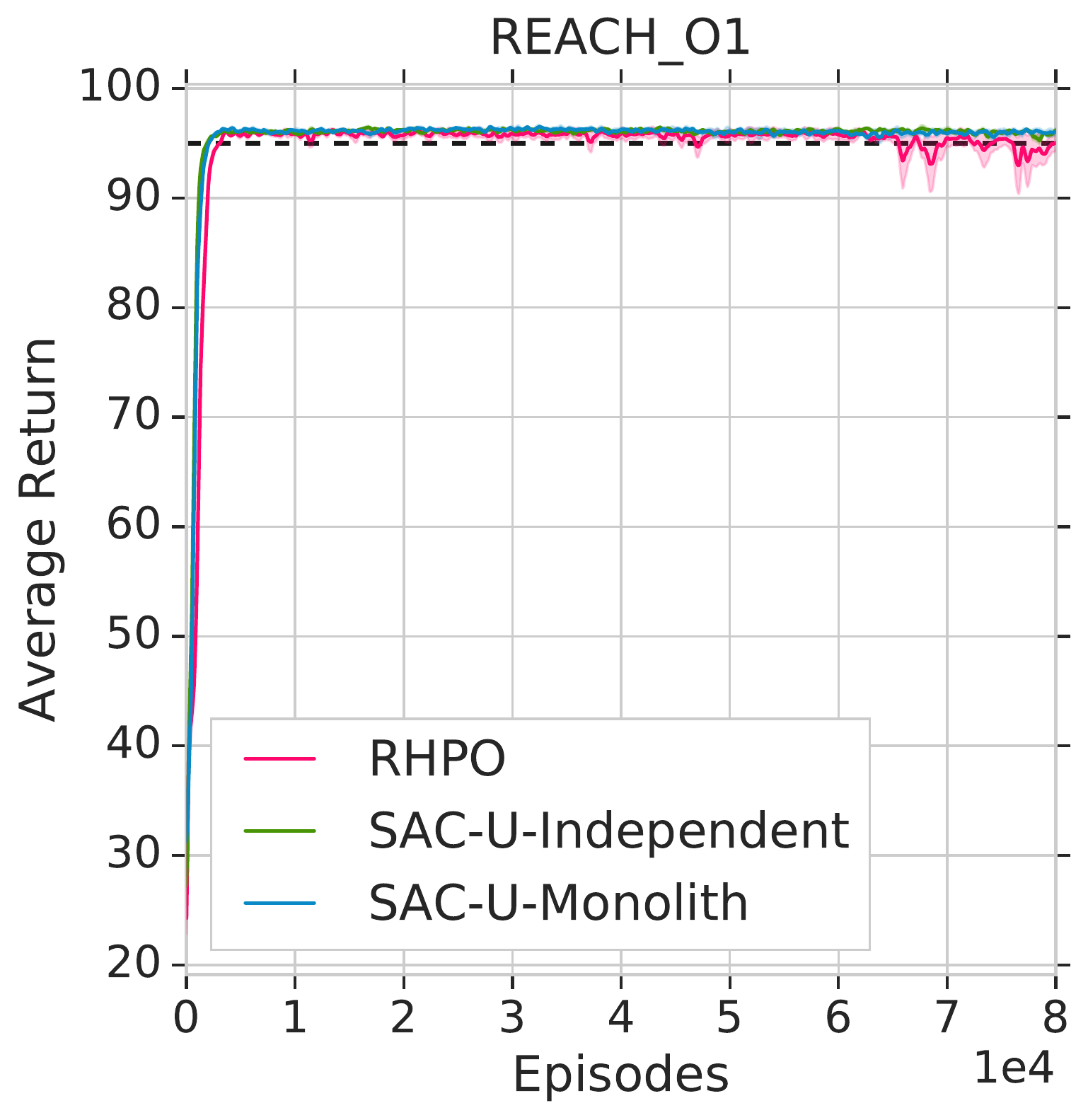}&  
    \includegraphics[width=.25\textwidth]{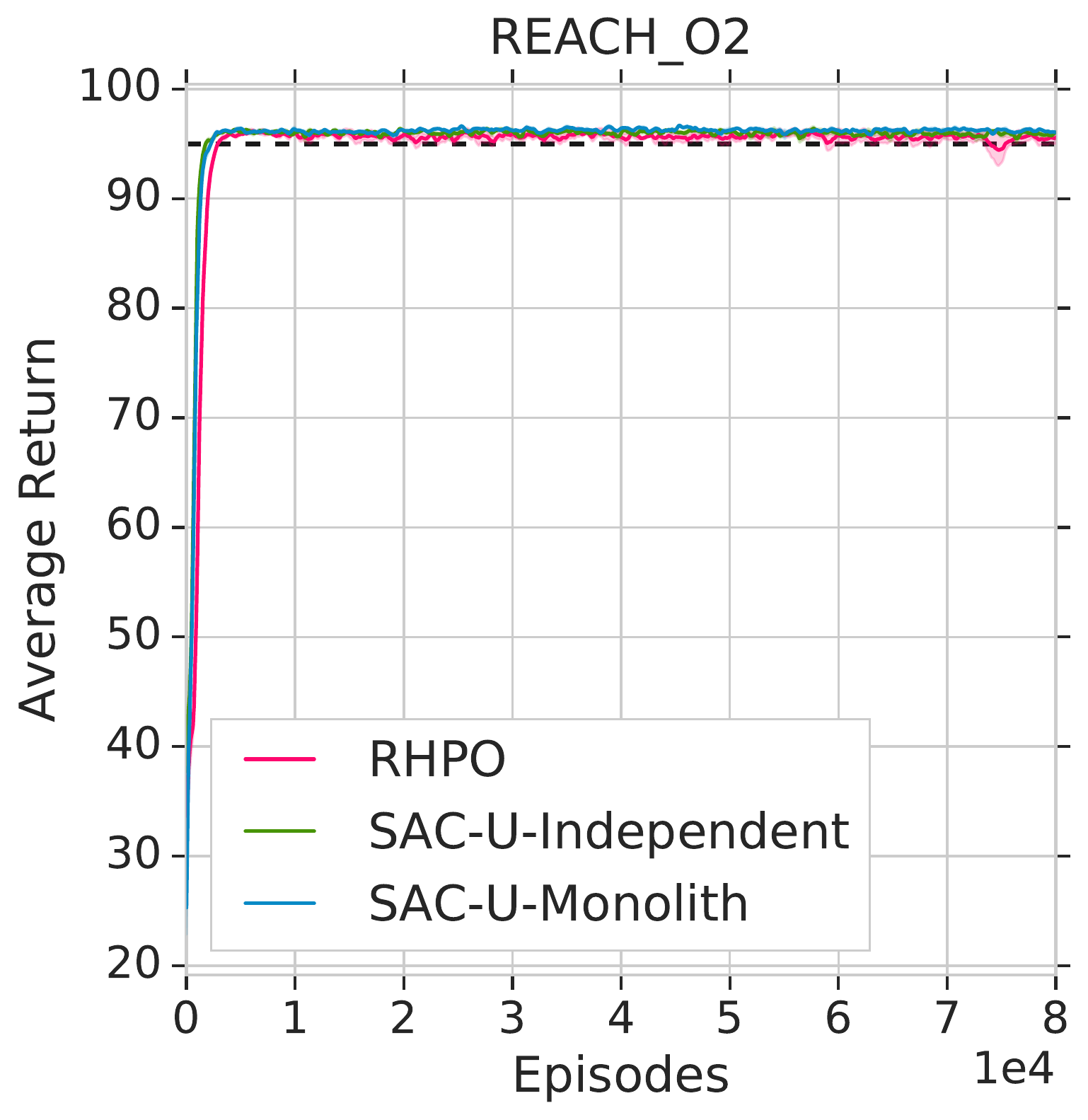}  & 
    \includegraphics[width=.25\textwidth]{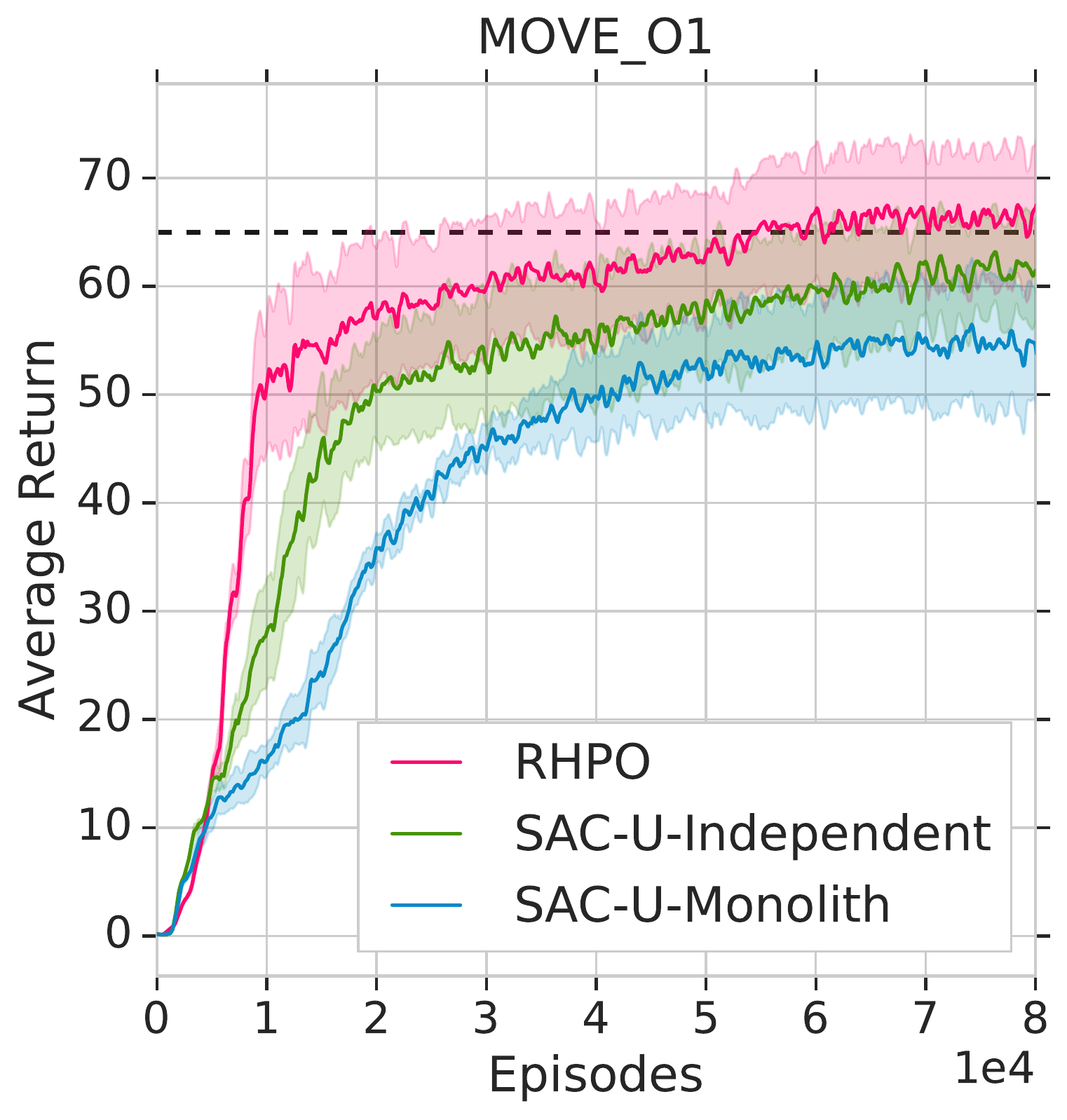} \\
    \includegraphics[width=.25\textwidth]{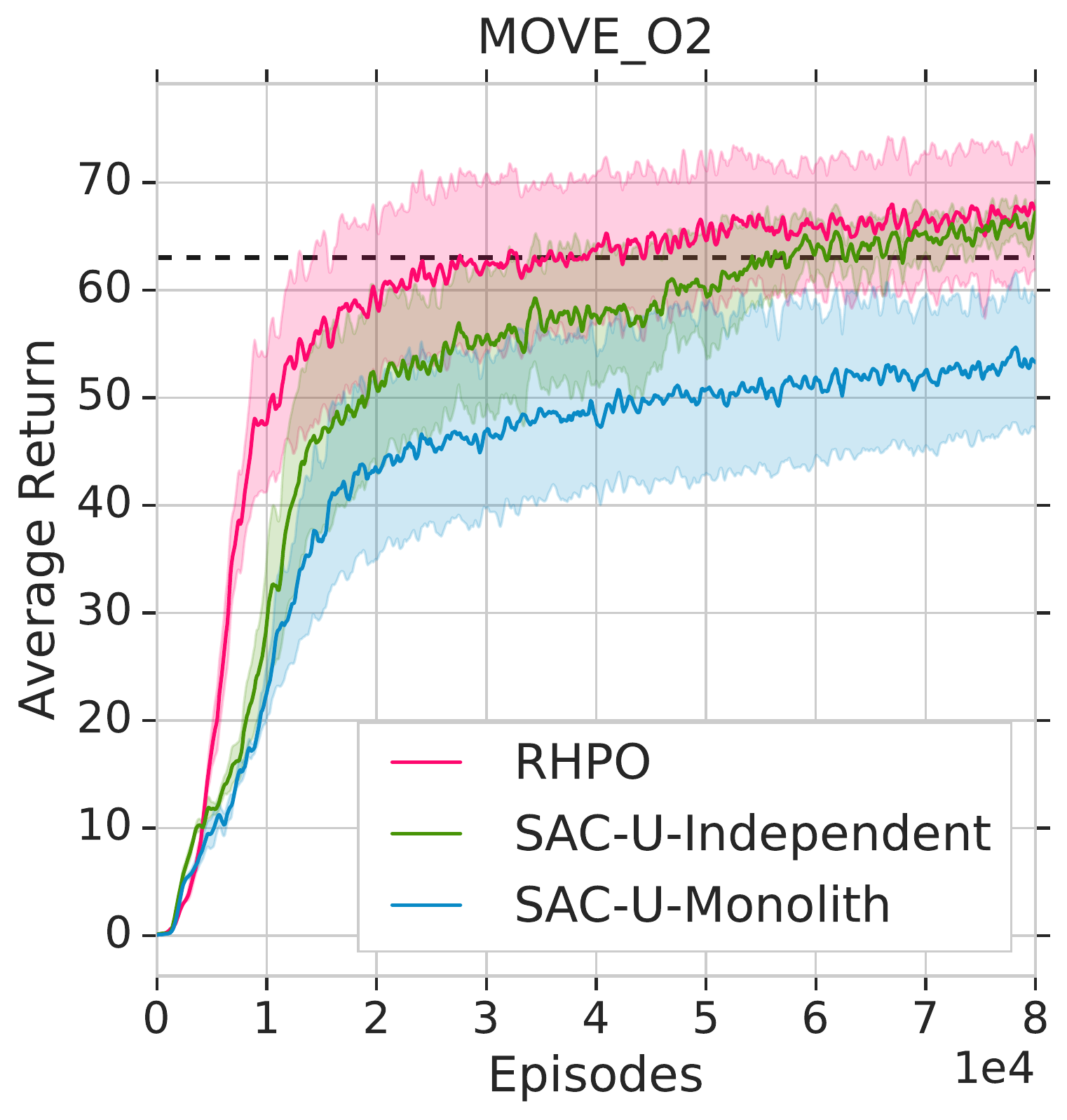}&  
    \includegraphics[width=.25\textwidth]{figures/pile2/pile2_4.pdf}  & 
    \includegraphics[width=.25\textwidth]{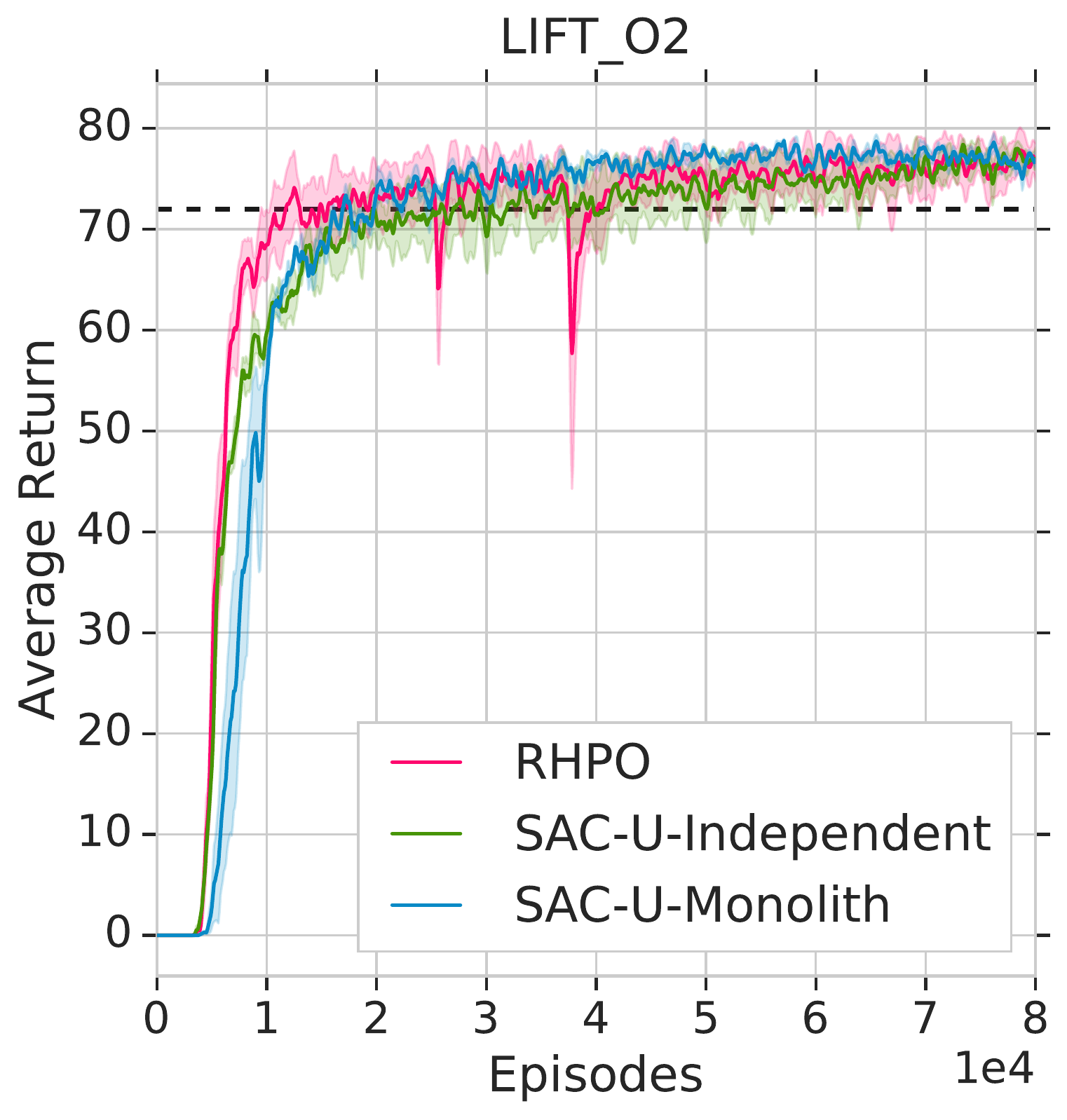}\\  
    \includegraphics[width=.25\textwidth]{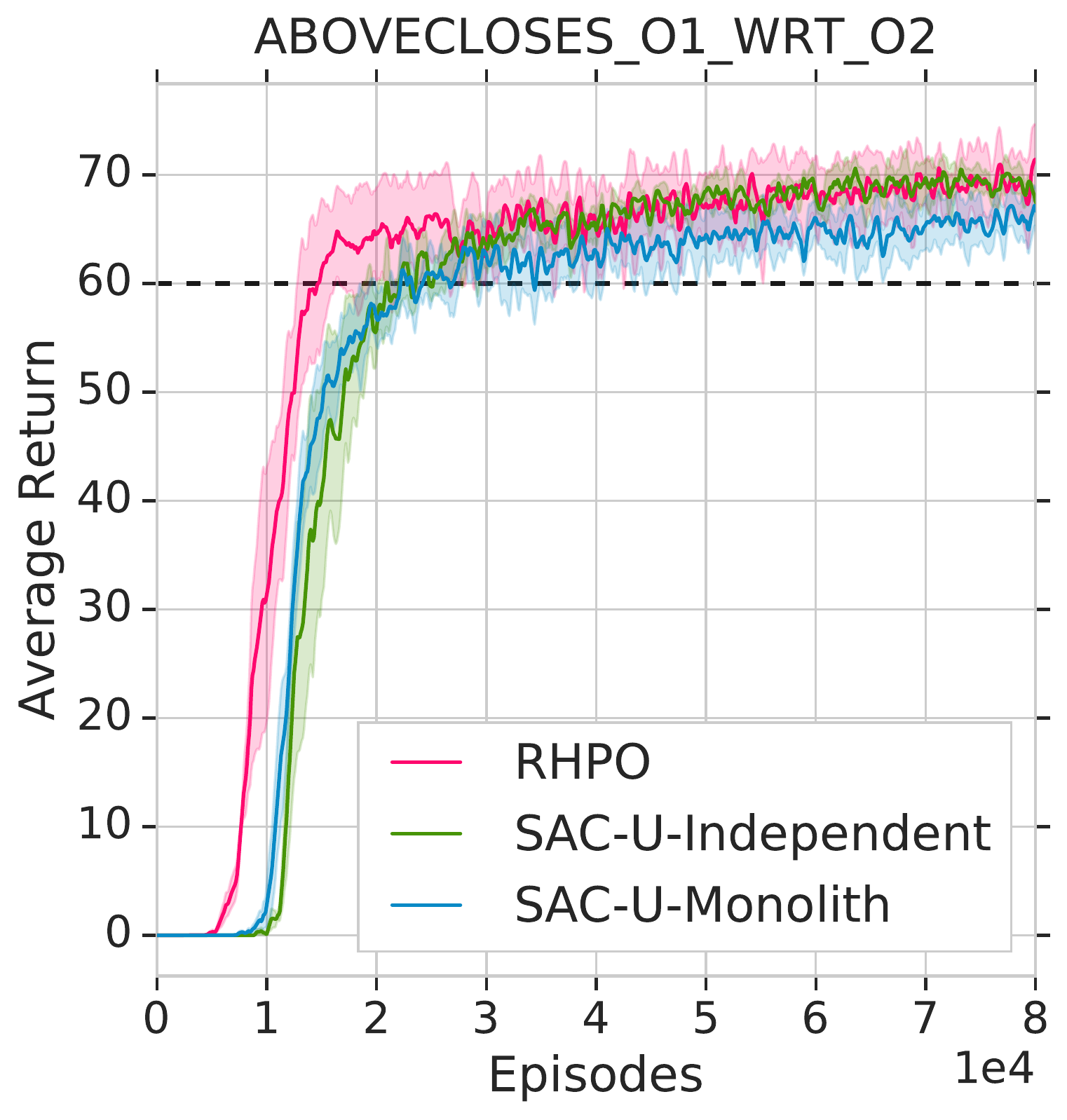}  & 
    \includegraphics[width=.25\textwidth]{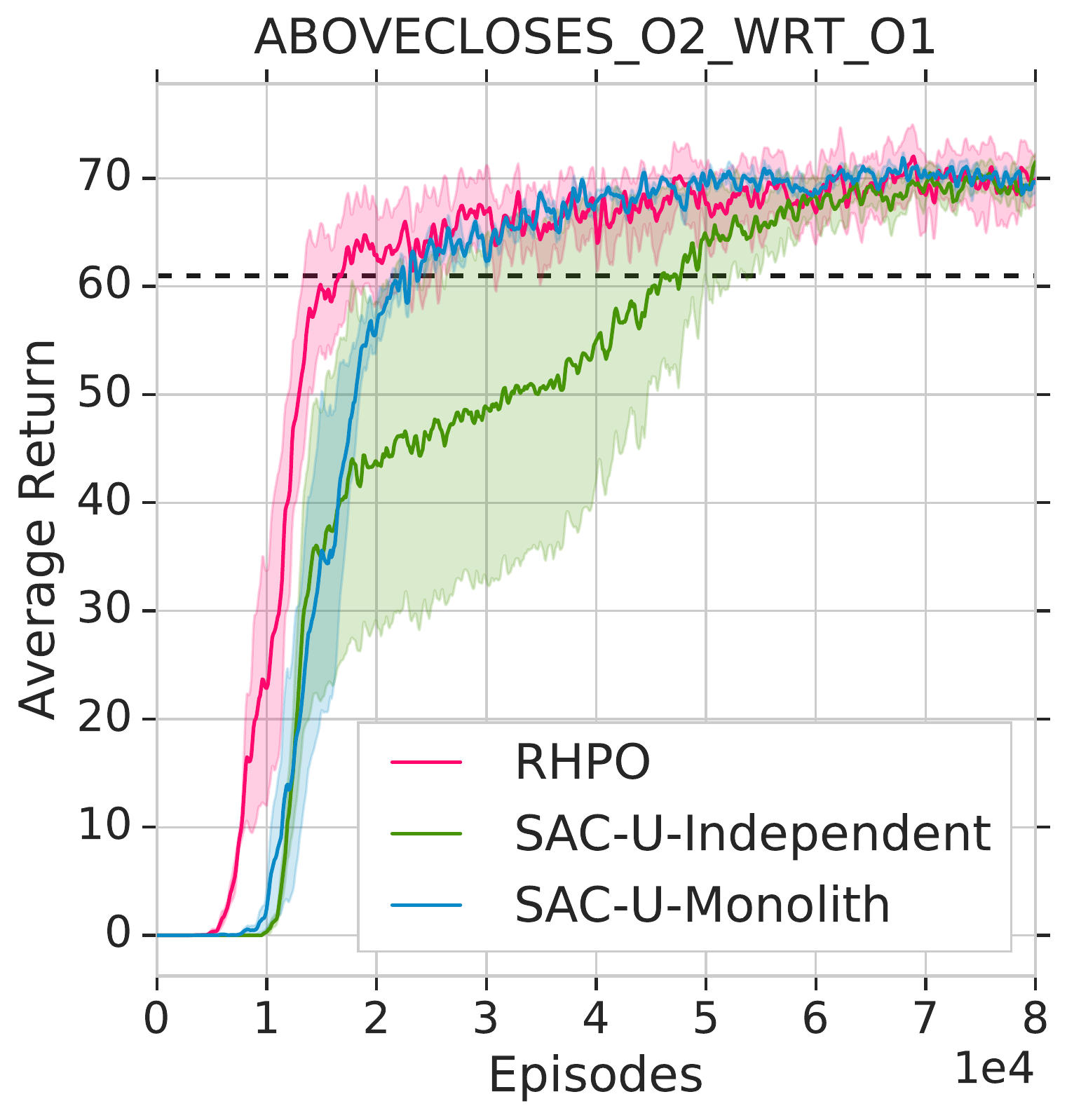} &
    \includegraphics[width=.25\textwidth]{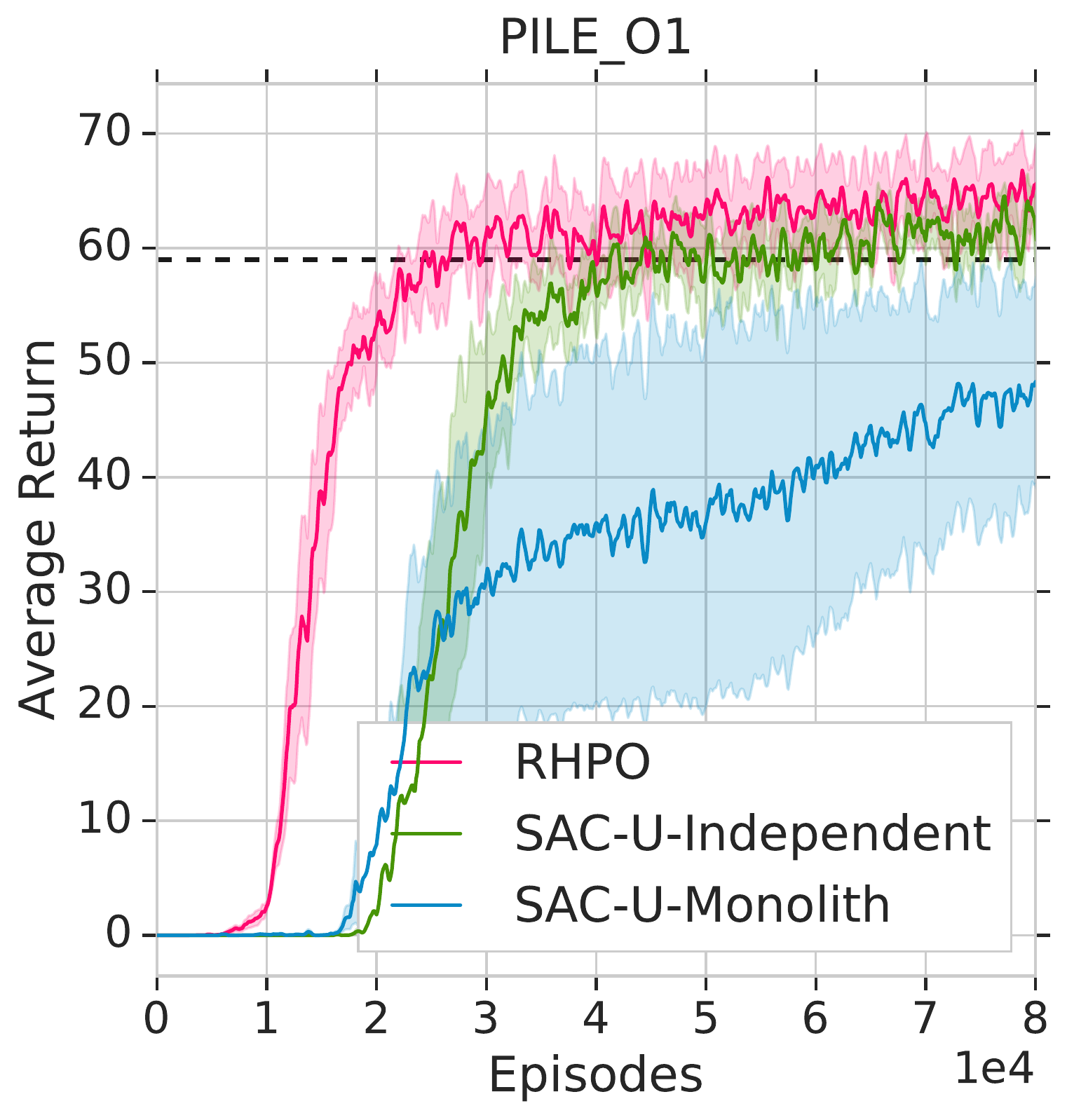}\\ 
    \includegraphics[width=.25\textwidth]{figures/pile2/pile2_9.pdf}   
    \end{tabular}
    \caption{\small Pile2: Complete results for all tasks from the Pile2 domain. Results show that using hierarchical policy leads to best performance. The dotted line represents standard SAC-U after the same amount of total training time.}
    \label{fig:multitask_experiments_pile2}
\end{figure}

\newpage

% \subsubsection{Cleanup2 -- All Tasks}

\begin{figure}[H]
    \centering
    \begin{tabular}{ccc}
    \includegraphics[width=.22\textwidth]{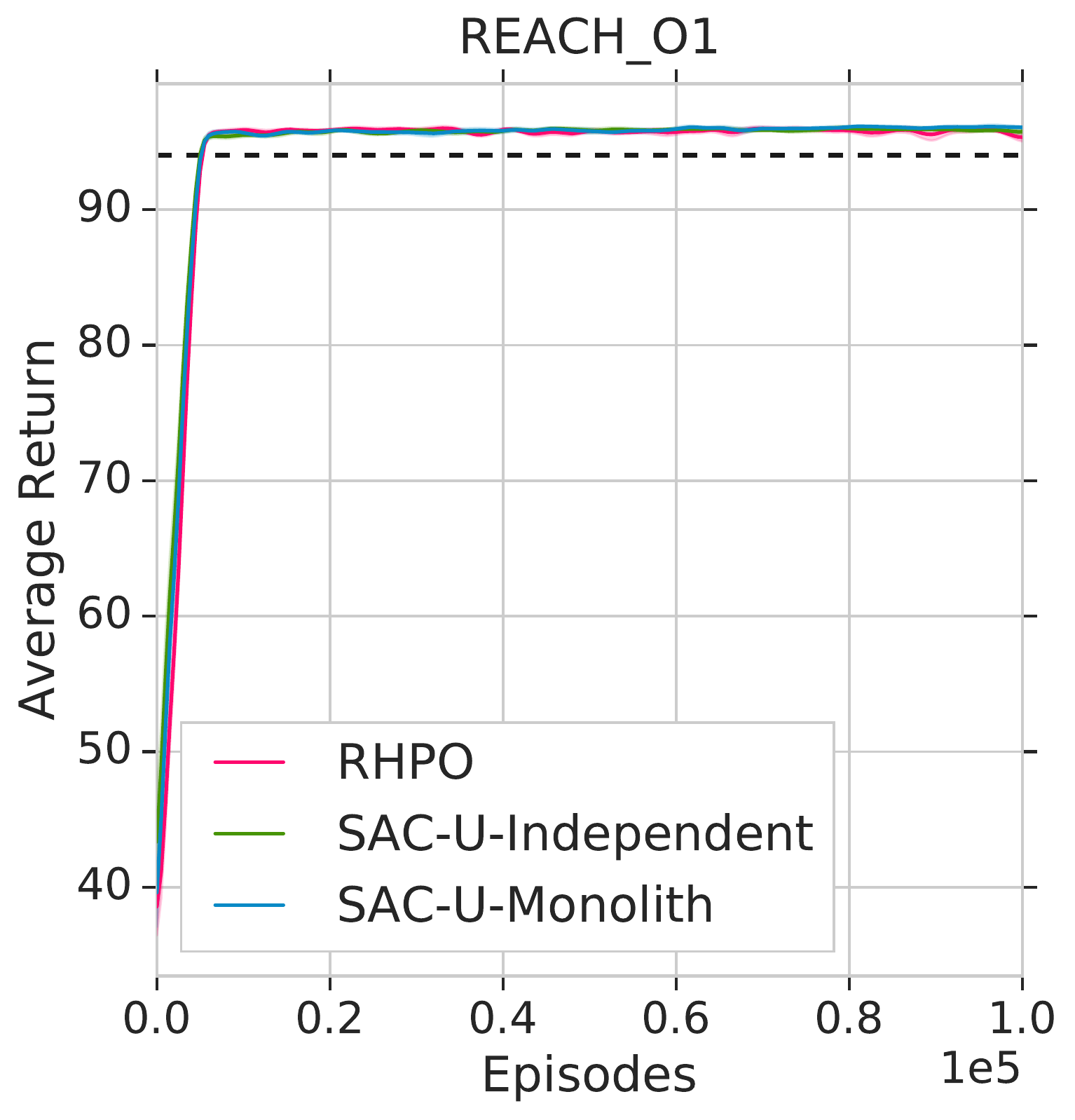}&  
    \includegraphics[width=.22\textwidth]{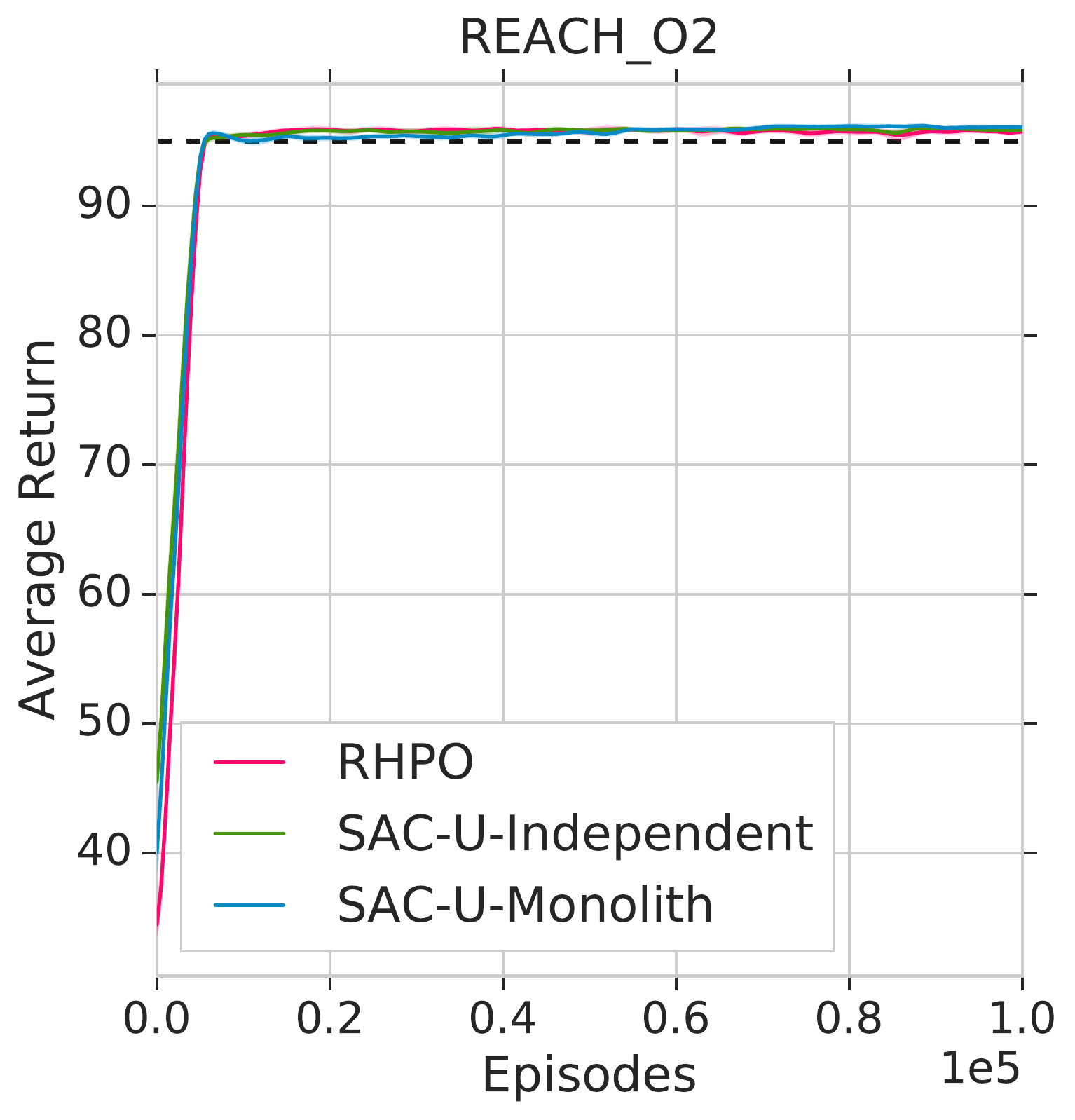}&
    \includegraphics[width=.22\textwidth]{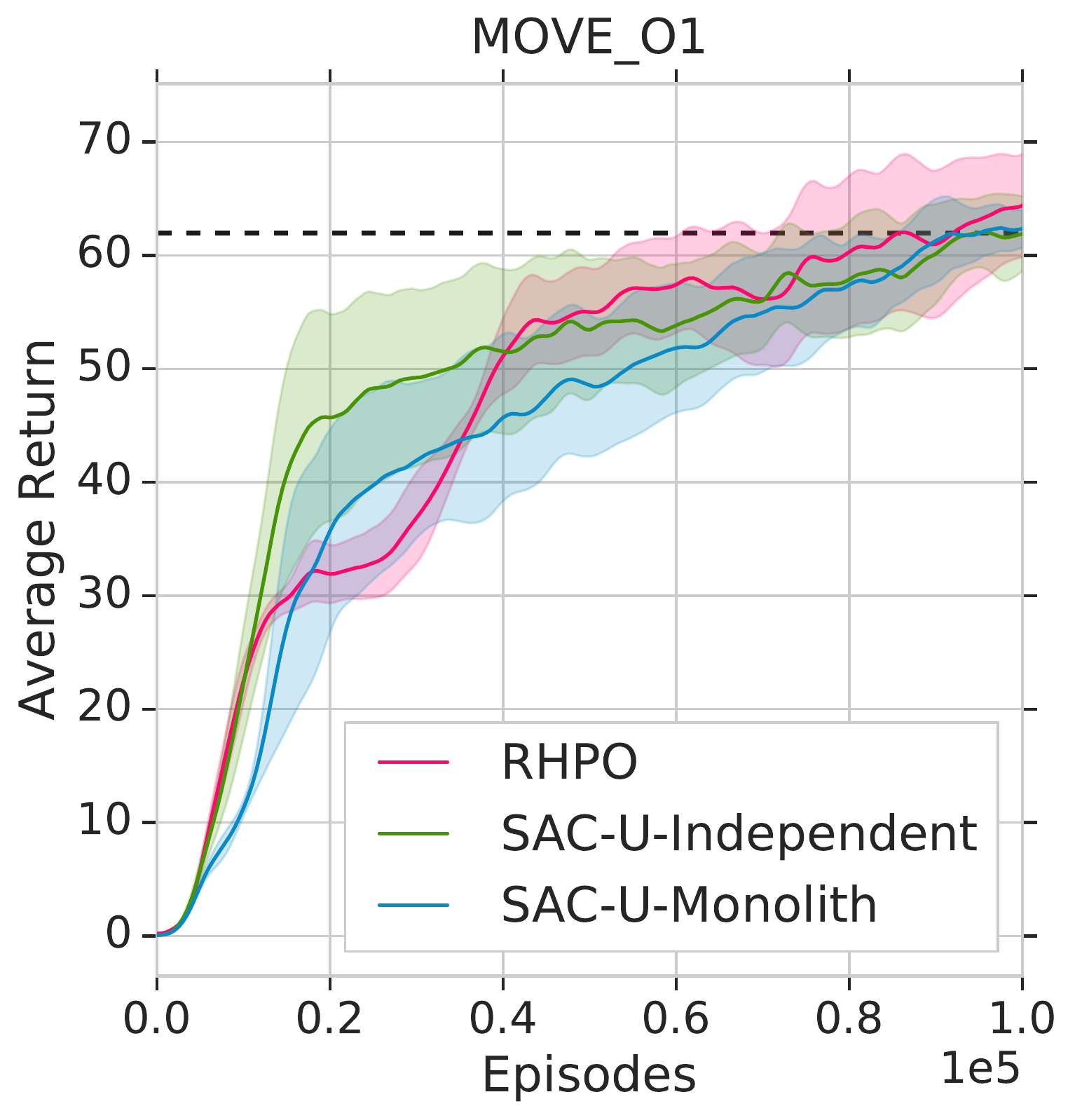} \\
    \includegraphics[width=.22\textwidth]{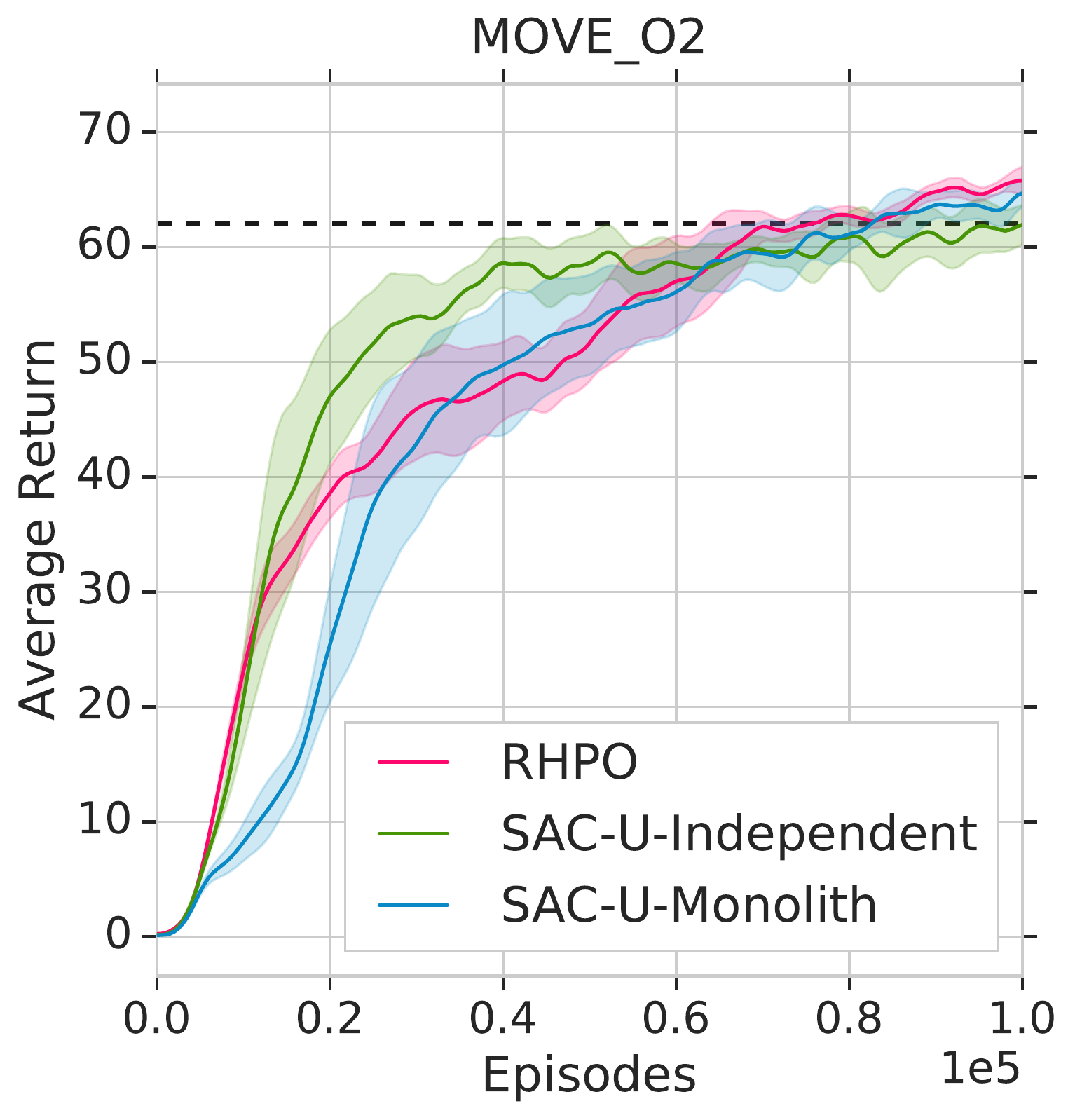}&  
    \includegraphics[width=.22\textwidth]{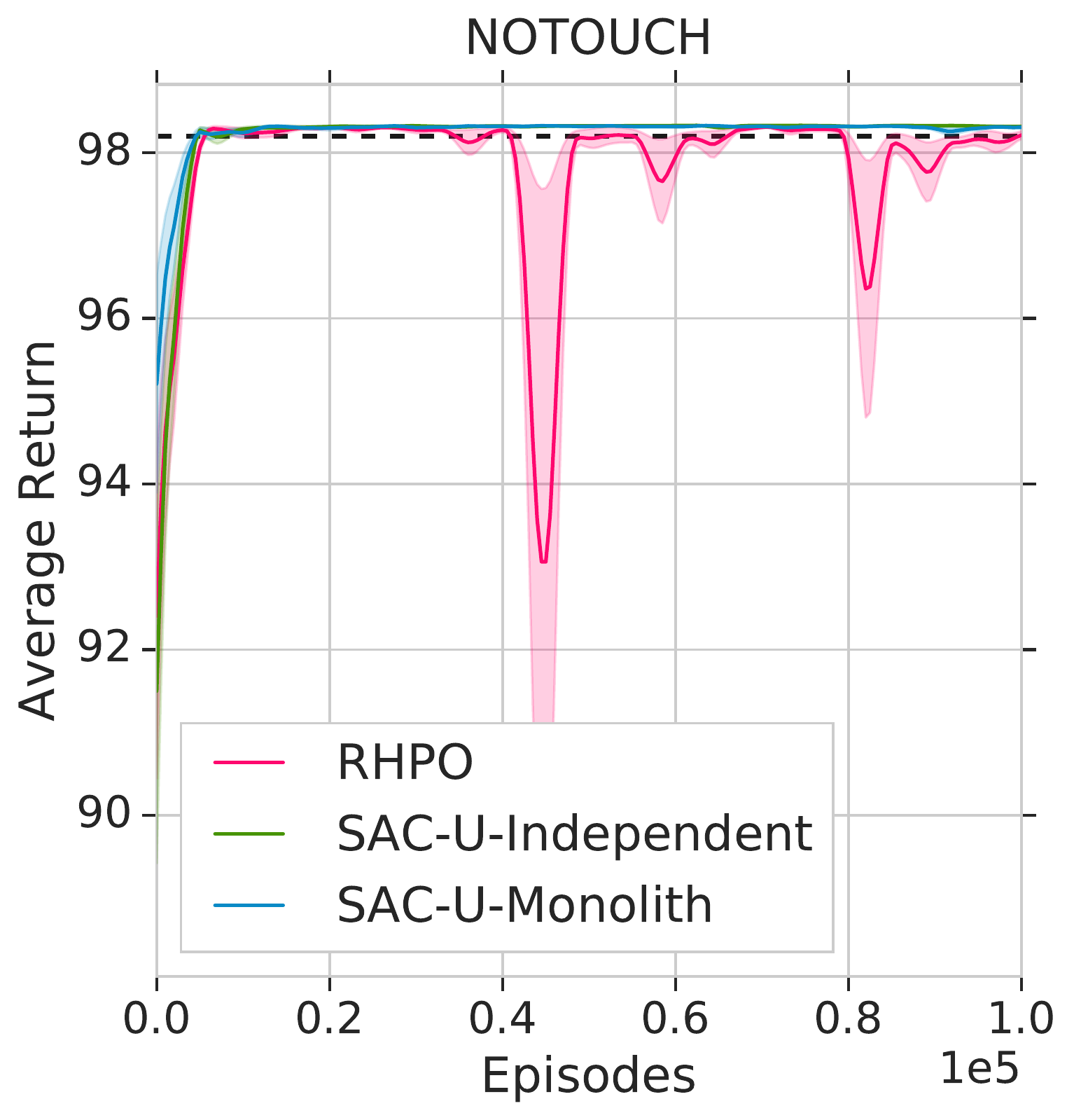}& 
    \includegraphics[width=.22\textwidth]{figures/cleanup2/cleanup2_5.pdf} \\
    \includegraphics[width=.22\textwidth]{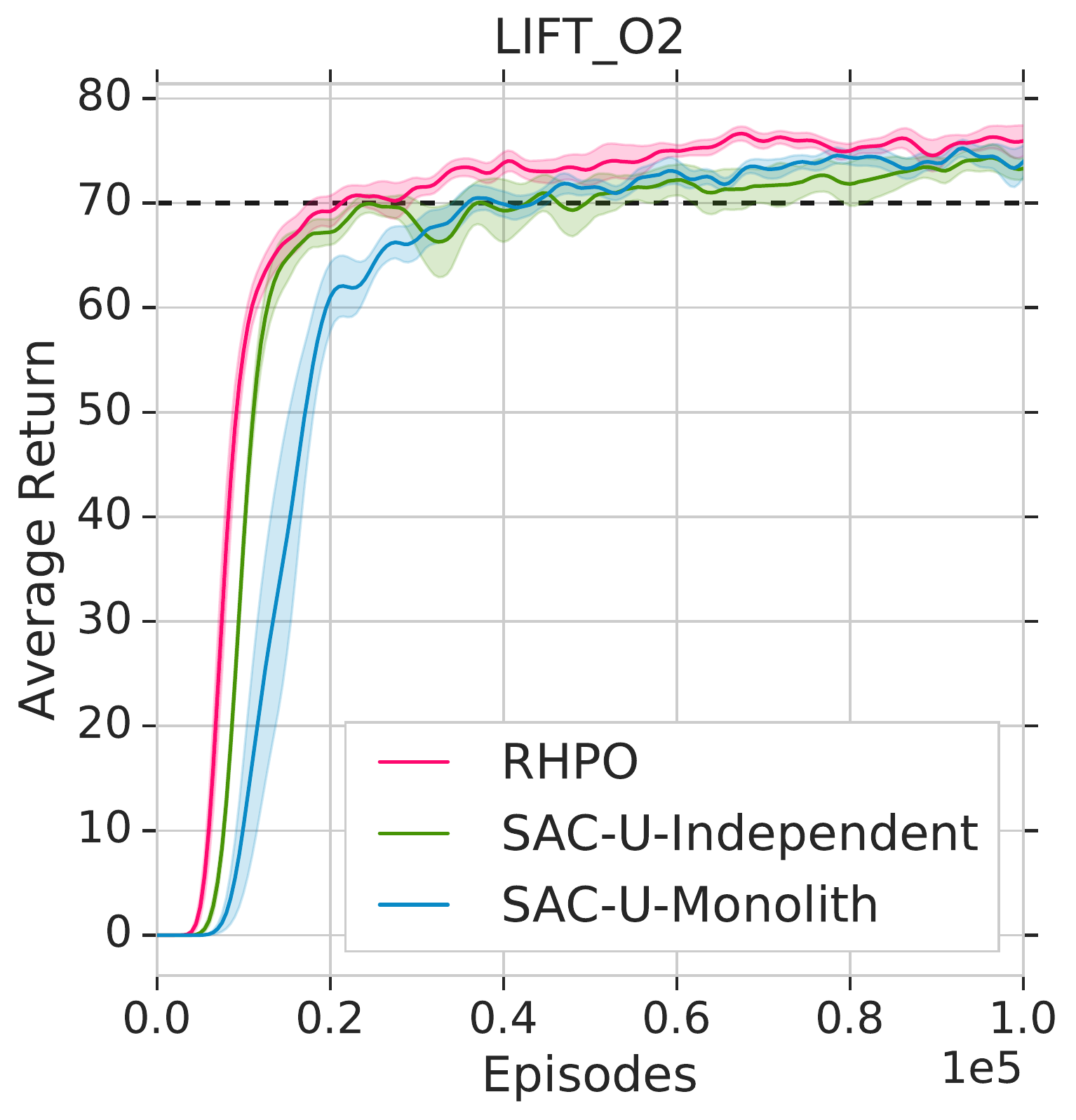}&  
    \includegraphics[width=.22\textwidth]{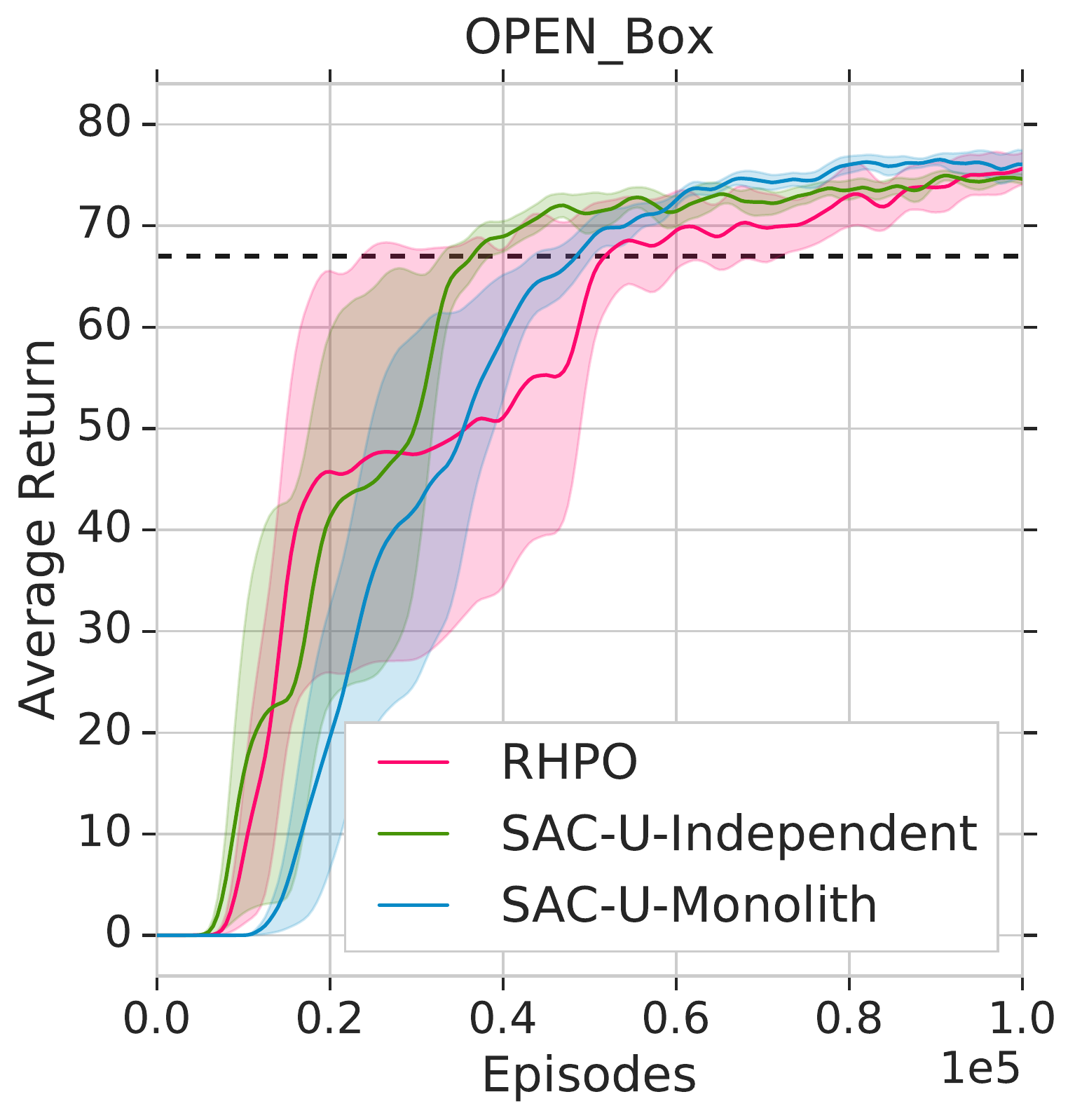}&
    \includegraphics[width=.22\textwidth]{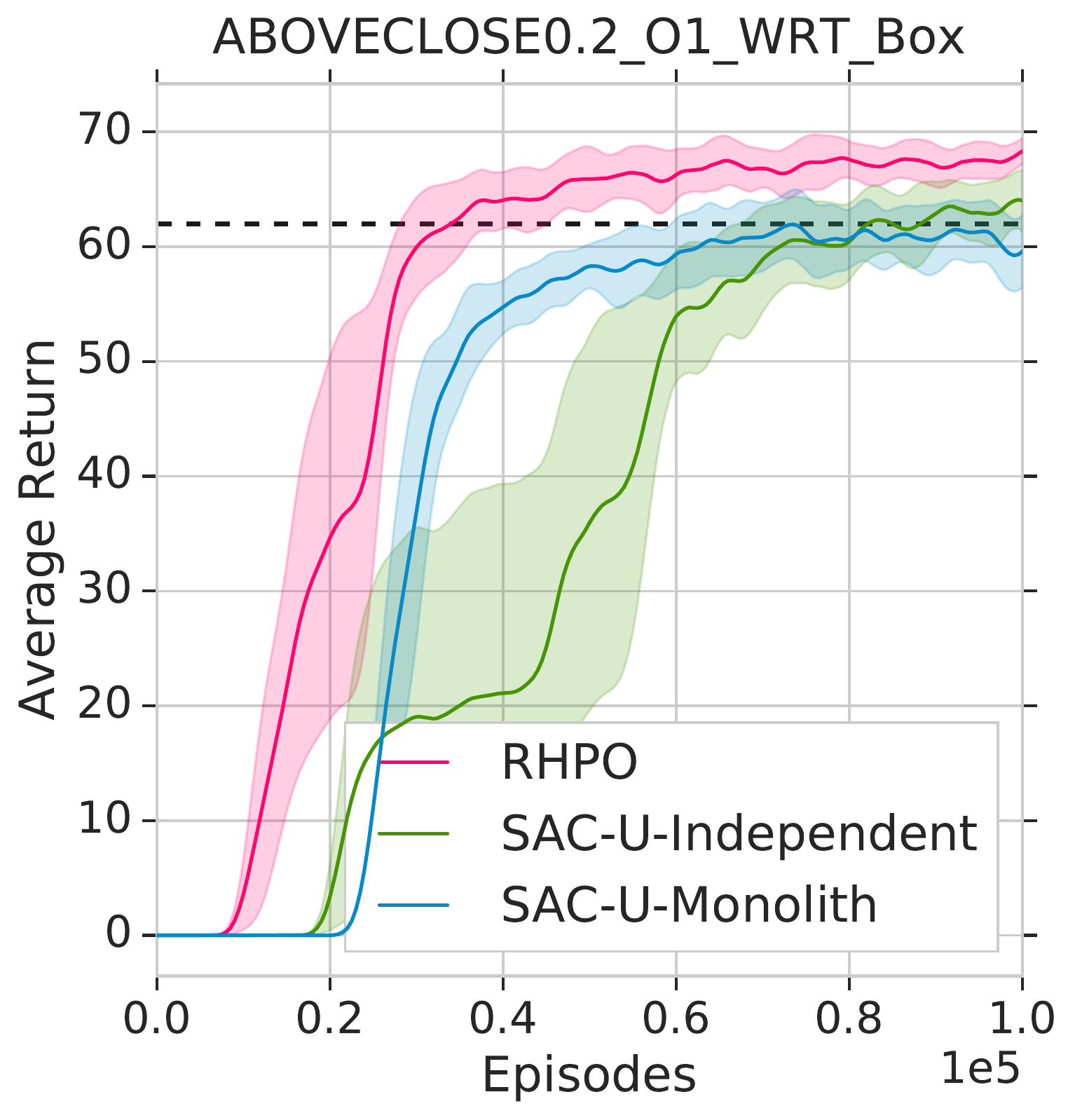} \\
    \includegraphics[width=.22\textwidth]{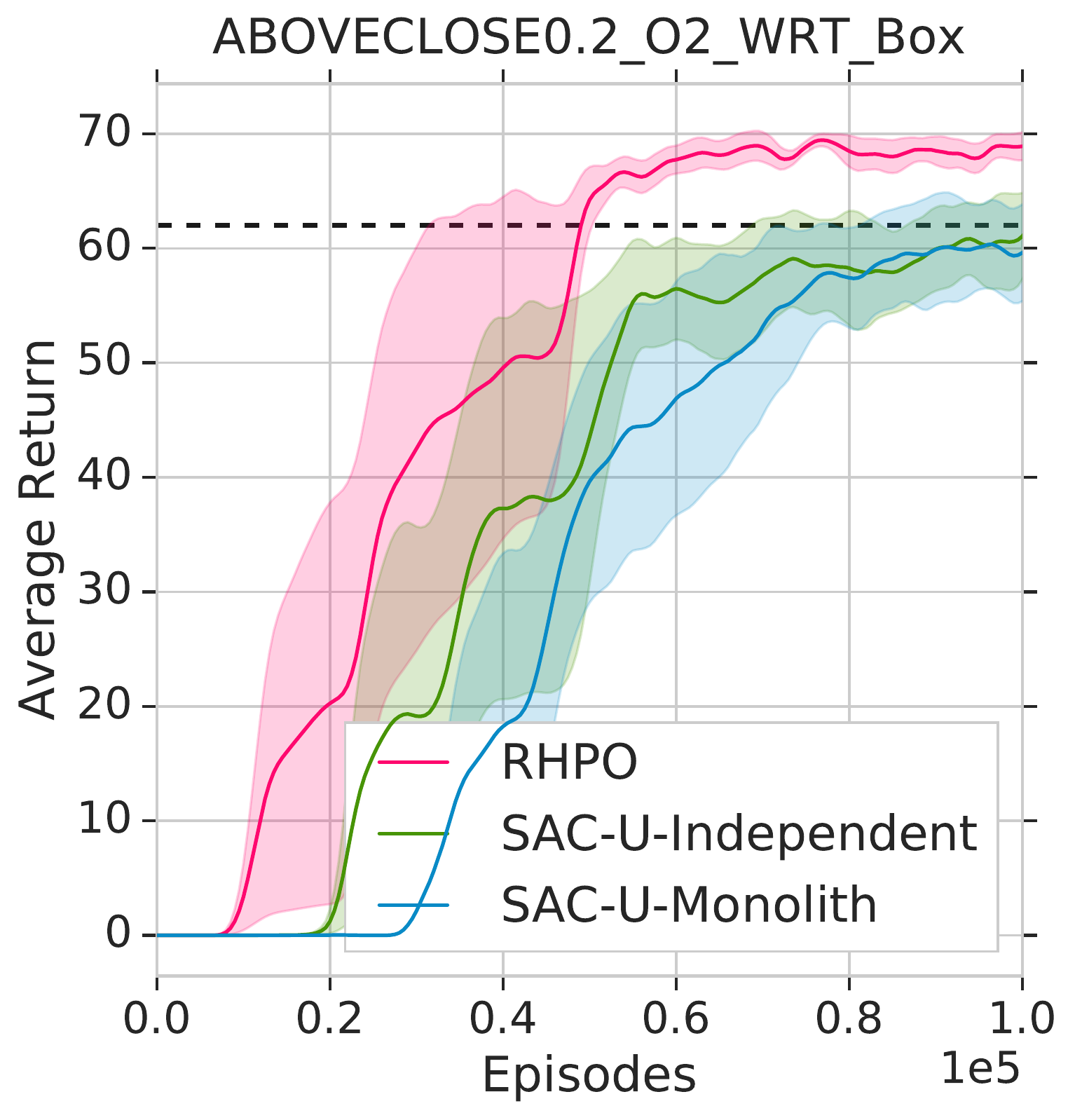}&  
    \includegraphics[width=.22\textwidth]{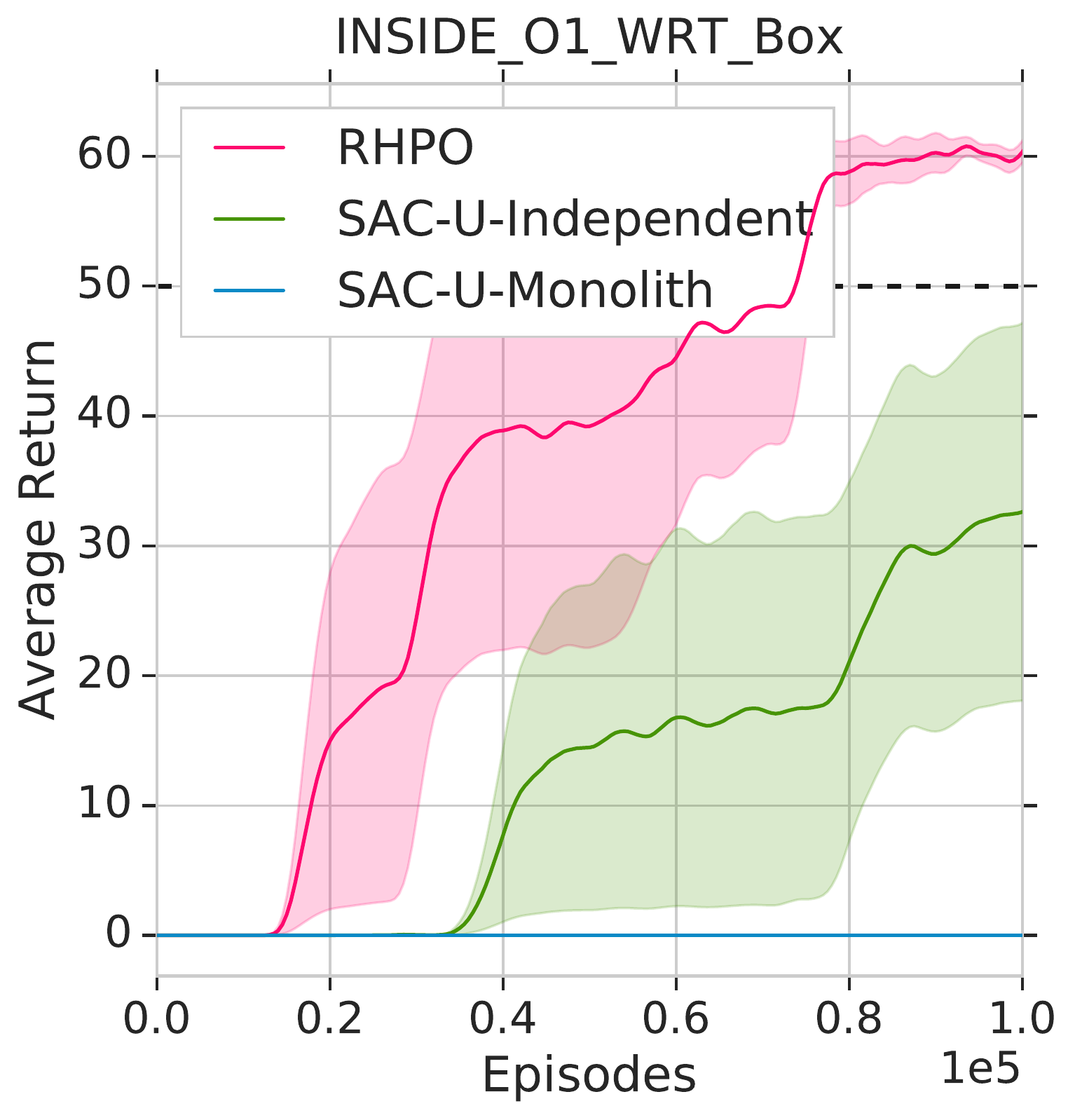}& 
    \includegraphics[width=.22\textwidth]{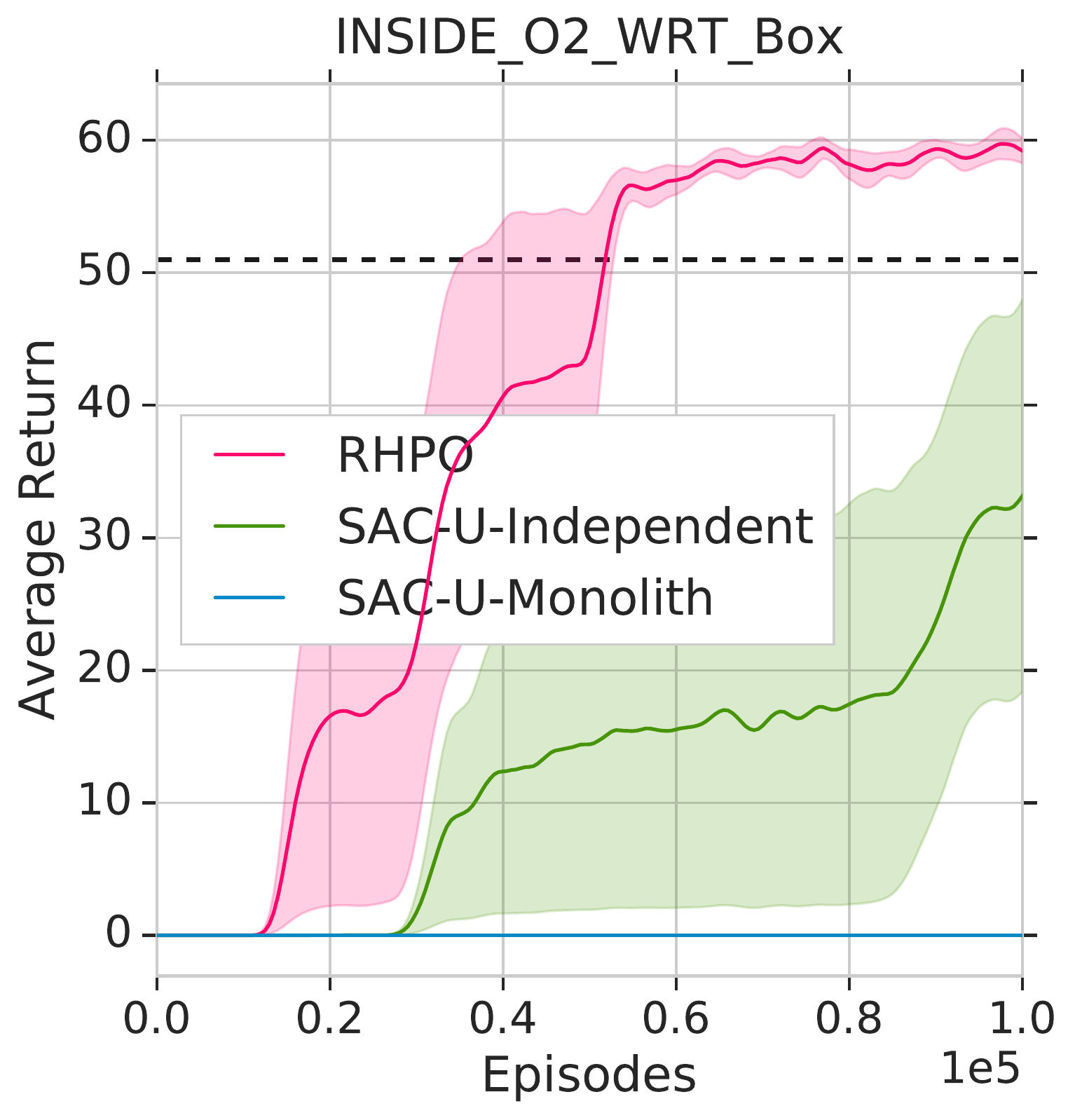} \\
    \includegraphics[width=.22\textwidth]{figures/cleanup2/cleanup2_12.pdf}& 
    \end{tabular}
    \caption{\small Cleanup2: Complete results for all tasks from the Cleanup2 domain. Results show that using hierarchical policy leads to best performance. The dotted line represents standard SAC-U after the same amount of total training time.}
    \label{fig:multitask_experiments_cleanup2}
\end{figure}

\newpage

% \subsection{Physical Robot Pile1 -- All Tasks}

\begin{figure}[H]
    \centering
    \begin{tabular}{ccc}
    \includegraphics[width=.3\textwidth]{figures/pile1real/real_pile_0.pdf}&
    \includegraphics[width=.3\textwidth]{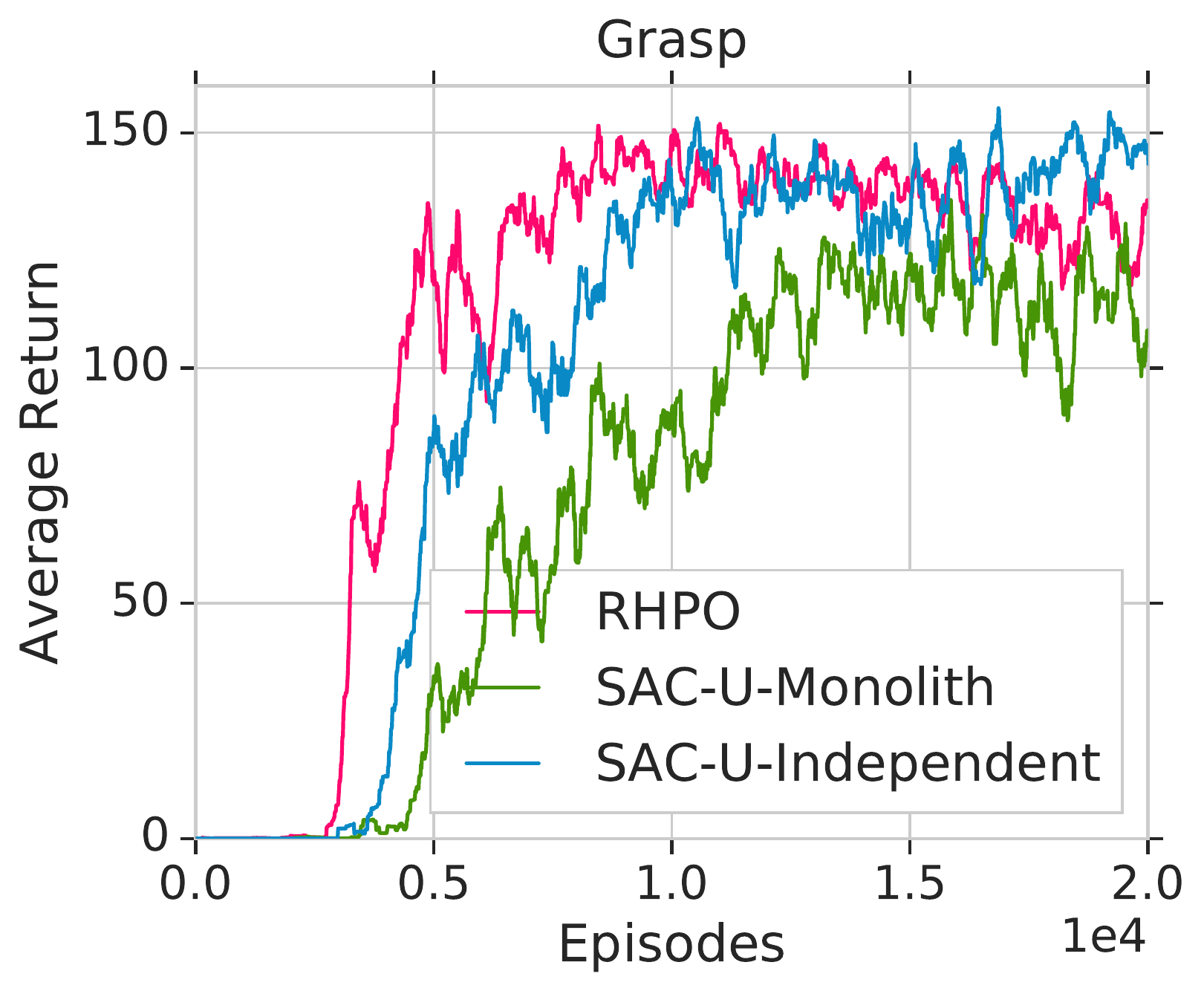}&
    \includegraphics[width=.3\textwidth]{figures/pile1real/real_pile_2.pdf} \\
    \includegraphics[width=.3\textwidth]{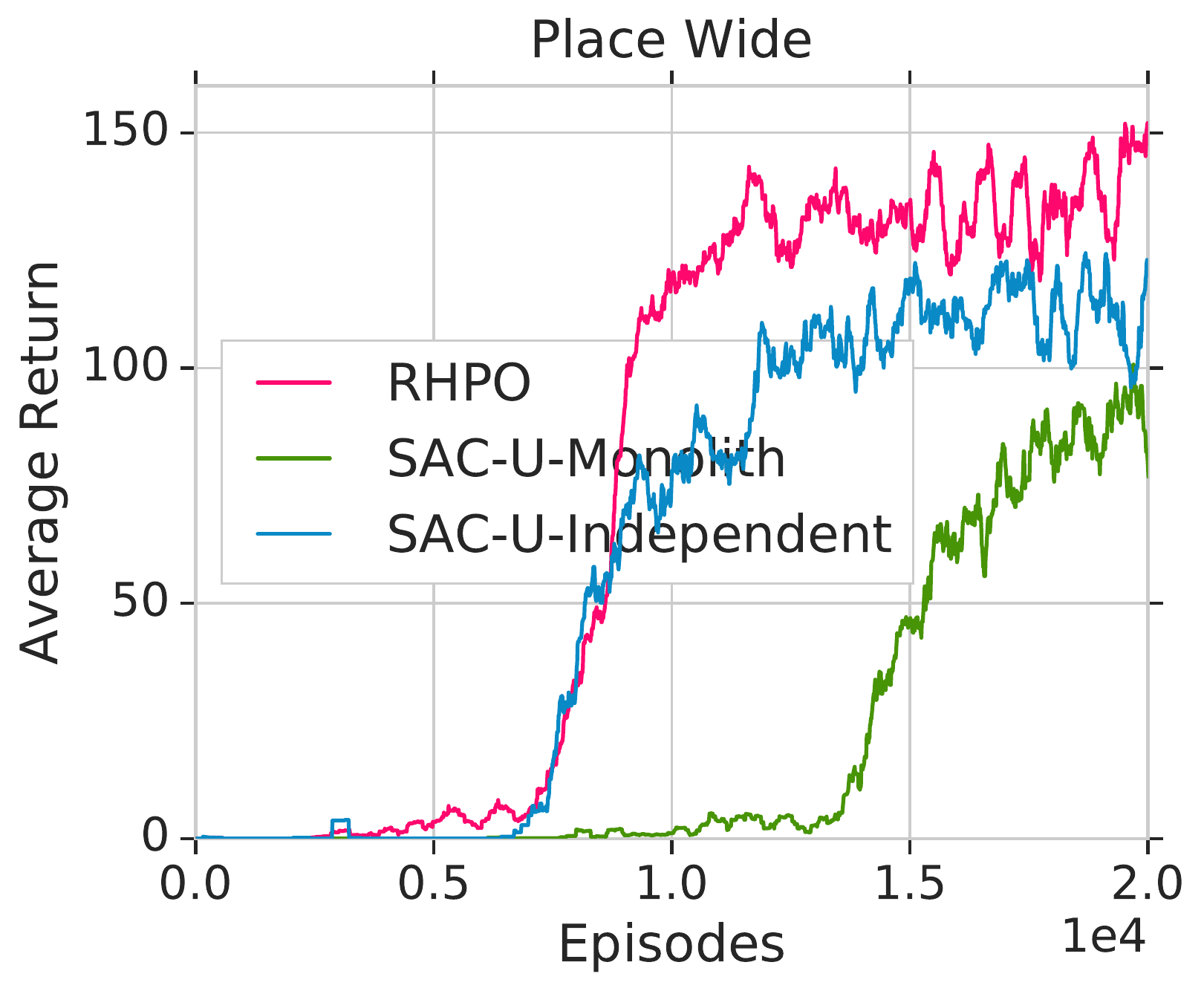}&
    \includegraphics[width=.3\textwidth]{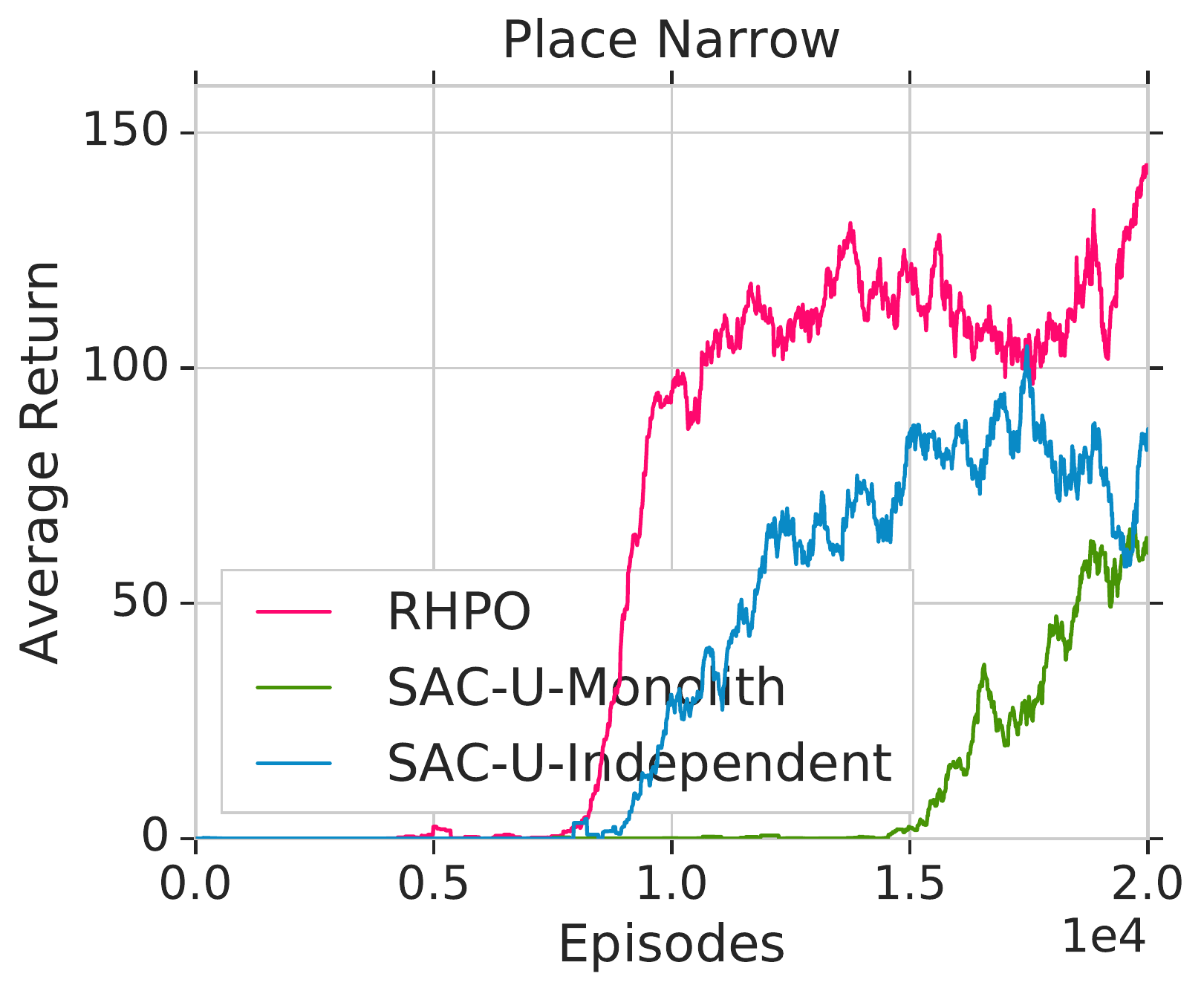}&
    \includegraphics[width=.3\textwidth]{figures/pile1real/real_pile_5.pdf} \\    
    \includegraphics[width=.3\textwidth]{figures/pile1real/real_pile_6.pdf}
    \end{tabular}
    \caption{\small Physical Robot Pile1: Complete results for all tasks on the real robot Pile1 domain. Results show that using hierarchical policy leads to best performance.}
    \label{fig:multitask_experiments_real}
\end{figure}

% \newpage

\subsection{Additional Multitask Ablations}

\subsubsection{Importance of Regularization} \label{app:regularisation}
Coordinating convergence progress in hierarchical models can be challenging but can be effectively moderated by the KL constraints. We perform an ablation study varying the strength of KL constraints on the high-level controller between prior and the current policy during training -- demonstrating a range of possible degenerate behaviors.

\begin{figure}[H]
    \centering
    \begin{tabular}{ccc}
    \includegraphics[width=.3\textwidth]{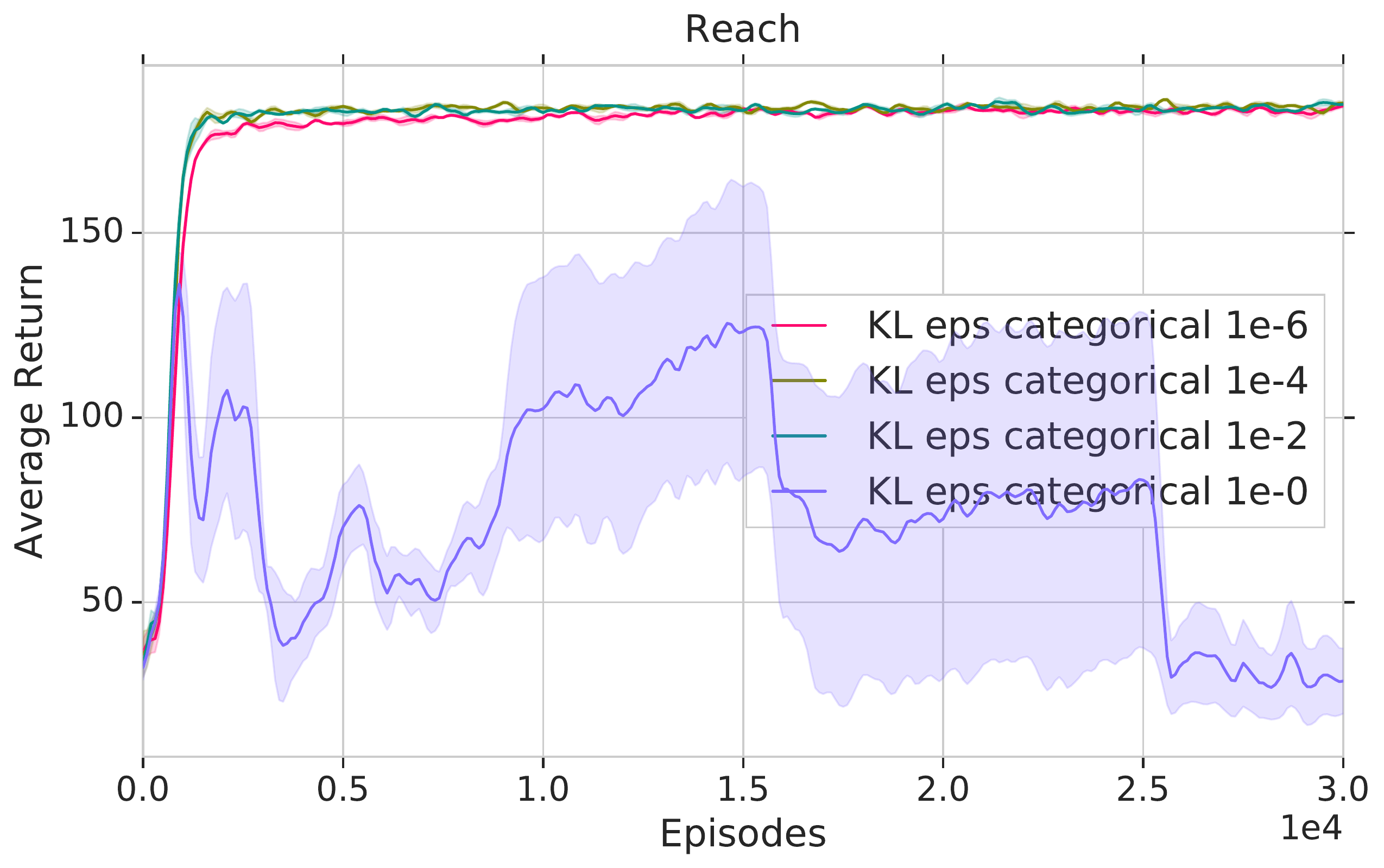}&  
    \includegraphics[width=.3\textwidth]{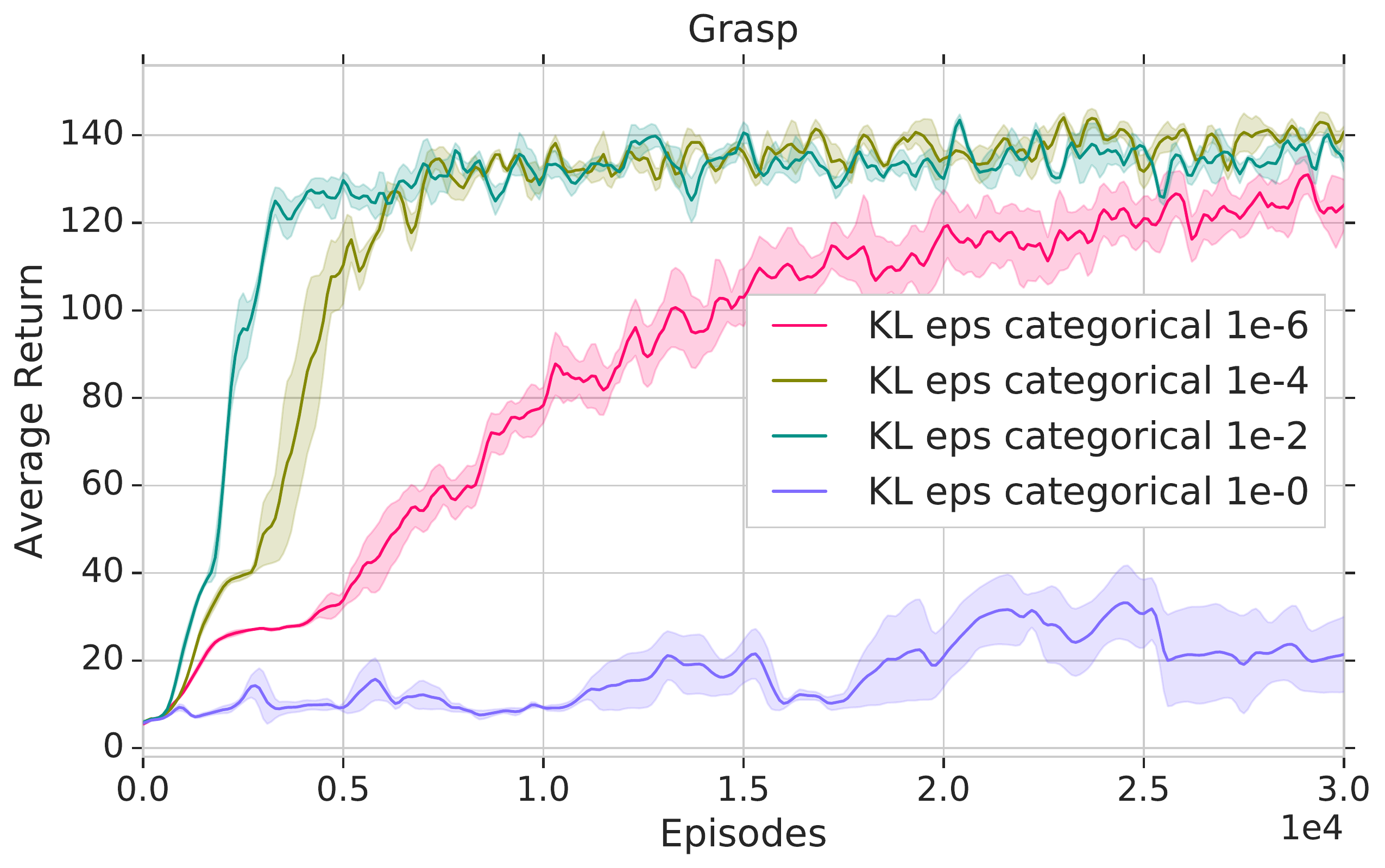}&
    \includegraphics[width=.3\textwidth]{figures/epsilon_ablations/epsilon_2.pdf} \\
    \includegraphics[width=.3\textwidth]{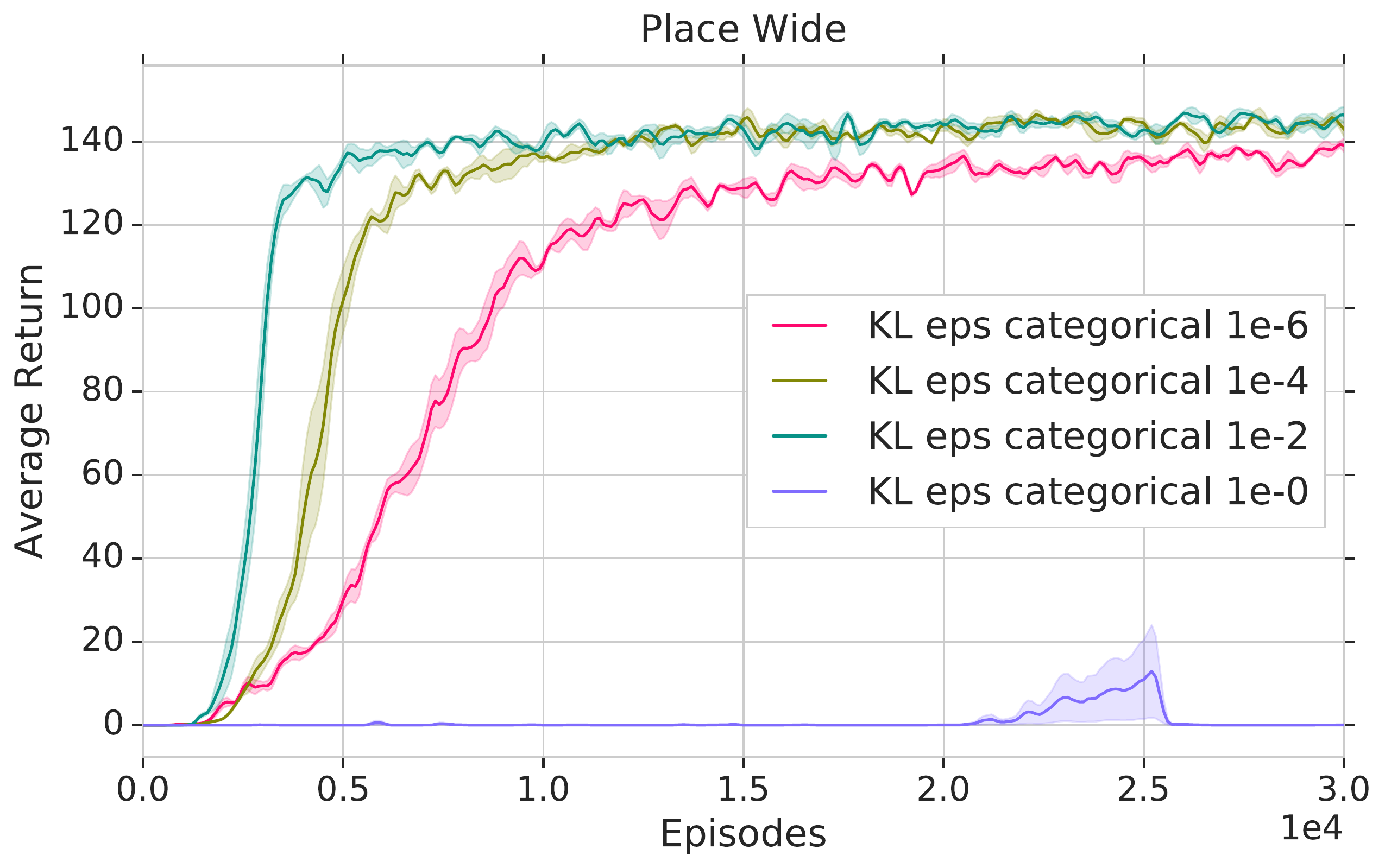}&  
    \includegraphics[width=.3\textwidth]{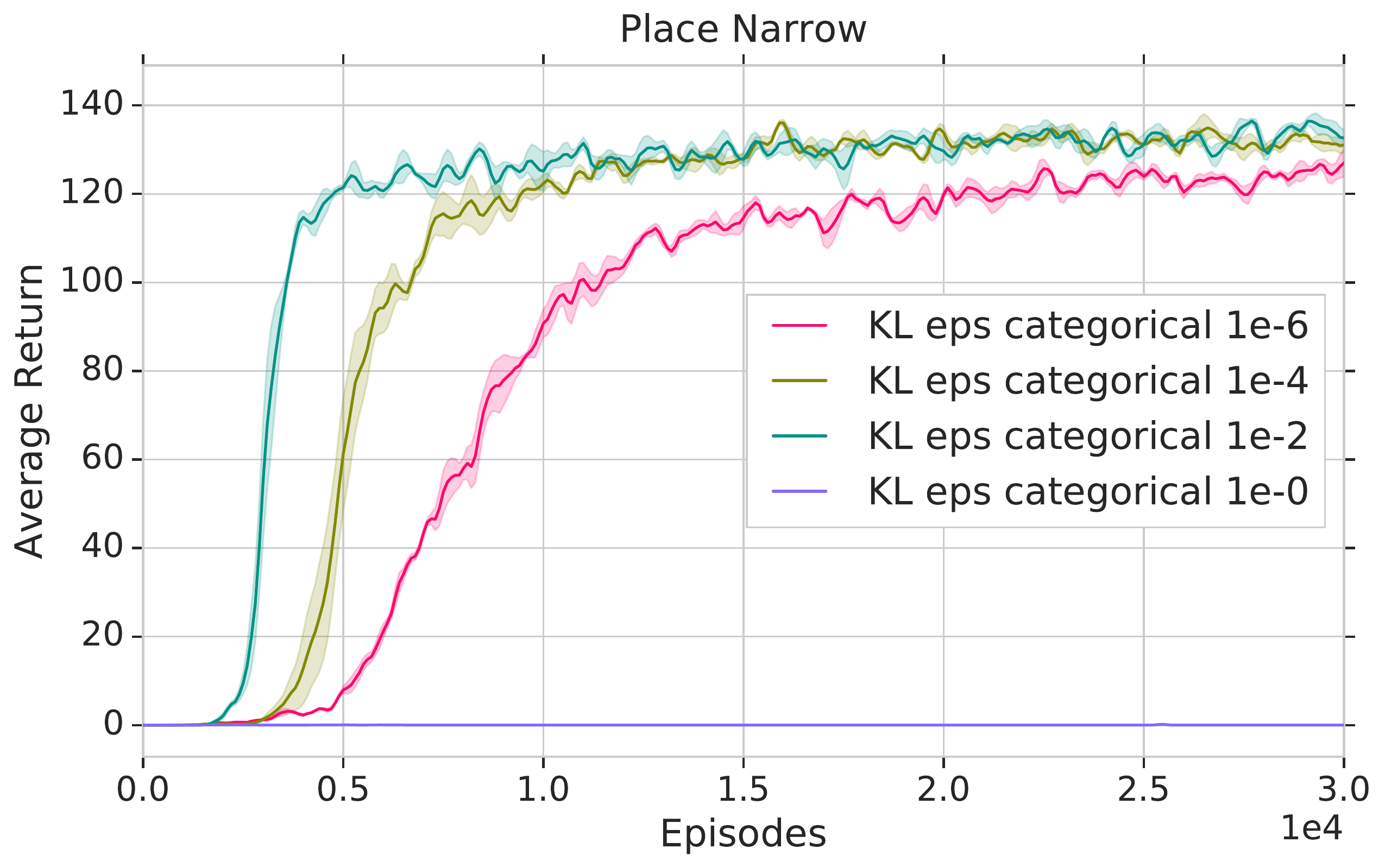}& 
    \includegraphics[width=.3\textwidth]{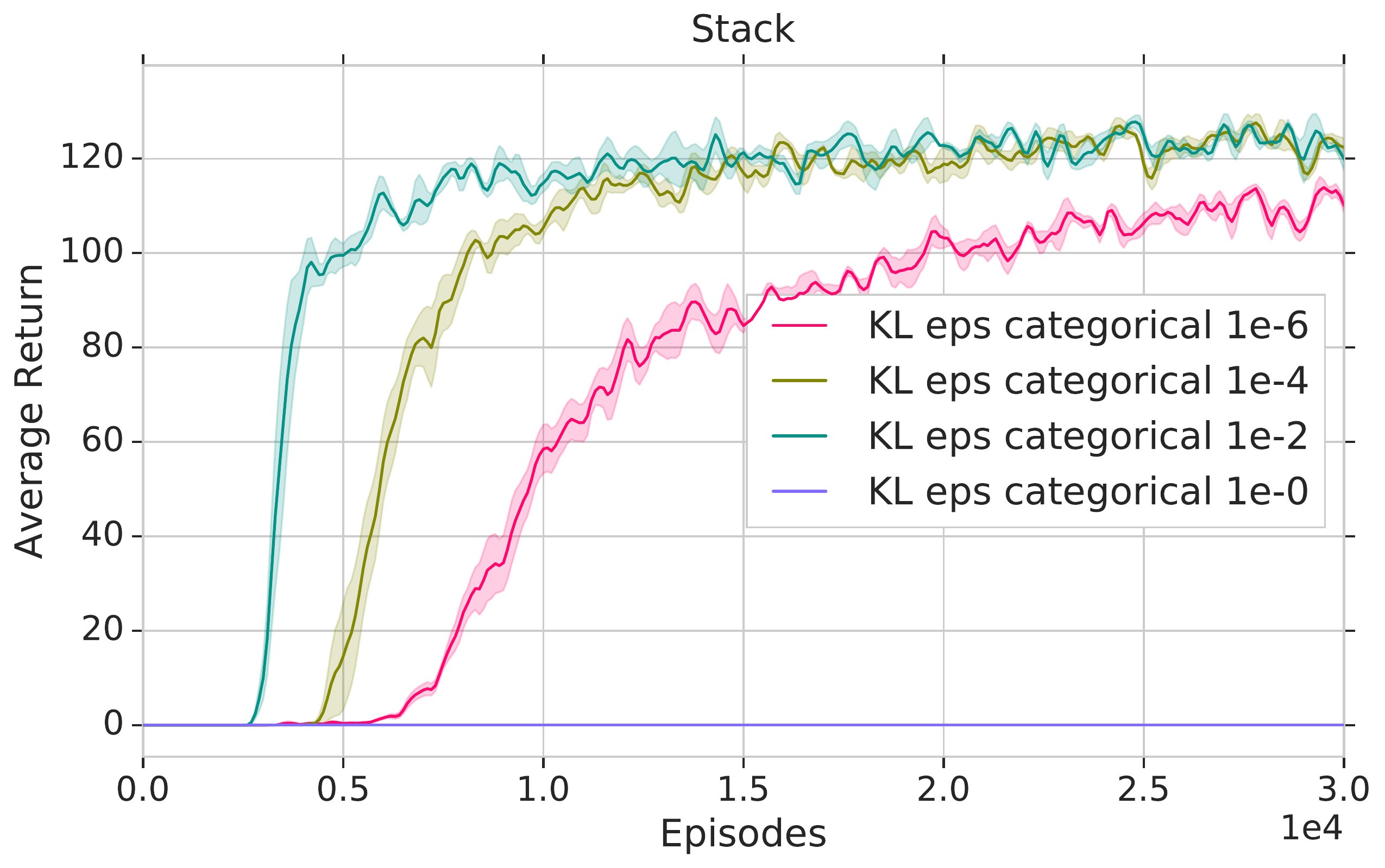} \\
    \includegraphics[width=.3\textwidth]{figures/epsilon_ablations/epsilon_6.pdf} 
    \end{tabular}
    \caption{\small Complete results for sweeping the KL constraint between 1e-6 and 1. in the Pile1 domain. For very weak constraints the model does not converge successfully, while for very strong constraints it only converges very slowly. }
    \label{fig:catkl_experiments}
\end{figure}

As depicted in Figure \ref{fig:catkl_experiments}, with a weak KL constraint, the high-level controller can converge too quickly leading to only a single sub-policy getting a gradient signal per step. In addition, the categorical distribution tends to change at a high rate, preventing successful convergence for the low-level policies. 
On the other hand, the low-level policies are missing task information to encourage decomposition as described in Section \ref{sec:method}. This fact, in combination with strong KL constraints, can prevent specialization of the low-level policies as the categorical remains near static, finally leading to no or very slow convergence. As long as a reasonable constraint is picked (here a range of over 2 orders of magnitude), convergence is fast and the final policies obtain high quality for all tasks. We note that no tuning of the constraints is required across domains and the range of admissible constraints is quite broad.

\subsubsection{Impact of Data Rate} \label{app:datarate}
\begin{figure}[H]
    \centering
    \begin{tabular}{c}
    \includegraphics[width=.5\textwidth]{figures/actor_ablations/actor_ablation_barplot_1_5_20_2.pdf}\\
    Lift\\
    \includegraphics[width=.5\textwidth]{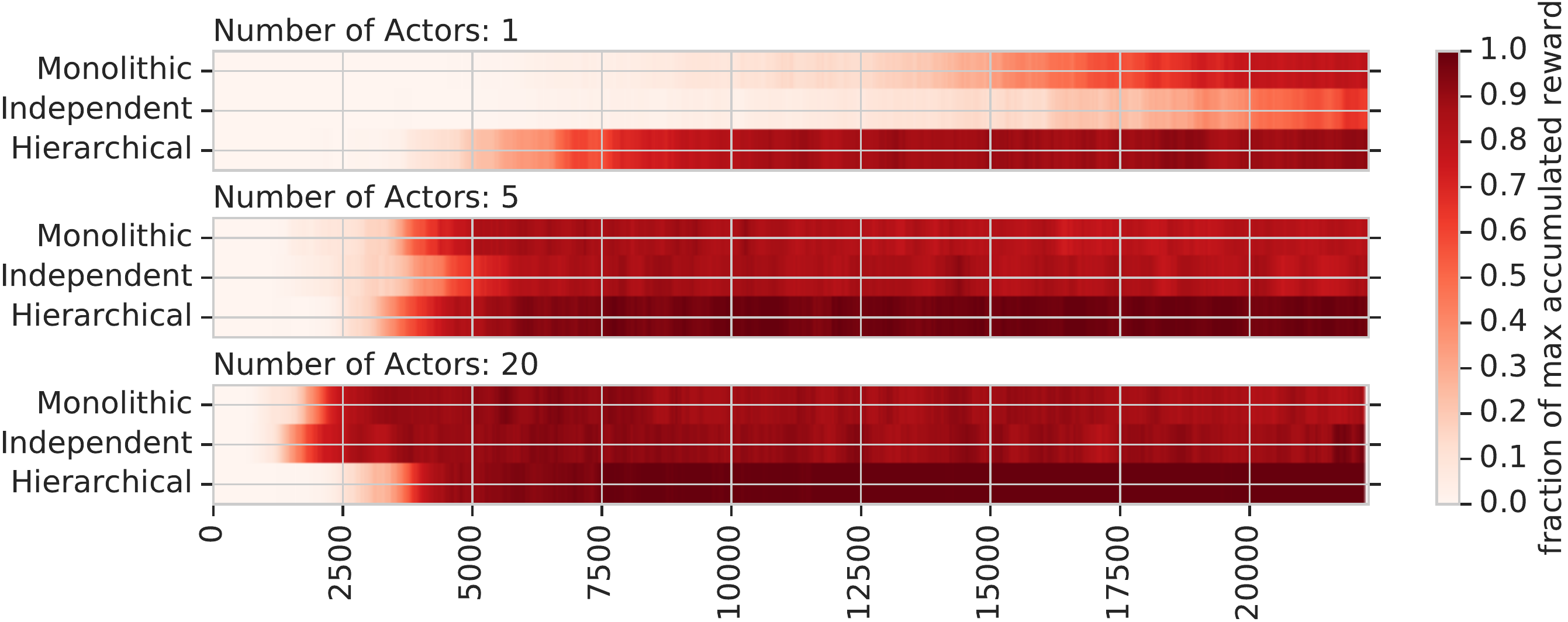}\\ 
    Place Wide\\
    \includegraphics[width=.5\textwidth]{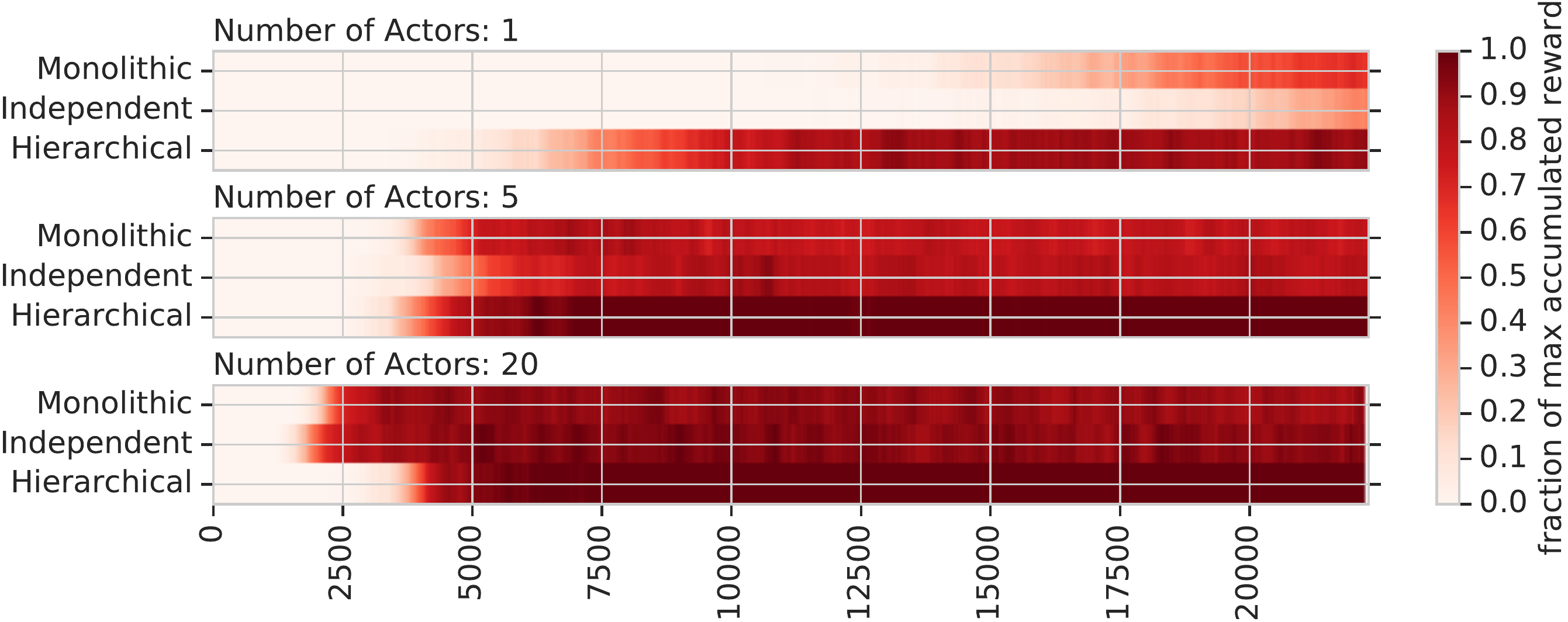}\\
    Place Narrow\\
    \includegraphics[width=.5\textwidth]{figures/actor_ablations/actor_ablation_barplot_1_5_20_5.pdf} \\
    Stack\\
    \includegraphics[width=.5\textwidth]{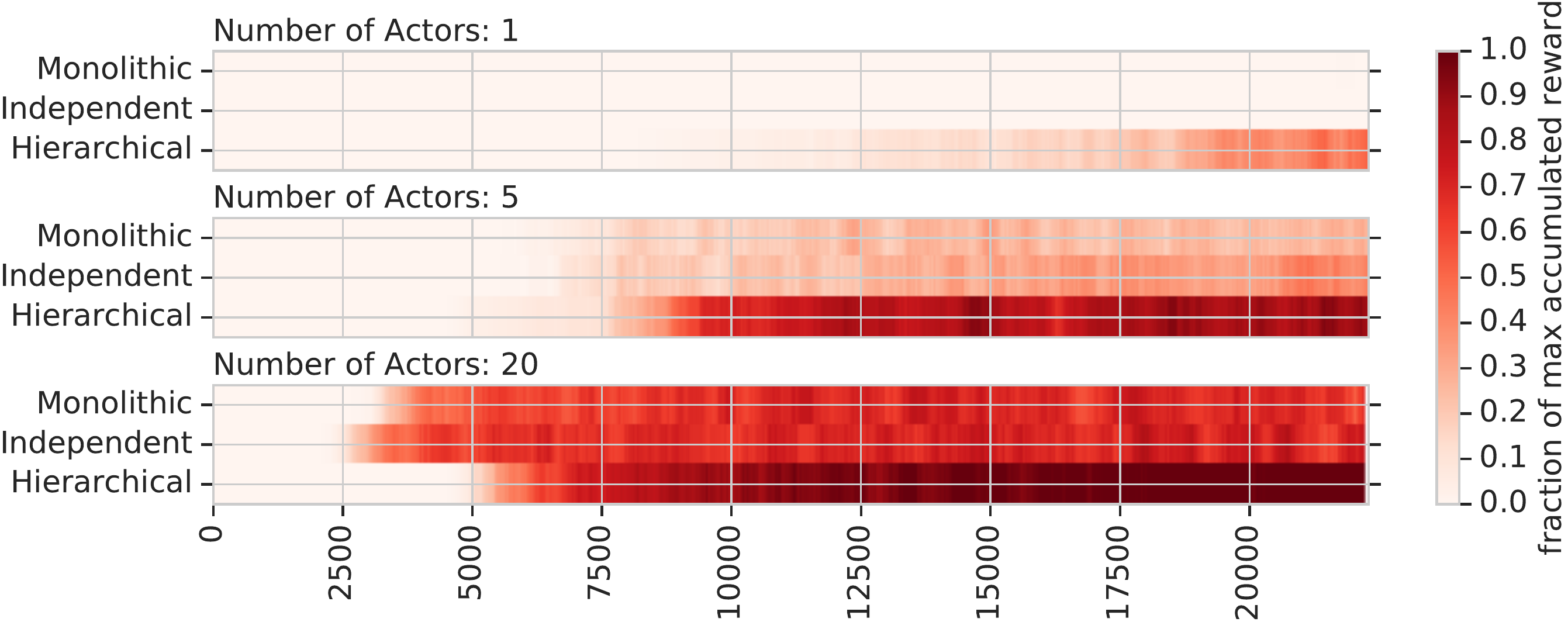} \\
    Stack and Leave
    \end{tabular}
    \caption{\small Complete results for ablating the number of data-generating actors in the Pile1 domain. We can see that the benefit of hierarchical policies is stronger for more complex tasks and lower data rates. However, even with 20 actors we see better final performance and stability}
    \label{fig:multitask_experiments_datarate}
\end{figure}

Evaluating in a distributed off-policy setting enables us to investigate the effect of different rates for data generation by controlling the number of actors.
Figure \ref{fig:multitask_experiments_datarate} demonstrates how the different agents converge slower lower data rates (changing from 5 to 1 actor). These experiments are highly relevant for the application domain as the number of available physical robots for real-world experiments is typically highly limited.
To limit computational cost, we focus on the simplest domain from Section \ref{sec:sim_multitask}, Pile1, in this comparison.  

\begin{figure}[H]
\centering
    \begin{tabular}{c}
    \includegraphics[width=.5\textwidth]{figures/actor_ablations/actor_ablation_barplot_1_5_20_0.pdf}\\
    Reach\\
    \includegraphics[width=.5\textwidth]{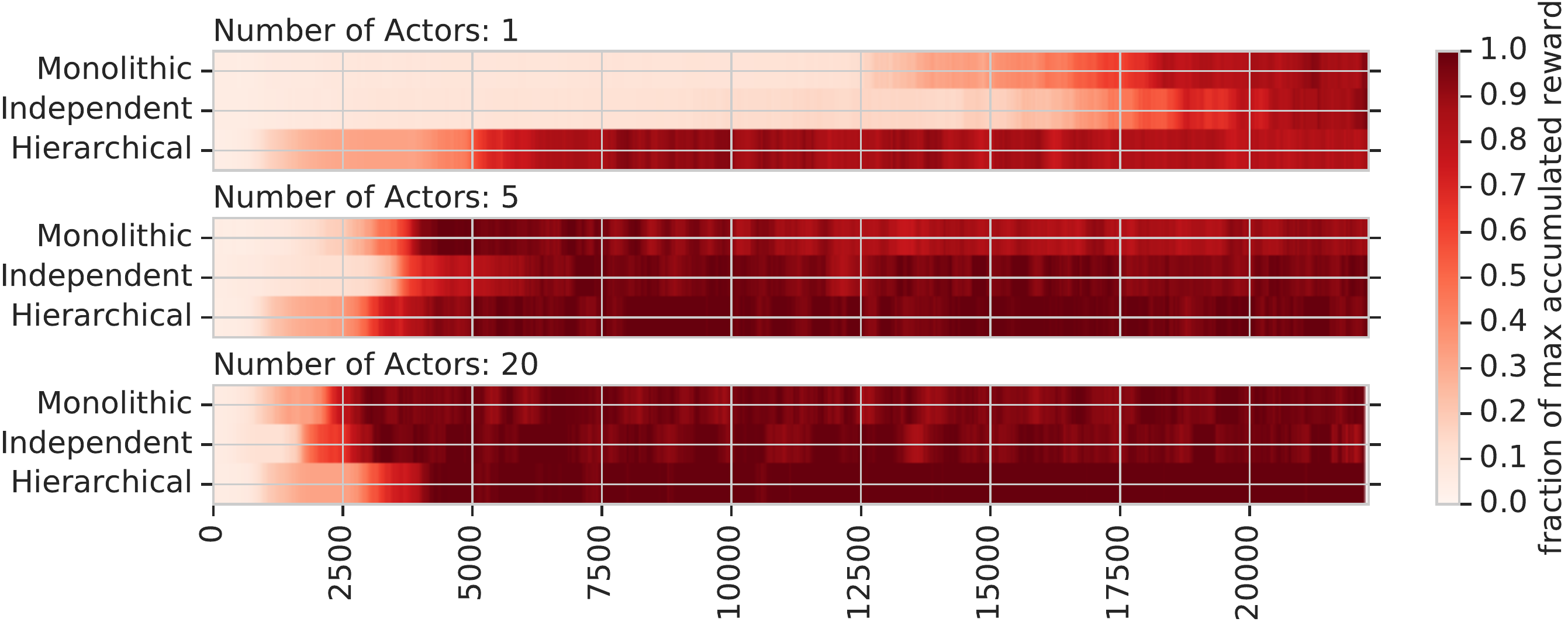}\\
    Grasp\\
    \end{tabular}
\end{figure}

\subsubsection{Number of Component Policies} \label{app:components}

\begin{figure}[H]
    \centering
    \begin{tabular}{ccc}
    \includegraphics[width=.3\textwidth]{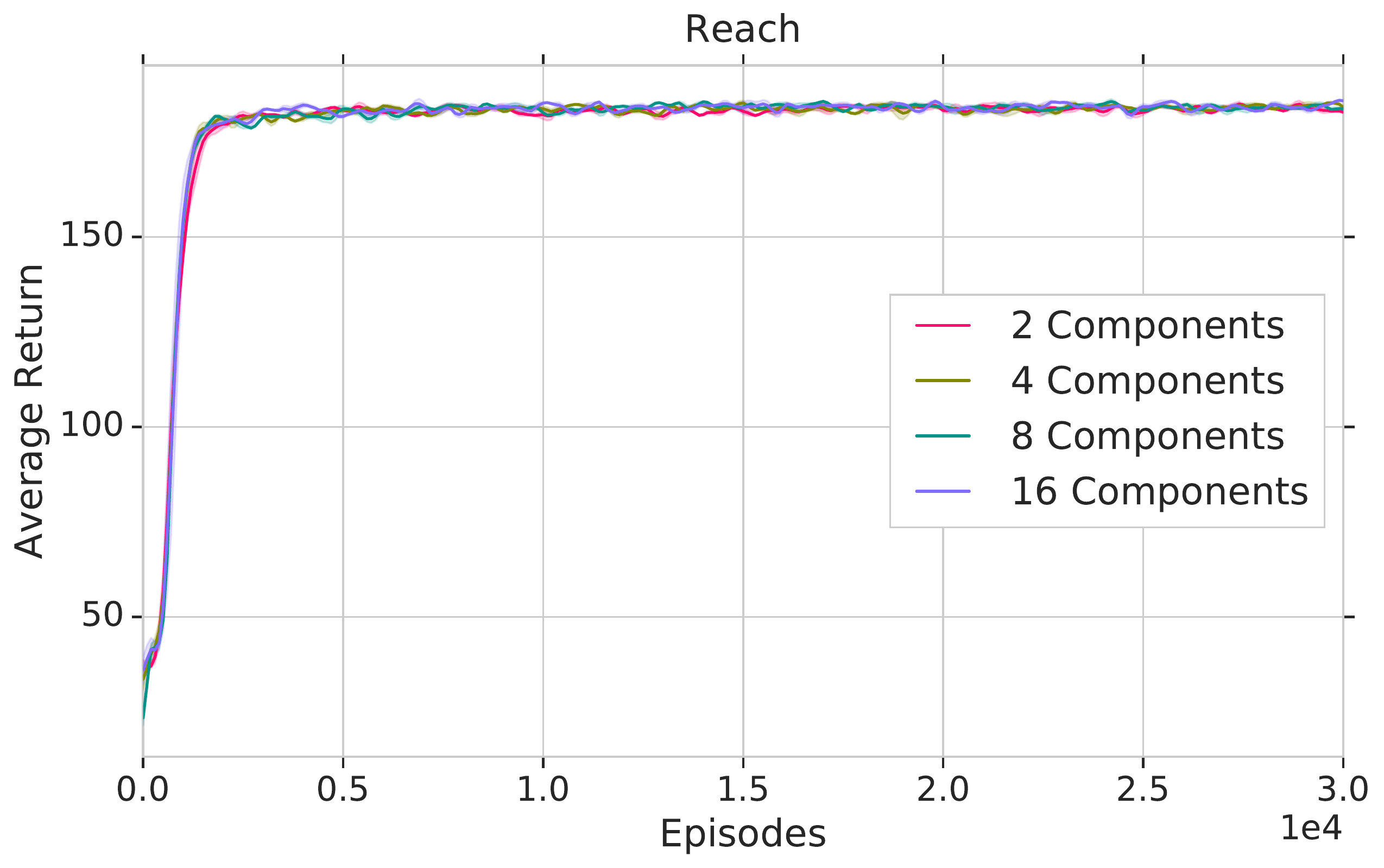}&  
    \includegraphics[width=.3\textwidth]{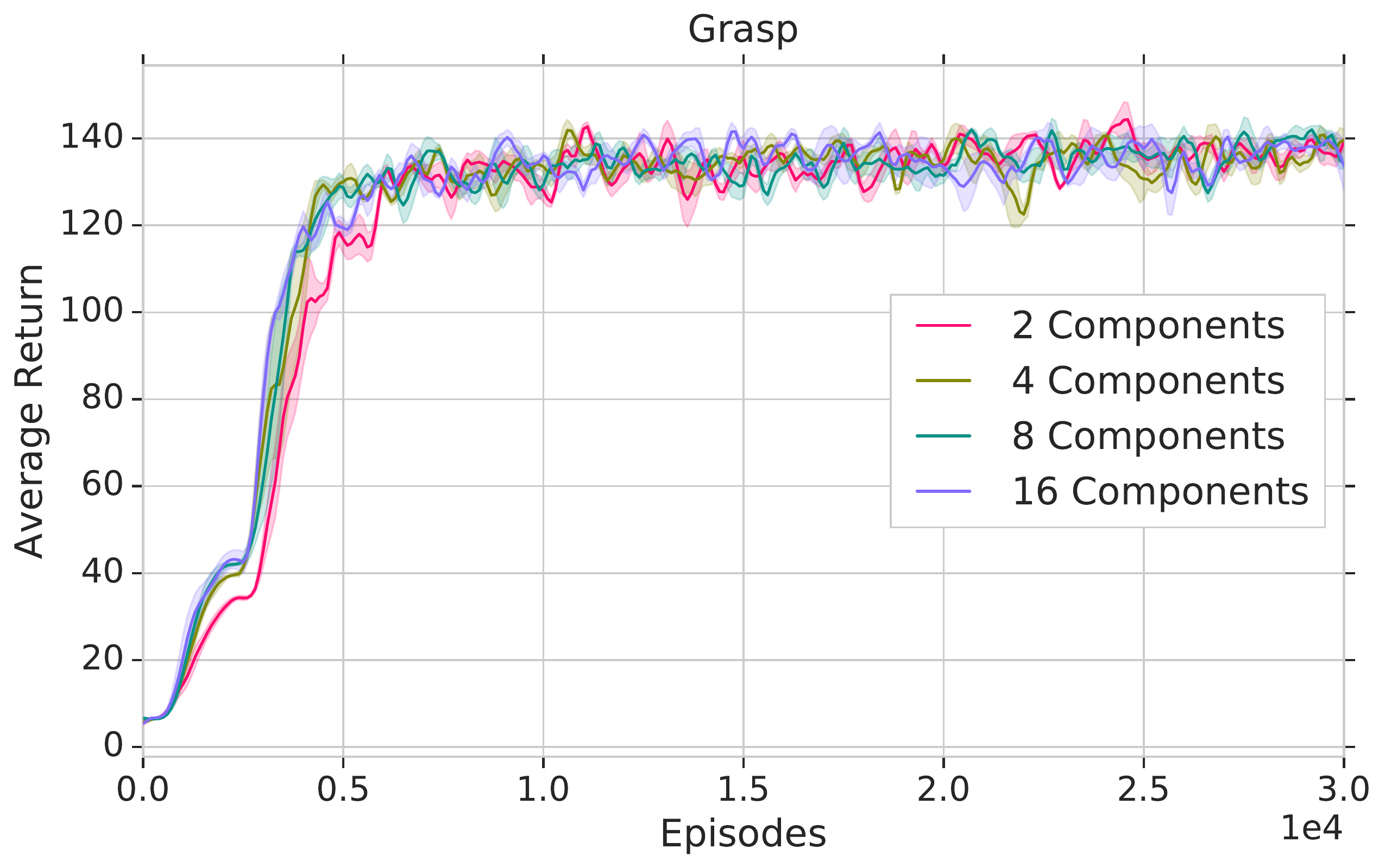}&
    \includegraphics[width=.3\textwidth]{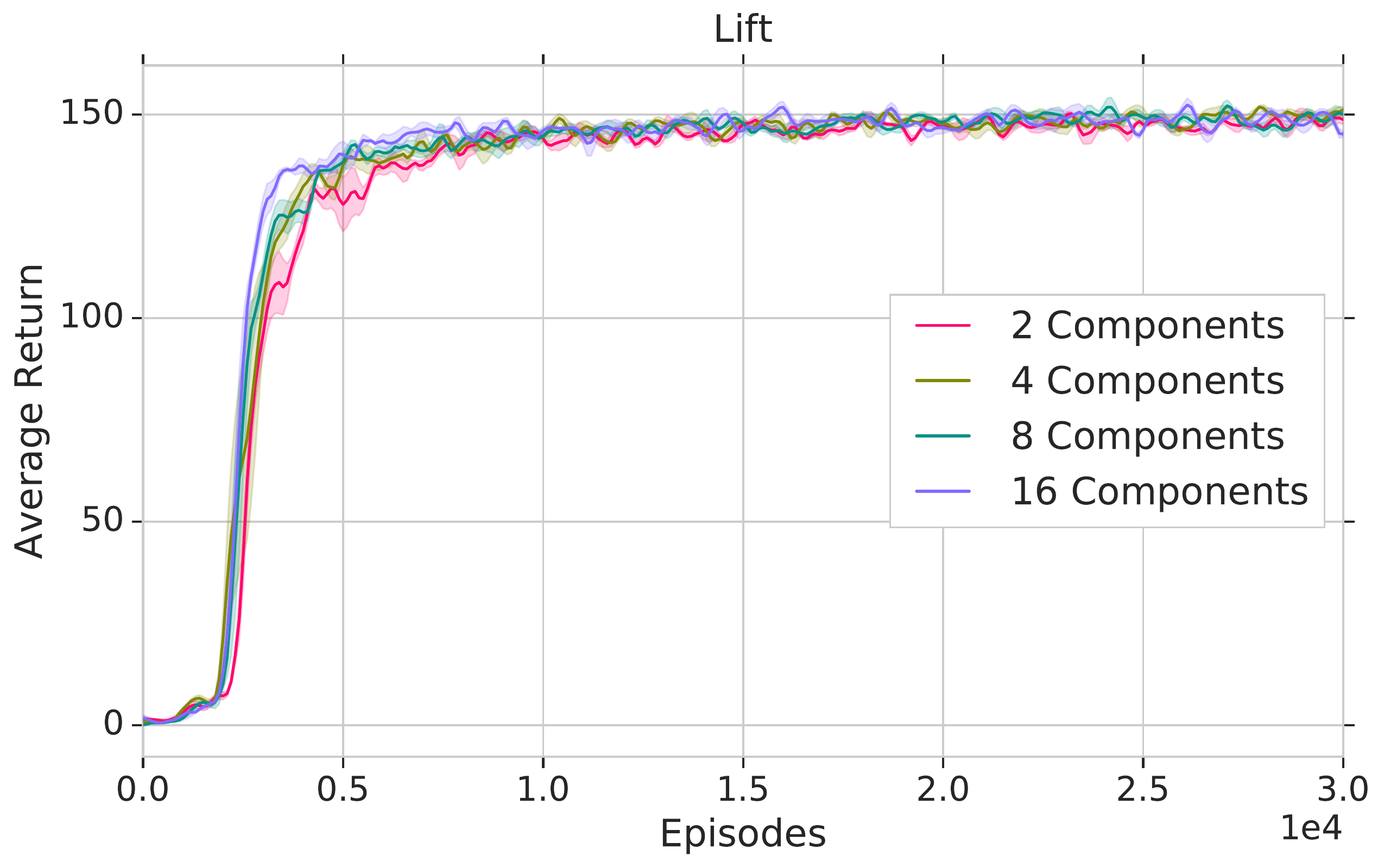} \\
    \includegraphics[width=.3\textwidth]{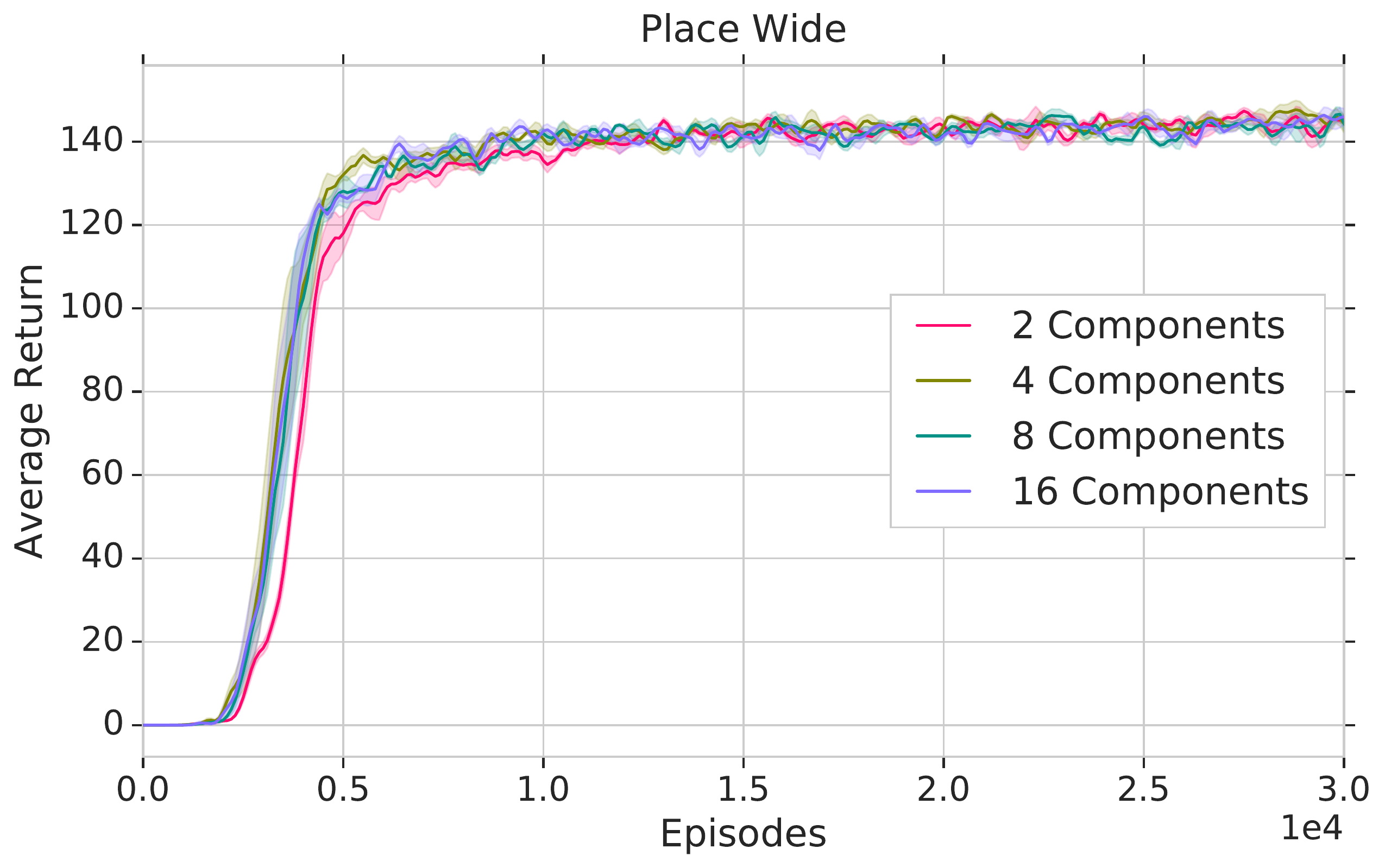}&  
    \includegraphics[width=.3\textwidth]{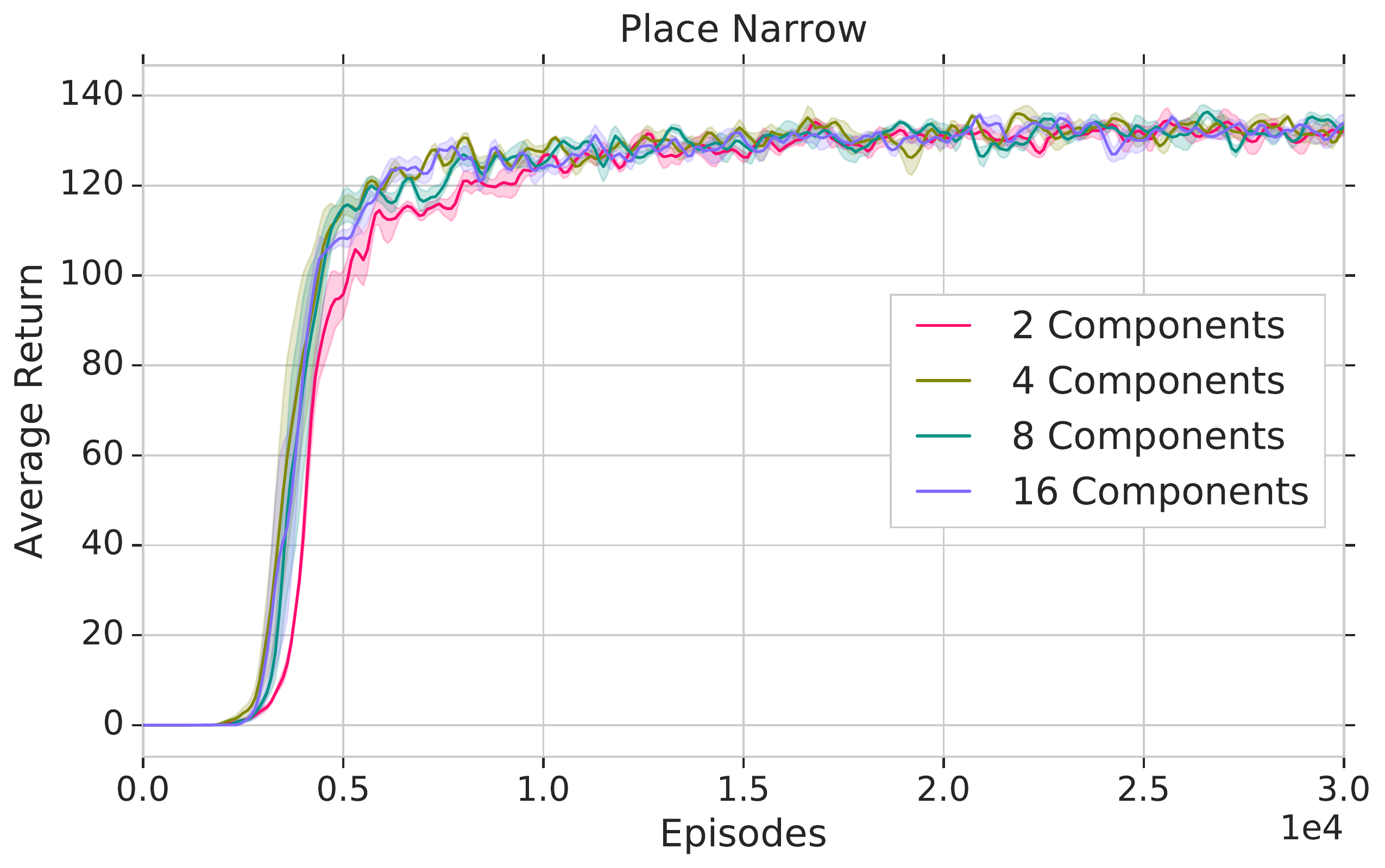}& 
    \includegraphics[width=.3\textwidth]{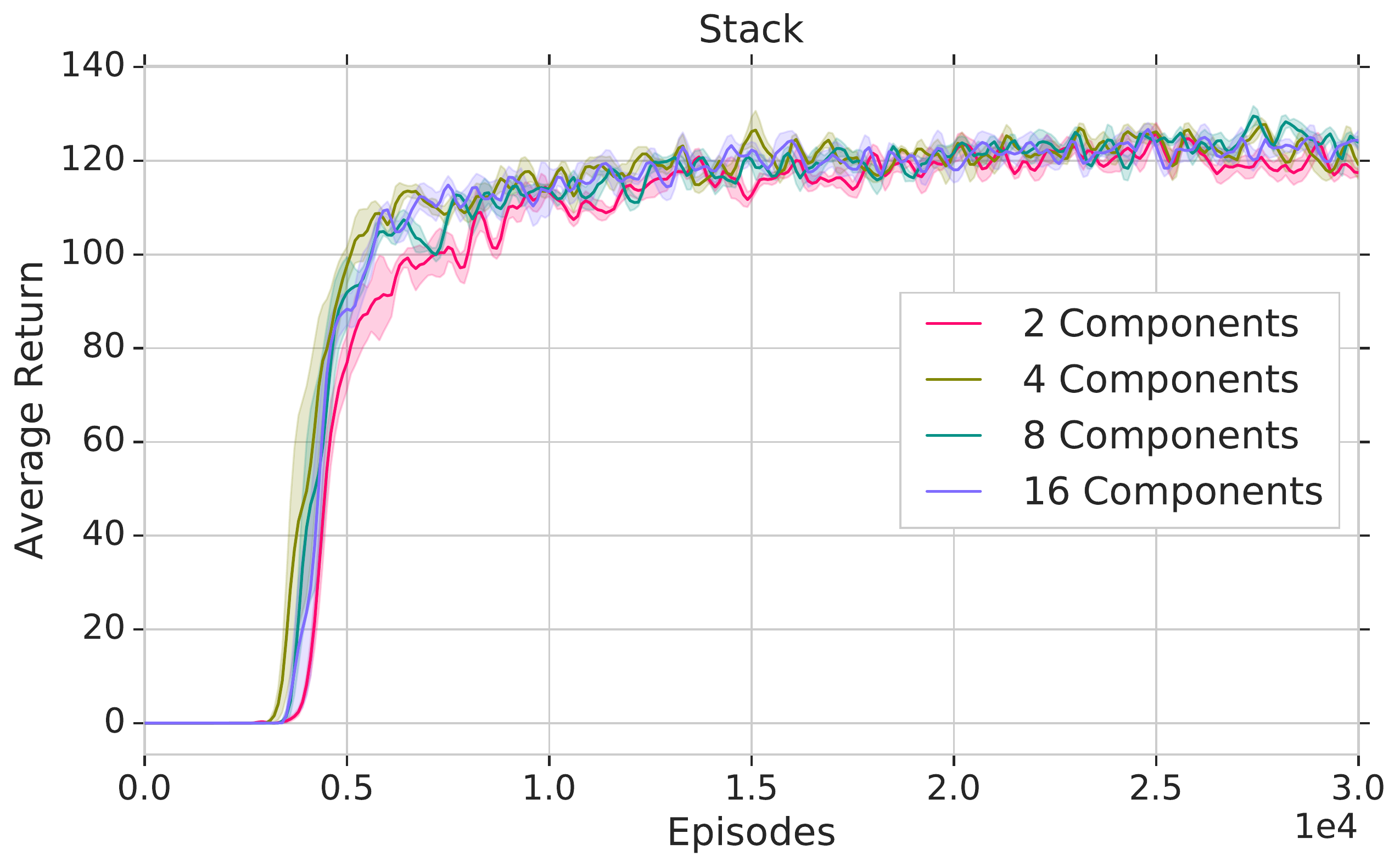} \\
    \includegraphics[width=.3\textwidth]{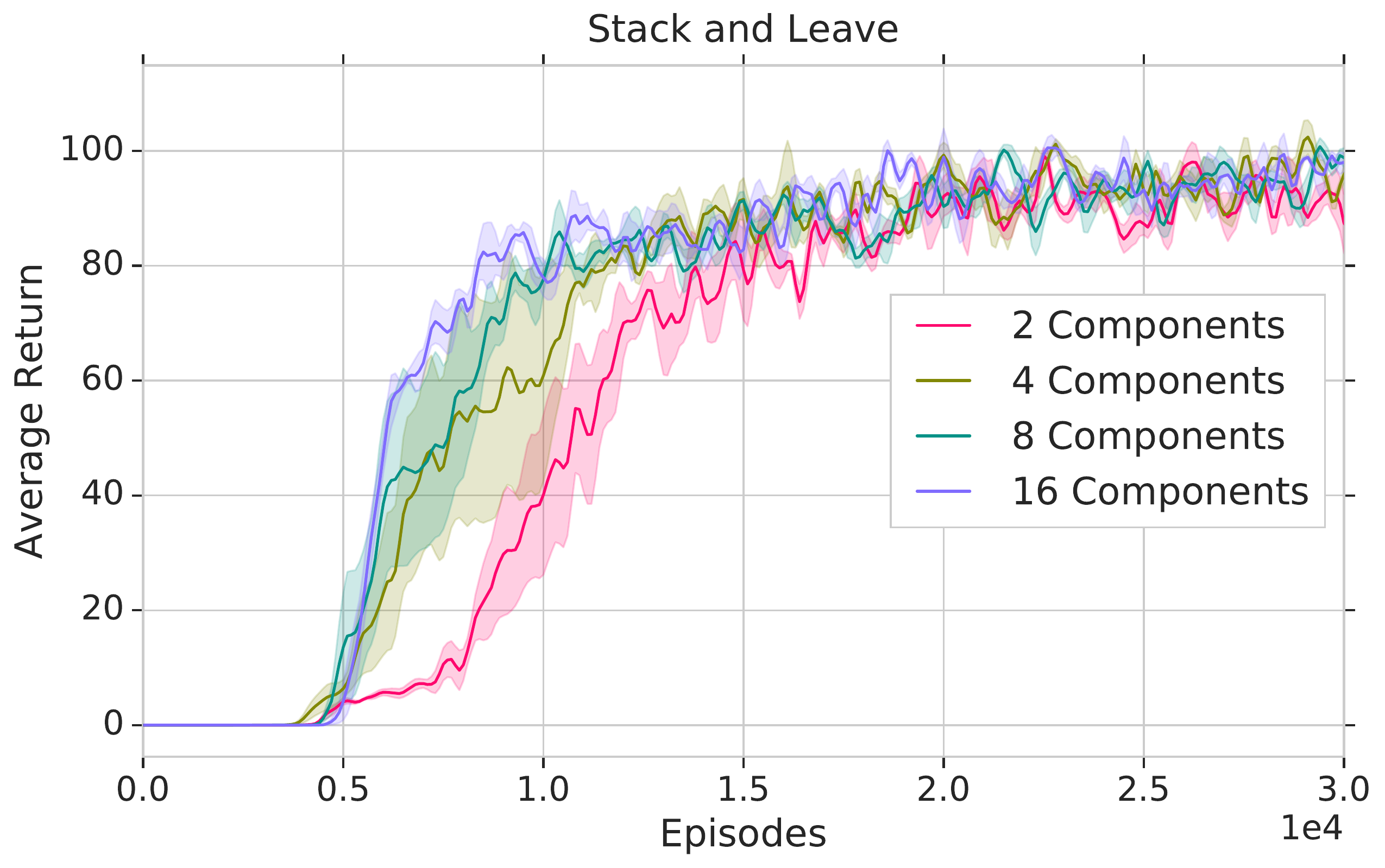} 
    \end{tabular}
    \caption{\small Complete results 2,4,8 and 16 low-level policies in the Pile1 domain. The approach is robust with respect to the number of sub-policies and we will build all further experiments on setting the components equal to the number of tasks.}
    \label{fig:multitask_experiments_components}
\end{figure}

\subsection{Hierarchical Policies in Reparameterization Gradient-based RL} \label{app:svg_experiments}
To test whether the benefits of a hierarchical policy transfer to a setting where a different algorithm is used to optimize the policy we performed additional experiments using SVG \citep{heess2015learning} in place of MPO. 
For this purpose we use the same hierarchical policy structure as for the MPO experiments but change the categorical to an implementation that enables reparameterization with the Gumbel-Softmax trick \citep{maddison2016concrete,jang2016categorical}. We then change the entropy regularization from Equation \eqref{eq:svg} to a KL towards a target policy (as entropy regularization did not give stable learning in this setting) and use 
%either a standard Gaussian distribution as the prior or 
a regularizer equivalent to the distance function (per component KL's from Equation \eqref{eq:mpo}) -- using a multiplier of 0.05 for the regularization multiplier was found to be the best setting via a coarse grid search. This is similar to previous work on hierarchical RL with SVG \citep{tirumala2019exploiting}.

This extension of SVG is conceptually similar to a single-step-option version of the option-critic \citep{bacon2017option}. Simplified, SVG is an off-policy actor-critic algorithm which builds on the reparametrisation instead of likelihood ratio trick (commonly leading to lower variance \citep{mohamed2019monte}). Since we do not build on temporally extended sub-policies, the algorithm simplifies to using a single critic (see Section \ref{sec:method}).

\begin{figure}[H]
    \centering
    \begin{tabular}{ccc}
    \includegraphics[width=.3\textwidth]{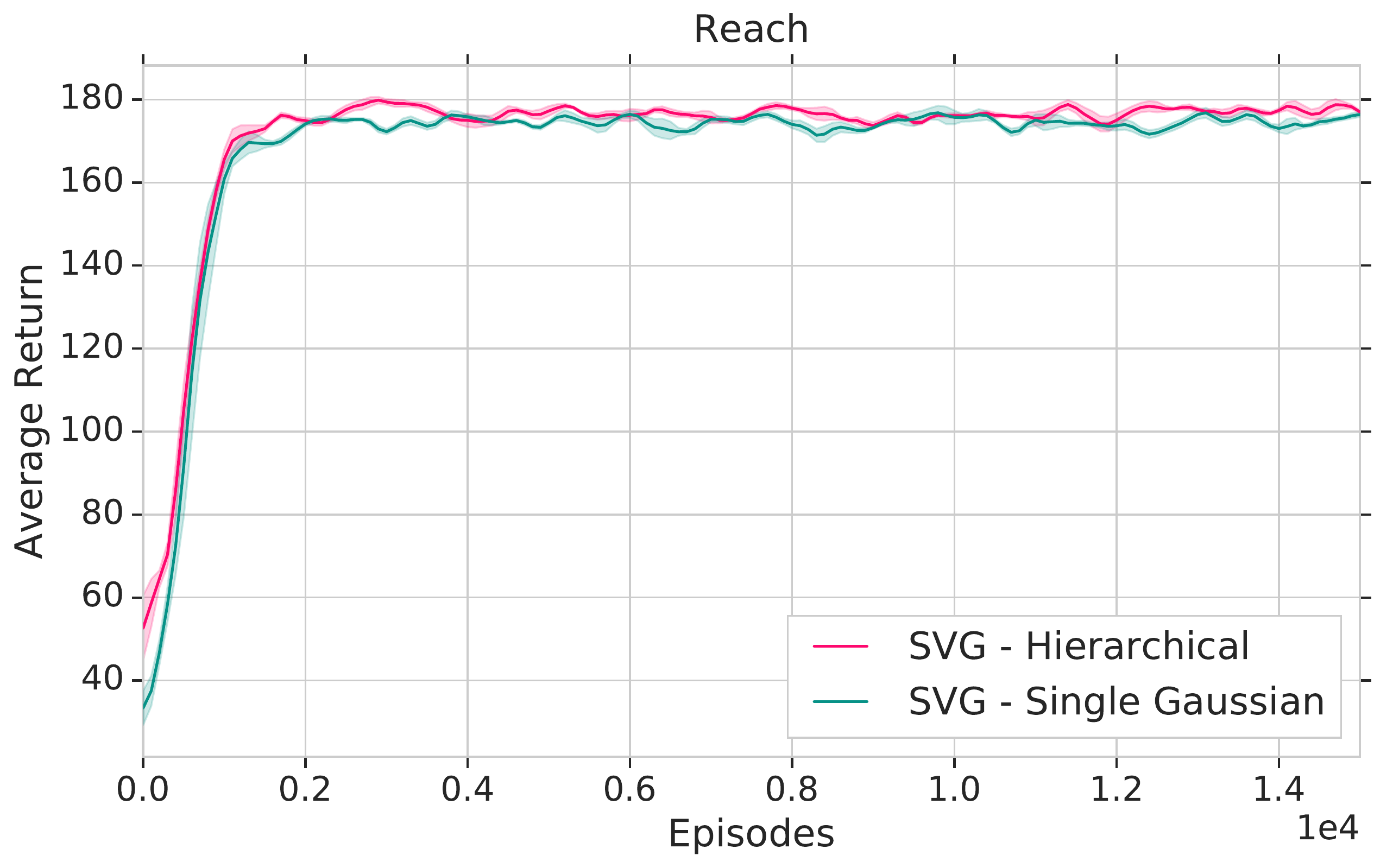}&  
    \includegraphics[width=.3\textwidth]{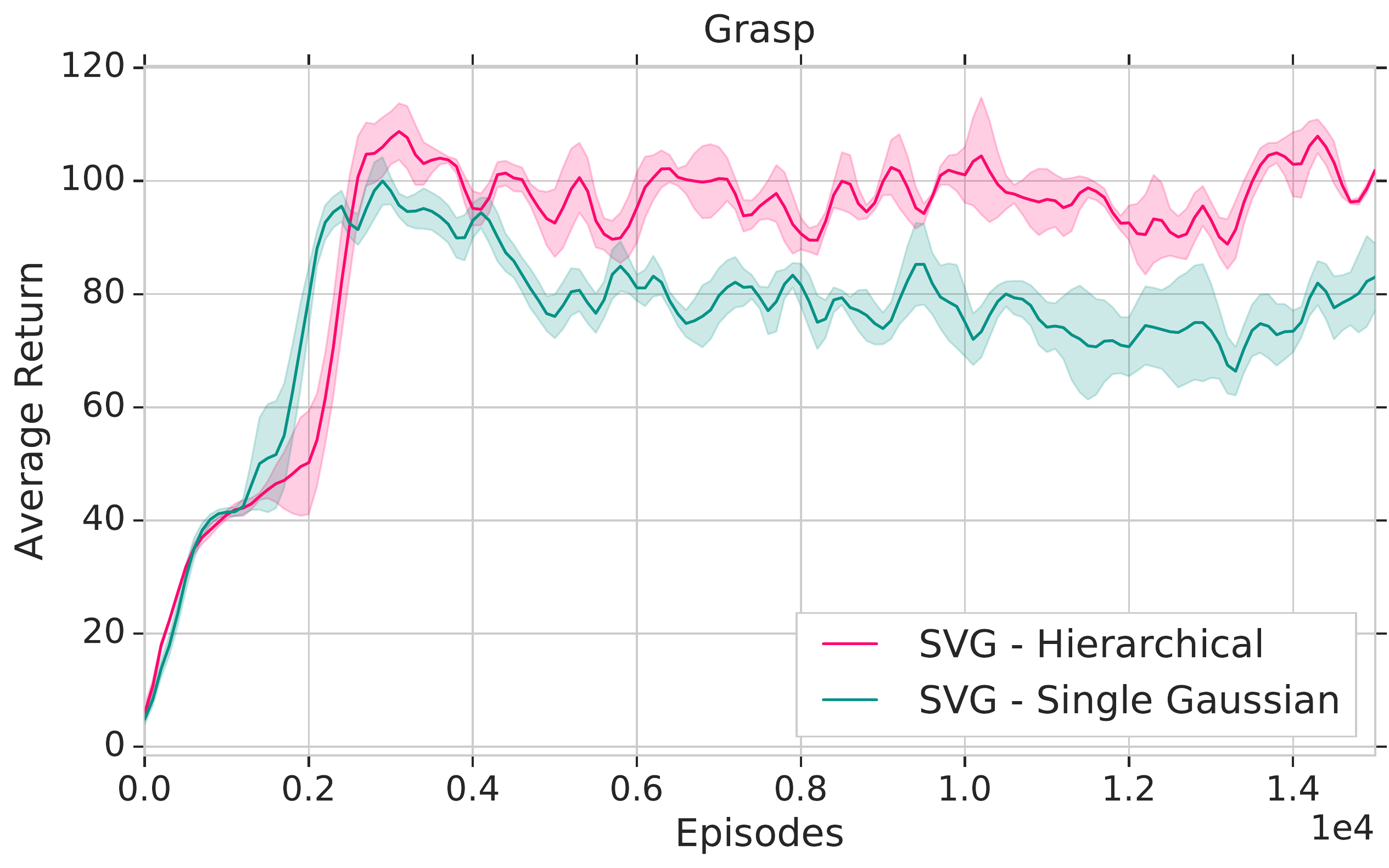}  & 
    \includegraphics[width=.3\textwidth]{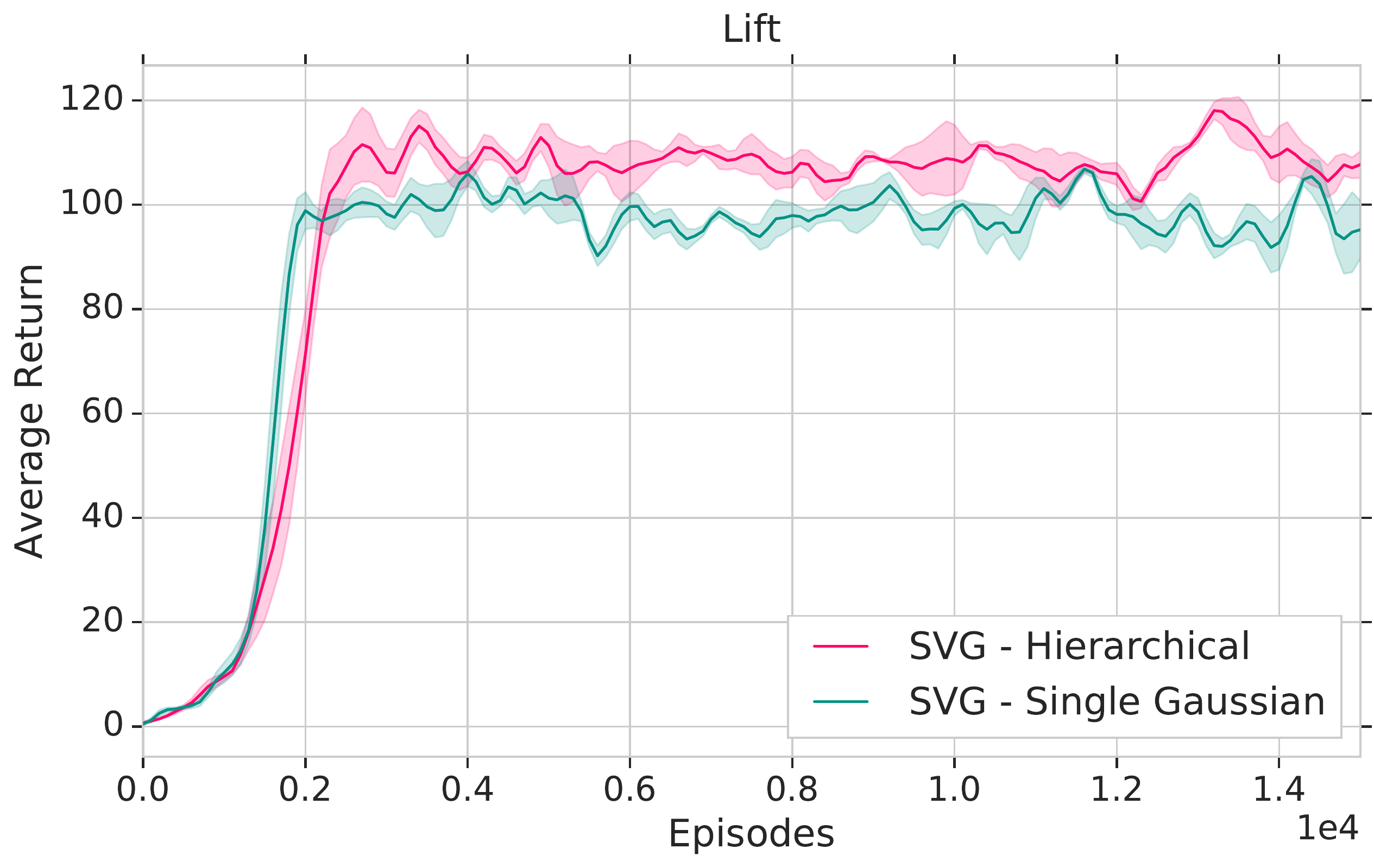} \\
    \includegraphics[width=.3\textwidth]{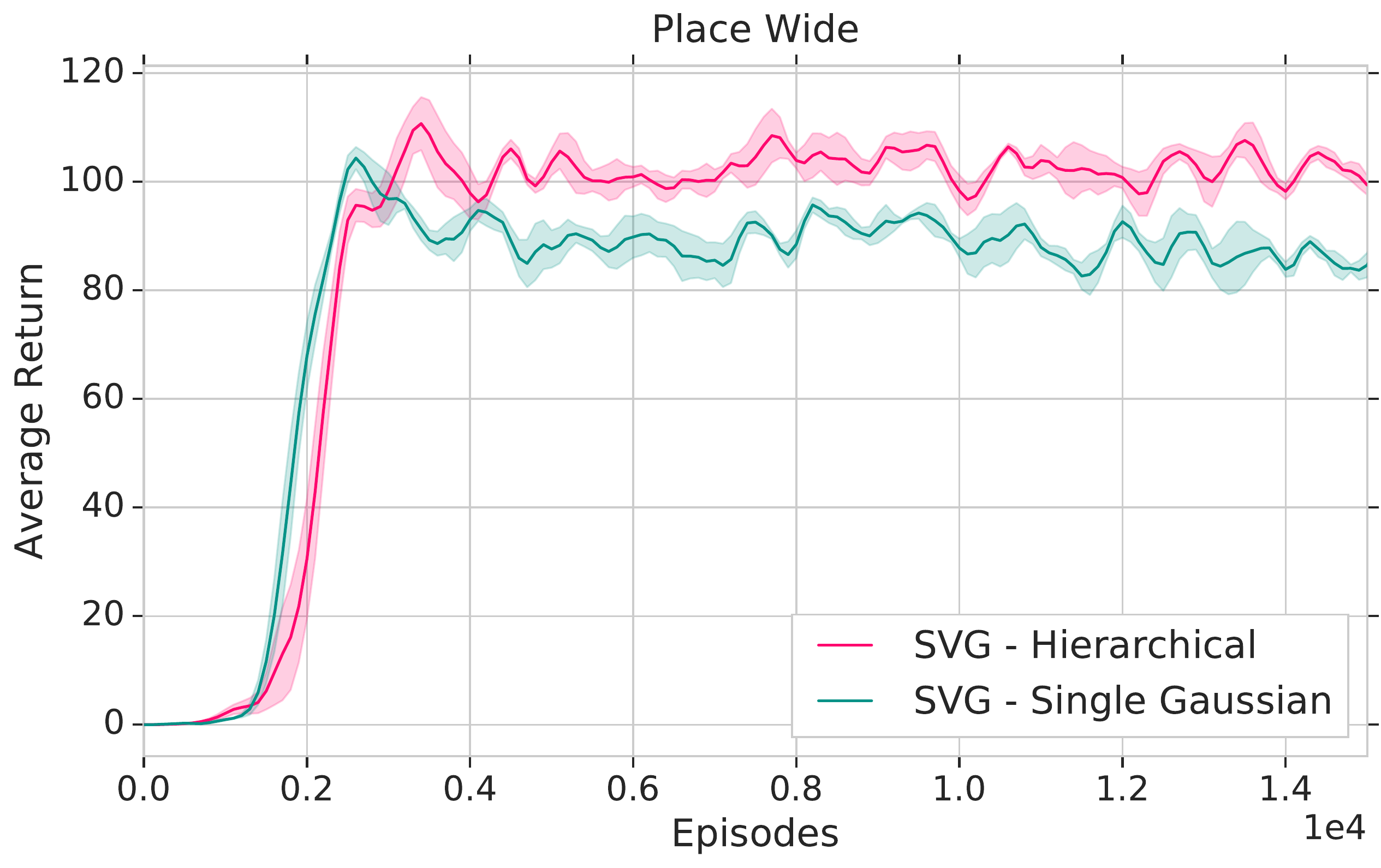}&  
    \includegraphics[width=.3\textwidth]{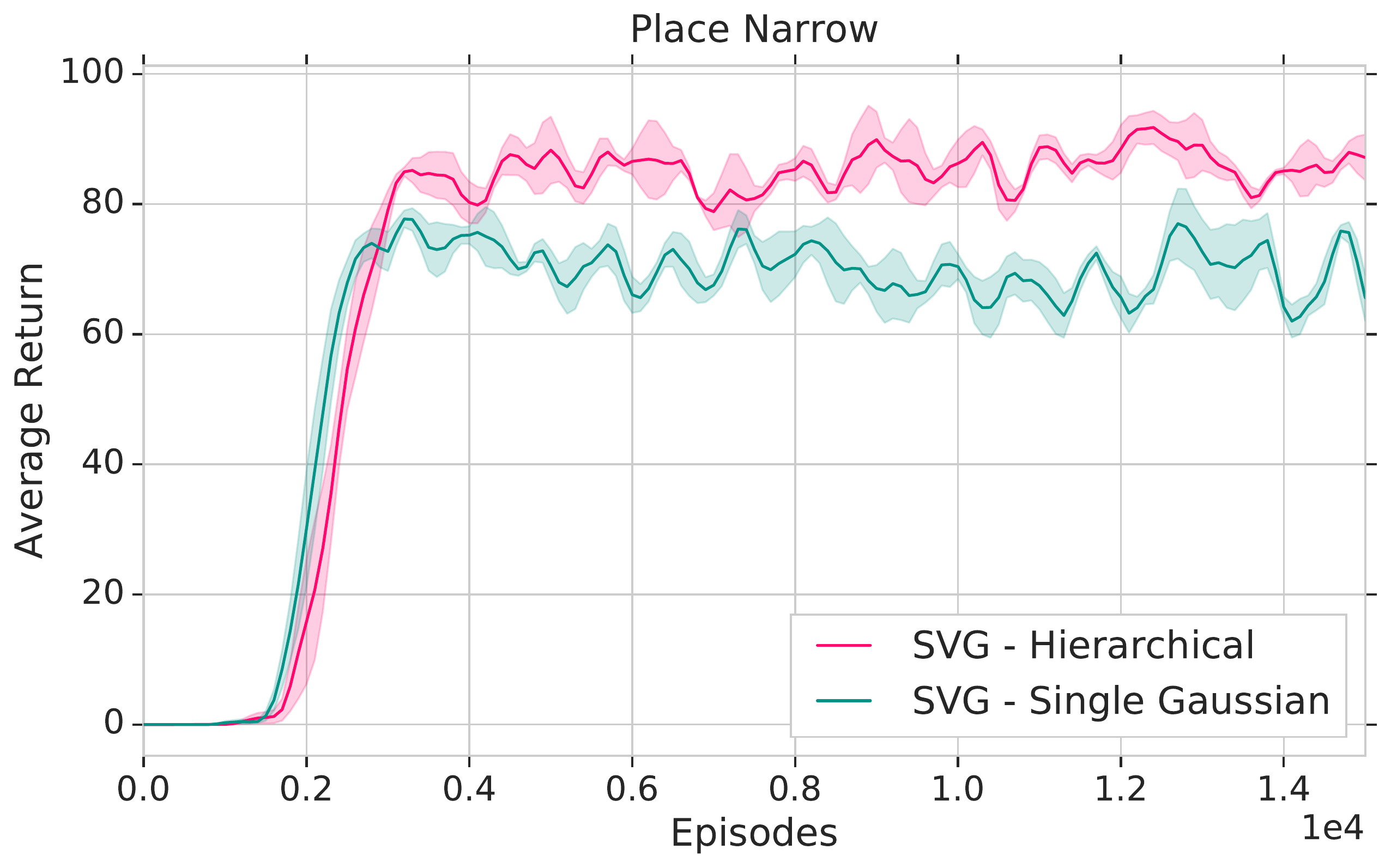}  & 
    \includegraphics[width=.3\textwidth]{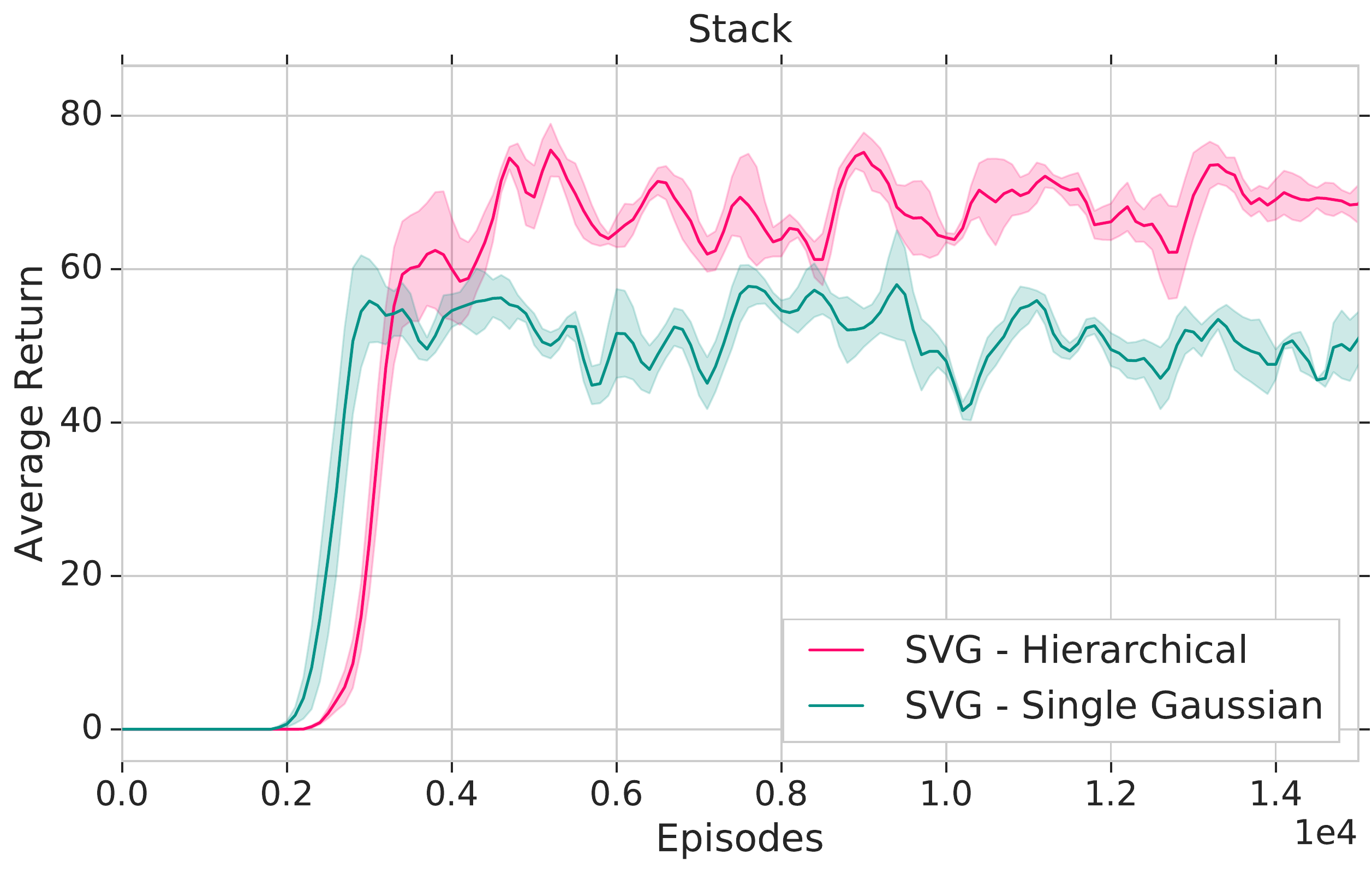}\\  
    \includegraphics[width=.3\textwidth]{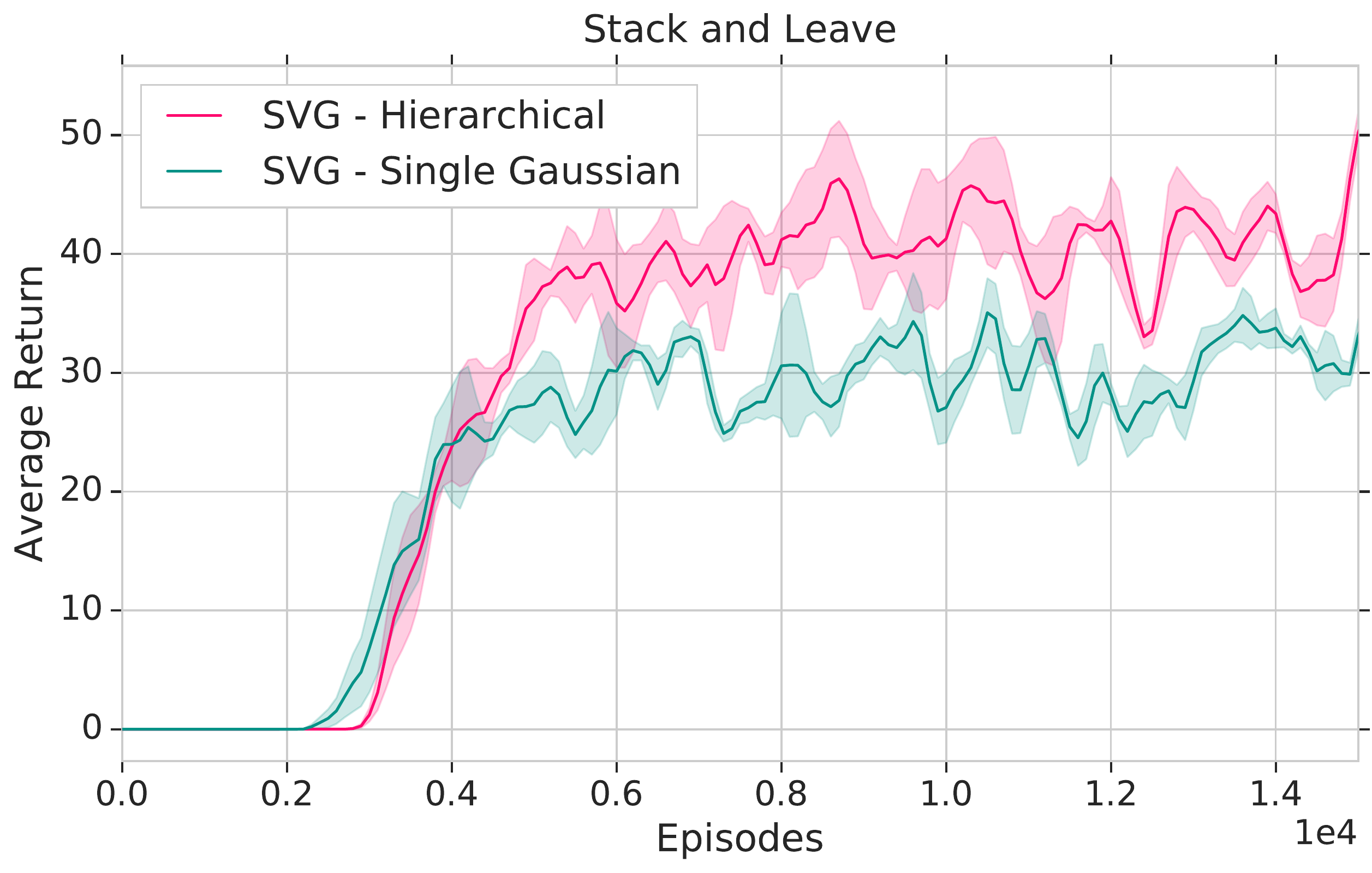}  
    \end{tabular}
    \caption{\small Complete results for evaluating SVG with and without hierarchical policy class in the Pile1 domain. Similarly to the experiments in the main paper, we can see that the hierarchical policy leads to better final performance -- here for a gradient-based approach. All plots are generated by running 5 actors in parallel.}
    \label{fig:multitask_experiments_gradient}
\end{figure}

The results of this experiment are depicted in Figure \ref{fig:multitask_experiments_gradient}, as can be seen, for this simple domain results in mild improvements over standard SAC-U.

\end{document}